\documentclass[journal]{IEEEtran}
\usepackage{savesym}
\usepackage{dingbat}
\savesymbol{checkmark}

\usepackage{amssymb}
\usepackage{tikz}

\newcommand\circledmark[1][red!50]{%
  \tikz\node[circle,fill=#1,inner sep=2.3pt]{$\checkmark$};%
}

\ifCLASSINFOpdf
\else
\fi
\usepackage{array}
\usepackage{graphicx}

\usepackage{pifont}
\newcommand{\xmark}{\ding{55}}%

 \usepackage{booktabs}
\usepackage{multirow}

\usepackage{dingbat}

\usepackage{booktabs}
\usepackage{graphicx}

\begin{document}
%
\title{Trustworthy Federated Learning: A Survey }
%
%
%

\author{Asadullah Tariq,~\IEEEmembership{}
        Mohamed Adel Serhani,~\IEEEmembership{}
        Farag Sallabi,~\IEEEmembership{}
        Tariq Qayyum,~\IEEEmembership{}
        Ezedin S. Barka,~\IEEEmembership{}
        Khaled A. Shuaib,~\IEEEmembership{}
\thanks{Asadullah Tariq is with CIT, United Arab Emirates university,
UAE e-mail: (700039114@uaeu.ac.ae).}
\thanks{ABC and ABC are with Anonymous University.}
\thanks{Manuscript received April 19, 2005; revised August 26, 2015.}}

%
%

\markboth{Journal of \LaTeX\ Class Files,~Vol.~14, No.~8, August~2023}%
{Shell \MakeLowercase{\textit{et al.}}: Bare Demo of IEEEtran.cls for IEEE Journals}
%



\maketitle

\begin{abstract}
Federated Learning (FL) has emerged as a significant advancement in the field of Artificial Intelligence (AI), enabling collaborative model training across distributed devices while maintaining data privacy. As the importance of FL increases, addressing trustworthiness issues in its various aspects becomes crucial. In this survey, we provide an extensive overview of the current state of Trustworthy FL, exploring existing solutions and well-defined pillars relevant to Trustworthy . Despite the growth in literature on trustworthy centralized  Machine Learning (ML)/Deep Learning (DL), further efforts are necessary to identify trustworthiness pillars and evaluation metrics specific to FL models, as well as to develop solutions for computing trustworthiness levels. We propose a taxonomy that encompasses three main pillars: Interpretability, Fairness, and Security \& Privacy. Each pillar represents a dimension of trust, further broken down into different notions. Our survey covers trustworthiness challenges at every level in FL settings. We present a comprehensive architecture of Trustworthy FL, addressing the fundamental principles underlying the concept, and offer an in-depth analysis of trust assessment mechanisms. In conclusion, we identify key research challenges related to every aspect of Trustworthy FL and suggest future research directions. This comprehensive survey serves as a valuable resource for researchers and practitioners working on the development and implementation of trustworthy FL systems, contributing to a more secure and reliable AI landscape.
\end{abstract}


%
\IEEEpeerreviewmaketitle

\section{Introduction}

%
%
%
%

\IEEEPARstart{T}{rustworthy} AI is an evolving concept within responsible AI that encompasses various existing ideas, such as ethical AI \cite{1}, robust AI \cite{2}, explainable AI (XAI) \cite{3} and fair AI \cite{4}, among others. In addition to trustworthiness, data privacy and protection have become increasingly important in today's society. AI systems, particularly those utilizing machine learning (ML) and deep learning (DL) models, are often trained on data owned and maintained by various stakeholders across distinct data silos. To address the challenge of preserving data privacy in AI, Google introduced Federated Learning (FL) \cite{5} in 2016 as a decentralized machine learning paradigm. FL enables collaborative model building among federation members while ensuring sensitive data remains within each participant's control. FL provides a solution to data silo and fragmentation issues resulting from legislation that restricts data sharing and requires data owners to maintain isolation. As such, FL is seen as a crucial component in the present and future of AI, with its worldwide market valuation is anticipated to grow from USD 127 million in 2023 to USD 210 million by 2028 \cite{6}.

FL is a cutting-edge AI that preserves privacy by  allowing clients, such as organizations or devices, to train models locally and construct a global model from local updates without sharing local data externally \cite{6.0}. Trustworthy FL is essential to responsible AI principles since it faces challenges like accountability and fairness due to the involvement of multiple parties and diverse client data distribution. This article introduces the concept of Trustworthy FL. Trustworthy AI aims to achieve high learning accuracy consistently while considering aspects like interpretability, fairness, privacy, robustness, accountability, and environmental wellbeing. As a promising trustworthy AI framework, FL ensures data privacy during AI model training by coordinating multiple devices. In a server-client structure for FL, individual devices carry out local training and transmit their updated local models to a central server, all while keeping their private raw data secure. The server aggregates parameters from local models to refine the global FL model before redistributing the enhanced model back to the devices. This process maintains privacy for sensitive applications in areas like healthcare, finance, and the Internet of Things (IoT).

However, existing FL schemes often face vulnerabilities to attacks and difficulties meeting security requirements in practical applications \cite{6.1}. Central servers, which distribute, aggregate, and update models, are attractive targets for malicious actors who might tamper with models to produce biased outcomes or desired results. Dishonest cloud service providers may deliver incorrect results to users or replace original models with simpler, less accurate ones to save on computational costs.
To bolster FL security, it is essential to verify the integrity and authenticity of model updates throughout training to prevent malicious attacks. Secure FL systems must be resilient against dropout clients who fail to submit model updates for aggregation due to issues like poor network connections, temporary unavailability, or energy constraints \cite{6.2}. Edge devices, with their widespread presence and easy internet access, are ideal candidates for quality training in various applications. However, participation is limited due to potential data leakage and inherent security issues, such as malicious participation, unfair contribution, biased aggregation, and central server bottlenecks \cite{6.3}. Addressing these challenges, a Trustworthy FL should maintain the following goals and objectives:
\begin{enumerate}
   
\item{Robust security with no single point of failure:} Protect the system even if a single component doesn't produce results or is attacked by adversaries.

\item{Local-to-global assurances: Provide a strong guarantee that high-quality models result from accurate aggregation of several local training models without manipulation.}

\item{Streamlined model verification and auditability:} Facilitate users to confirm a model update's accuracy and easily access verifiable data for specific update versions.

\item{Unalterable model update history:} Showcase a globally uniform record of global model updates without changes. Within this log, each update corresponds to a distinct entry that cannot be edited or deleted once created.

\item{Dependable client and model identification:} Select trustworthy clients and models to enhance the FL process.

\item{Reliable contribution evaluation and incentive allocation:} Implement a credible assessment of contributions and provide incentives to encourage FL clients' participation in future sessions. Investigators must concentrate on novel approaches to address the challenges in FL by creating equitable and dependable metrics for worker selection and evaluation.

\item{Dynamic authoritative keys management:} Allow updates for authoritative keys, even if some, but not most, keys are compromised.

\item{Timely Monitoring:} Timely monitoring methods for workers and model assessment schemes are necessary. Strengthening security and privacy features is crucial for Trustworthy FL.

\end{enumerate}

By emphasizing these objectives, crucial areas and fostering cooperation among stakeholders, the creation of secure and dependable FL systems can facilitate more responsible and ethical AI applications. Within the current research landscape, there is a noticeable absence of studies investigating the crucial pillars of trustworthiness, specifically in relation to FL. To the best of our knowledge, this is the pioneering work encompassing all facets of Trustworthy FL. In an effort to bridge the gaps in existing literature, this study presents the following contributions:
\begin{enumerate}

\item {We furnish an overview of the fundamental principles underlying trustworthy FL, accompanied by a comprehensive architecture of Trustworthy FL and trust assessment mechanisms.}

\item{We develop a taxonomy centered on the most pertinent trustworthiness pillars within FL contexts. In order to construct this taxonomy, we identify three primary pillars as the foundational elements: Interpretability, Fairness, and Security \& Privacy. Each pillar embodies a dimension of trust, which is further delineated into distinct concepts for each pillar. Additionally, we explore the key challenges and future directions in the realm of Trustworthy FL.}
\item{In conclusion, we highlighted the essential research challenges associated with each facet of Trustworthy FL, as well as the direction for future investigations.}
\end{enumerate}

The remainder of this paper is organized as follows: 
The structure of this paper is as follows: We commence with an introduction to FL and its classification in Section 2, succeeded by a discussion on trust in FL, fundamental principles of trustworthiness, and the architecture of Trustworthy FL systems in Section 3. Section 4 presents a literature review, emphasizing existing research in the domain. Sections 5-8 delve into trust evaluation, trust-aware interpretability, fairness-aware trustworthiness, and trust-aware security \& privacy-preserving FL systems, respectively. Throughout the paper, we examine topics such as data and feature selection, data sharing, model selection, explainability, client selection, contribution assessment, incentive mechanisms, accountability, auditability, secure and data aggregation, privacy preservation, and more. In conclusion, we explore open challenges and future research avenues in the field of Trustworthy FL, laying the foundation for the creation of secure and dependable FL systems that facilitate responsible and ethical AI applications.
\section{Fedrated Learning an Overview}
FL is a transformative approach to distributed machine learning, enabling the collaborative training of models across multiple devices or clients while preserving data privacy \cite{6.4}. Various FL architectures have been developed to address the diverse challenges and requirements in data distribution, communication patterns, and coordination methods. These architectures can be broadly classified into three main categories: data distribution-based architectures, such as Horizontal FL (HFL), Vertical FL (VFL), and Federated Transfer Learning (FTL); scale and communication-based architectures, including Cross-Silo FL (CSFL) and Cross-Device FL (CDFL); and coordination-based architectures, which encompass Centralized FL, Decentralized FL, and Hierarchical FL. Each of these architectures caters to specific needs and constraints, ensuring optimal model development across a wide range of applications while maintaining data privacy and security.

\subsection{Data Distribution-Based FL Architectures}
\subsubsection{Vertical FL} VFL involves federated training of datasets that share the same sample space but differ in feature spaces. It is apt for scenarios in which data is divided vertically based on feature dimensions, with different parties holding homogeneous data that partially overlaps in sample IDs. Entity alignment and encryption methods can be employed to address data sample overlapping during local training. One instance of VFL in the healthcare sector is the collaboration between hospitals and insurance companies, jointly training an AI model using their respective datasets. Although VFL provides data privacy and joint training benefits, it presents more implementation challenges than HFL due to the requirement for entity resolution and the limitations of current techniques for complex machine learning models.

\subsubsection{Horizontal FL} In HFL, multiple participants possessing datasets with identical feature spaces but varying sample spaces collaborate in training a joint global model. The datasets are used locally, and a server merges the local updates received from the participants to develop a global update without accessing the local data directly. One example of HFL in the healthcare domain is speech disorder detection, where users with different voices speak identical sentences, and the local updates are combined to create a comprehensive speech recognition model. HFL is primarily utilized in smart devices and IoT applications. It allows leveraging data from numerous sources without sharing raw data, thus maintaining privacy. However, HFL may encounter challenges when dealing with a limited number of labeled entities.

\begin{figure}[!ht]
\centering
\includegraphics[width=7cm,height=9cm,keepaspectratio]{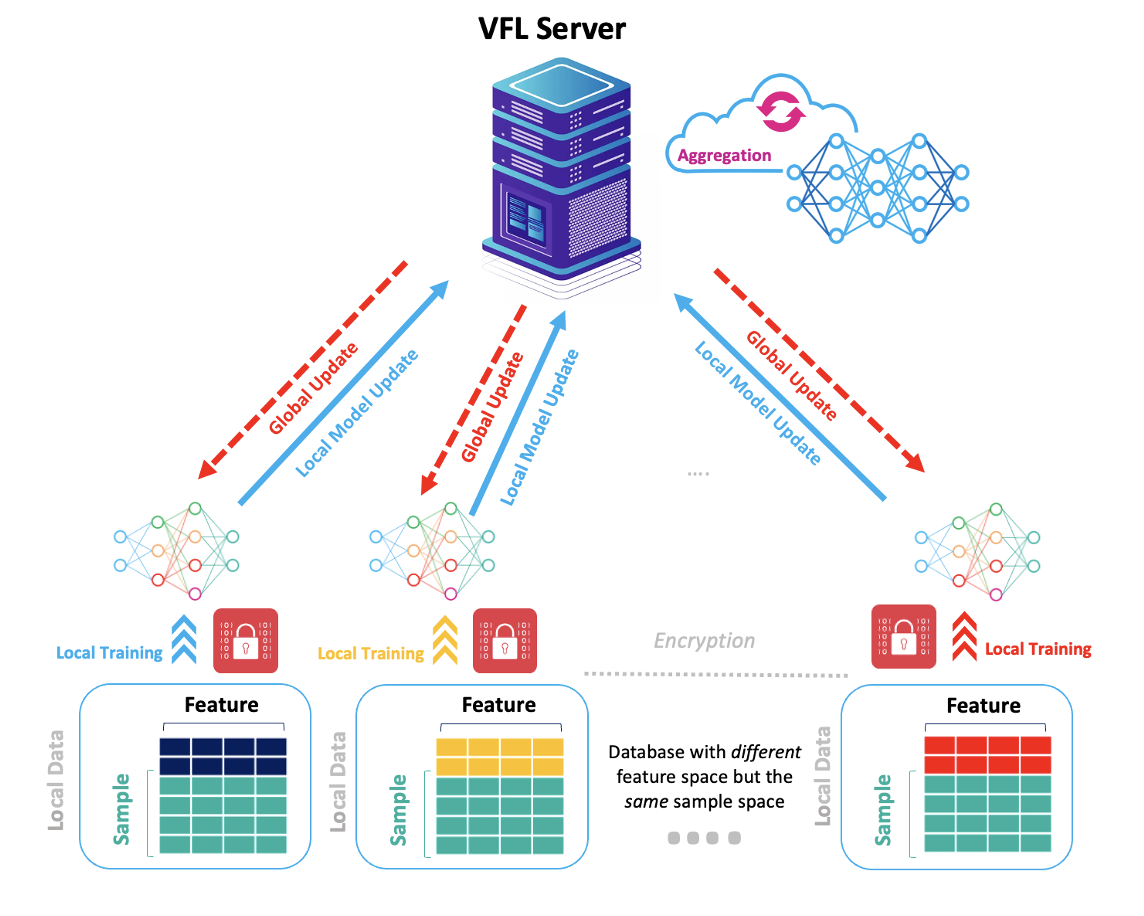}
\label{fog1}
\end{figure}
\begin{figure}[!ht]
\centering
\includegraphics[width=7cm,height=9cm,keepaspectratio]{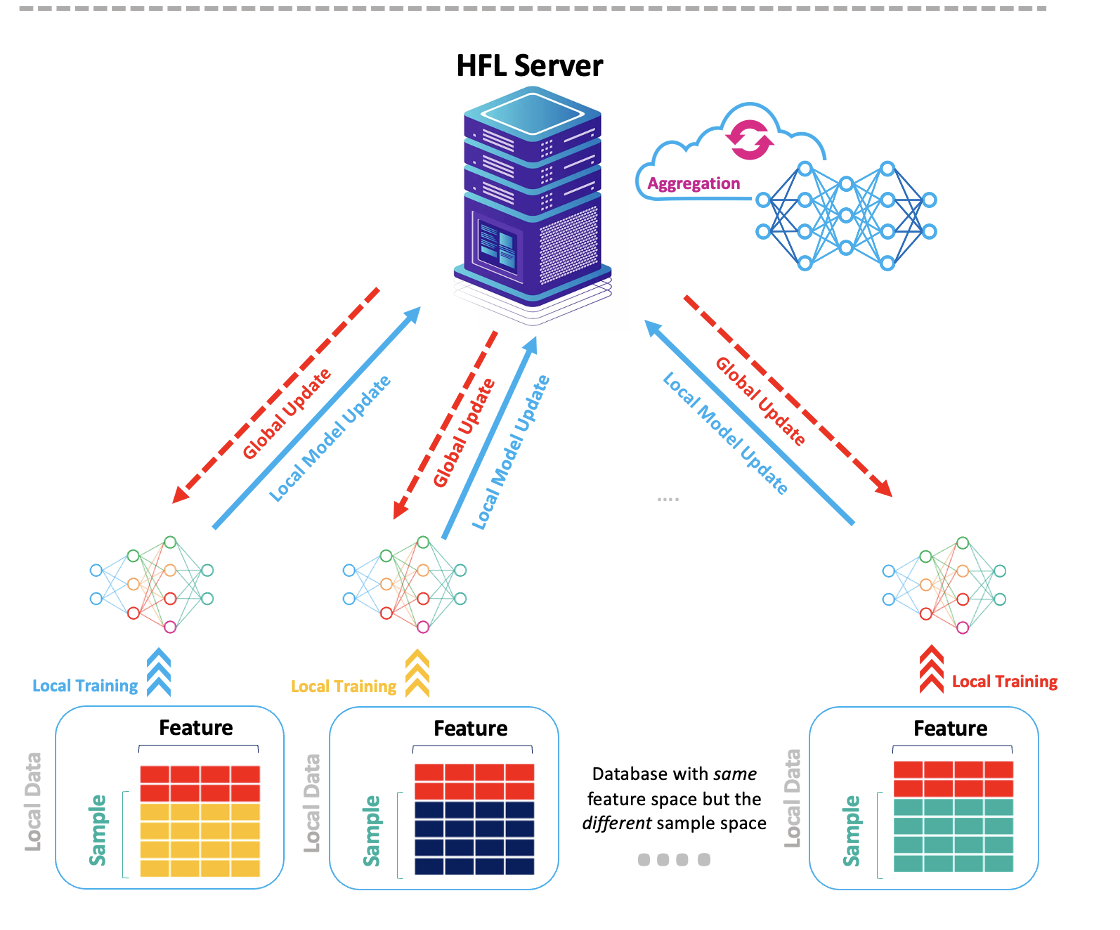}
\label{fog2}
\end{figure}
\begin{figure}[!ht]
\centering
\includegraphics[width=7cm,height=9cm,keepaspectratio]{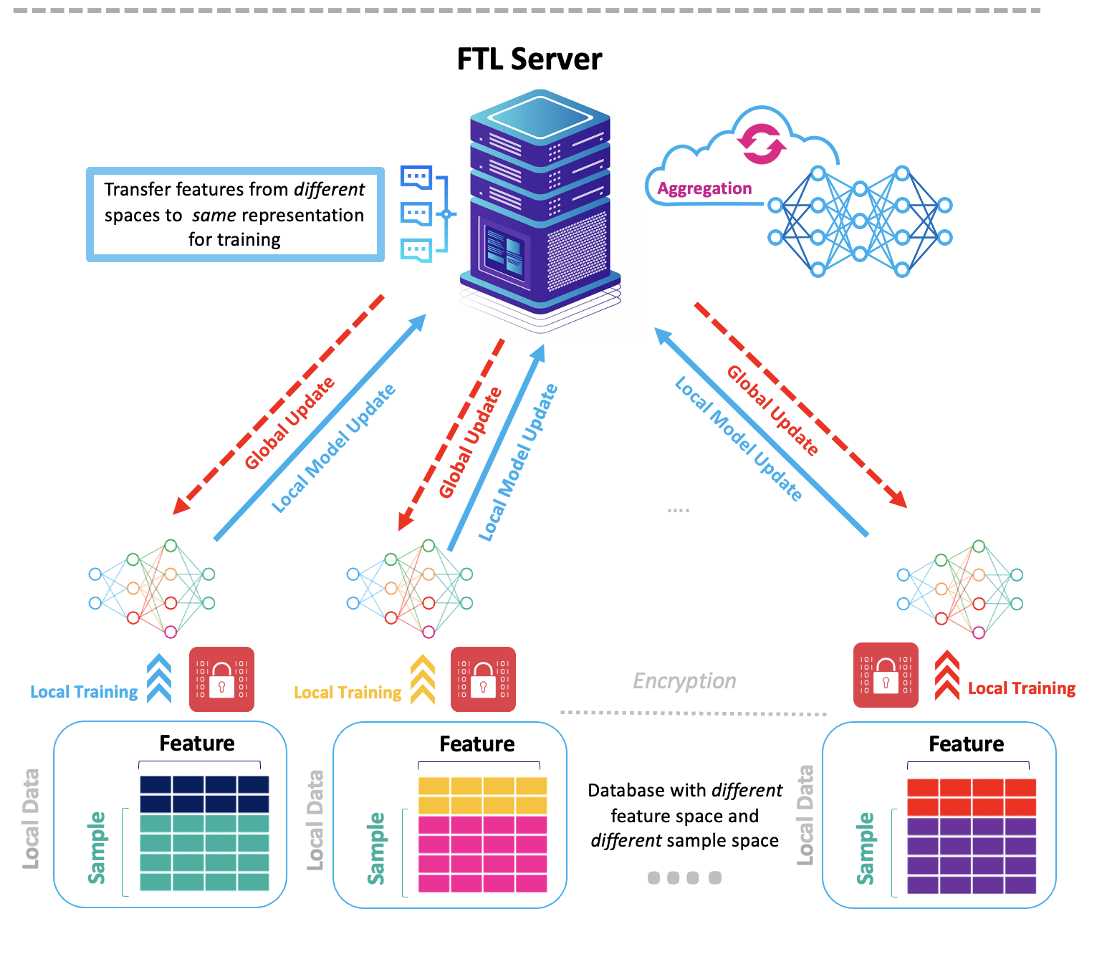}
\caption{A depiction of data distribution-based FL architectures.}
\label{fog3}
\end{figure}
\subsubsection{Federated Transfer Learning} FTL is tailored to handle datasets that differ in both sample spaces and feature spaces. Transfer learning techniques are used to compute feature values from distinct feature spaces into a uniform representation, which is then employed for local dataset training. Privacy protection mechanisms, such as random masks, can be implemented during the gradient exchange between clients and servers. FTL can support smart healthcare use cases like disease diagnosis by enabling collaboration among multiple hospitals with distinct patients and treatment plans. Although FTL has potential, it remains an evolving research area with challenges related to communication efficiency and flexibility in dealing with diverse data structures. Nevertheless, it provides a promising solution for ensuring data security and user privacy while addressing data island problems. An illustration of these data distribution-based FL architectures is presented in Fig. 1.

\begin{figure*}[!ht]
\centering
\includegraphics[width=18cm,height=15cm,keepaspectratio]{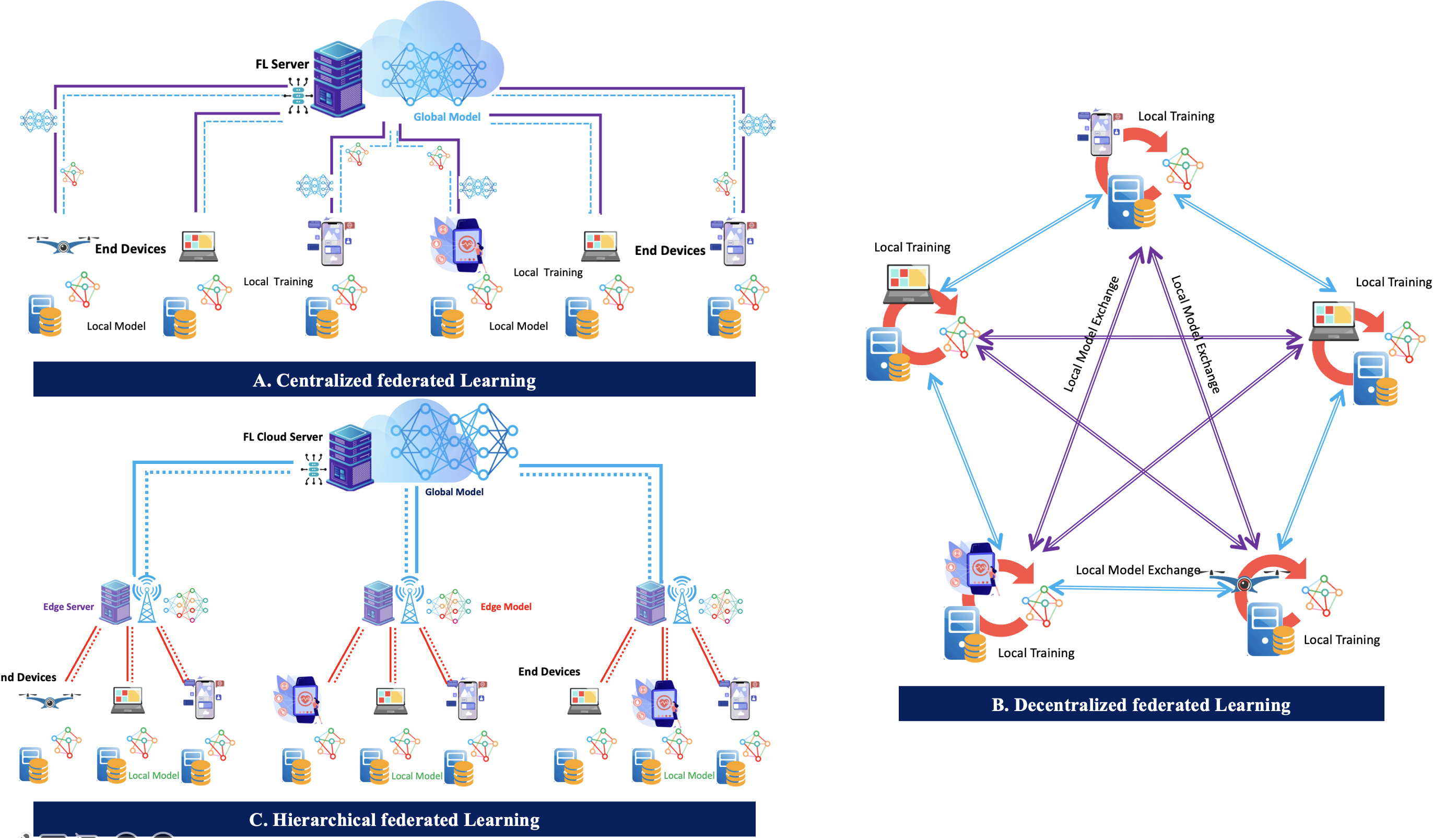}
\caption{An illustration of  coordination-based FL architectures.}
\label{fog4}
\end{figure*}
\subsection{Scale and Communication-Based FL Architectures}
\subsubsection{Cross-Silo FL}
CSFL is an architectural approach within the FL paradigm that focuses on collaboration between a limited number of clients, each possessing large-scale datasets. Typically, CSFL is employed in scenarios where organizations or data centers aim to jointly train a shared model while preserving data privacy and adhering to regulatory compliance. By leveraging a central server for aggregating model updates, clients can effectively contribute to the global model without directly sharing raw data.
\subsubsection{ Device-Silo FL}
DSFL, also known as Cross-Device FL, involves a substantial number of clients with relatively smaller-scale datasets, such as those found in smartphones or Internet of Things (IoT) devices. The primary objective of DSFL is to harness the collective knowledge of numerous devices to develop a global model while minimizing data movement and maintaining user privacy. This approach is particularly well-suited for edge computing environments, where devices have limited computational resources and intermittent network connectivity.

Both CSFL and DSFL represent distinct FL architectures, catering to different requirements in terms of scale, data distribution, and communication patterns. By tailoring the architecture to the specific challenges and constraints of a given application, these methods enable efficient and privacy-preserving collaborative learning.

\subsection{Coordination-Based FL Architectures}

\subsubsection{Centralized FL} This approach to FL involves the collaboration of multiple clients with a central server to train a global model. Clients independently train local models on their data and send updates to the central server. The server aggregates these updates, refines the global model, and distributes it back to clients. This iterative process maintains privacy by keeping data localized while benefiting from the collective knowledge of participating clients.

\subsubsection{Decentralized FL} In Decentralized FL, clients directly communicate with each other, eliminating the need for a central server. Clients exchange model updates in a peer-to-peer manner, allowing the system to be more resilient to failures and less reliant on a single point of control. This method offers enhanced privacy and scalability, making it suitable for large-scale networks with potentially unreliable or untrustworthy central authorities.

\subsubsection{Hierarchical FL} HFL introduces a multi-layer structure to the system, combining elements of centralized and decentralized approaches. Clients are organized into clusters or groups, with each group having a local aggregator. Clients send their local model updates to their respective aggregators, which then share aggregated updates with a central server or other aggregators. This hierarchical structure optimizes communication efficiency, enhances scalability, and provides a more flexible framework for diverse network topologies. An illustration of these three coordination-based FL architectures is presented in Fig. 2.
\section{Trustworthy Federated Learning an overview}
\subsection{What is trust in FL?}
In FL scenarios, trust refers to the assurance one node has in another's ability to execute specific tasks and serve as a reliable partner. Trust is subjective and depends on the domain and context, indicating the degree of control granted by one party to another. It is often portrayed as risk management, as it can be reduced but never fully eliminated. For example, when patients consent to data collection, they demonstrate trust in healthcare institutions. This data is valuable for ML algorithms, which strive to predict specific treatments or conditions. Trust involves a variety of factors that collectively ensure security, reliability, and privacy in different settings.

In IoT networks, the establishment and evaluation of trust allow end devices to form secure and efficient connections with other nodes or networks based on their trust values. The trustworthiness of devices within the network contributes to secure routing, facilitating stable data transmission paths and the selection of an efficient mobility model. Nodes in the network must assess each other's trustworthiness to maintain trusted communication. Trust is a psychological state reflecting a trustor's willingness to accept risks on behalf of a trustee without monitoring or external control. Trustworthiness is a prerequisite for choosing to trust someone and focuses on the trustee's attributes. Both AI and IoT fields consider trustworthiness crucial, but they approach it differently. AI technologies, such as machine learning systems, are designed to possess human-like traits. Rising public skepticism has driven governments and organizations to devise AI frameworks that outline principles for creating trustworthy AI. Key frameworks emphasize aspects like privacy, transparency, accountability, safety and security, and explainability, human control of technology, professional responsibility, fairness and non-discrimination, and the promotion of human values.

Trustworthiness is a quantitative expression of an entity's trust level, which changes dynamically based on the entity's actions. Factors like data security, authenticity, accuracy, service and processing capabilities of nodes, and the reliability and real-time performance of connections contribute to trustworthiness. Trust can be built through cryptographic techniques or distributed reputation systems. In the current era of information abundance, truth discovery (TD) methods have been applied to evaluate data quality in various real-world applications, such as mobile crowdsensing and knowledge bases. Truth discovery algorithms, like conflict resolution in heterogeneous data (CRH) and the Gaussian truth model (GTM), determine true and reliable values by aggregating data from multiple sources and assessing the reliability of those sources.
\begin{figure*}[!ht]
\centering
\includegraphics[width=17cm,height=14cm,keepaspectratio]{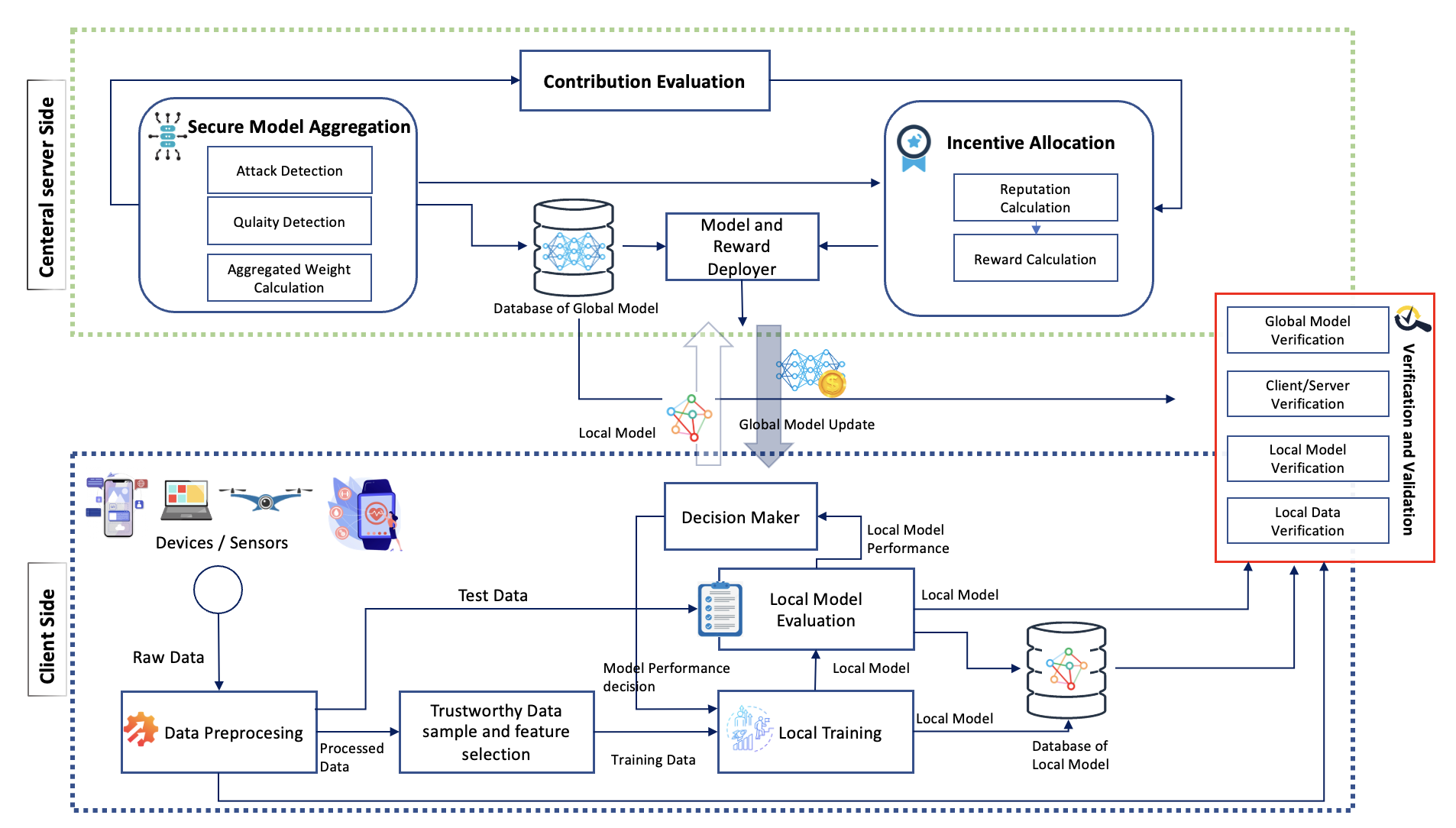}
\caption{A diagram illustrating the proposed comprehensive architecture for Trustworthy FL, highlighting the primary phases along with their respective components, which collectively ensure trustworthiness, data integrity, fairness, security, and privacy throughout the FL system.}
\label{fog5}
\end{figure*}
\subsection{Fundamentals of Trustworthiness in FL}
In this section, we will discuss a set of criteria for evaluating trust assessment methods based on machine learning and FL. These criteria pertain to the trustworthiness of FL systems, addressing questions such as: "Is the client device's local model non-adversarial and trustworthy?", "Has the local model genuinely been trained using the device's local data?", and "Can the client rely on the central server for the accuracy of the global model it provides?". The main accountability challenges in FL systems involve auditing the data used to train each local model, assessing the various local models provided by multiple client devices, and evaluating the global model derived from these individual models.

In this discussion, we will address a few crucial aspects of trust evaluation methods to ensure accuracy and effectiveness. A trustworthy evaluation method must provide precise trust values for trustees and offer reliable evaluation results. Various indicators, such as accuracy, precision, recall, and F-score, can represent effectiveness. When examining trust evaluation within machine learning, we must consider two critical components: the data used to train the model and the algorithm that builds the model. Selecting appropriate data and algorithms contributes to a highly accurate evaluation, so it's essential to explore the impact of these choices on trust evaluation. During the trust evaluation process, attacks such as conflict behavior, on-off, collision, Sybil whitewashing, and bad-mouthing may occur. Trust evaluation methods should aim to prevent these attacks to maintain robustness and ensure evaluation results remain unaffected. It's essential to protect users' private information when handling data for trust evaluation. Trust evaluation methods must prioritize privacy protection for both users and trust evidence to gain acceptance and recognition.

Context-awareness and subjectivity are fundamental characteristics of trust, so trust evaluation methods should support these attributes. By incorporating context-awareness, the evaluation scheme can adapt to changes in the application scenario, context, or environment. Ensuring subjectivity in trust evaluation brings the expression of trust closer to reality. Standardizing and operationalizing trustworthiness concerns in machine learning systems is crucial for incorporation into specifications and objective evaluation in applications.
In FL, client selection is crucial for trustworthiness. Many selection methods emphasize server goals, such as faster convergence or better model accuracy, potentially disadvantaging clients. Common threshold-based approaches can lead to unfairness, typically categorized as over-representation, under-representation, and non-representation. In systems sensitive to network speeds, clients with slower connections may be under-represented or excluded entirely, while those with faster connections are over-represented. This imbalance can cause global model bias, impacting overall performance negatively.
Achieving fairness in client selection for FL involves balancing the interests of both the FL server and clients, while considering client heterogeneity. Biases introduced during the FL model optimization process can lead to unequal performance across clients, which is seen as unfair treatment. Recent research explores fairness issues in the FL model optimization process, focusing on two main approaches: objective function-based and gradient-based methods. These strategies aim to mitigate biases and performance discrepancies during model training.
Contribution evaluation in FL is crucial for trustworthiness, as it measures each client's influence on the global model without exposing local data. This evaluation is vital for equitable client selection and incentive allocation. Traditional data valuation methods aren't directly applicable to FL. Current FL contribution evaluation strategies consist of self-reported data, utility game, individual evaluation, empirical methods, and Shapley value. Reputation systems track past contributions to determine reliability, assisting in client selection and rewards distribution, fostering trustworthiness in both centralized and decentralized FL systems.
Improving explainability in the context of trustworthiness is valuable for addressing biases in machine learning. In FL, enhancing explainability can potentially promote fairness. FL clients often lack mechanisms to determine if they are treated fairly, and this uncertainty may negatively affect their future decisions to participate in FL. The objective of developing explainability is to offer a comprehensive understanding of the FL server's decision-making process and its impact on each client's interests, ultimately fostering trust between the parties. However, it is essential to conduct explainability research within the framework of privacy preservation in FL to avoid conflicting with its primary goal.
\subsection{Architecture of Trustworthy FL}
In this section, we present a comprehensive architecture for Trustworthy FL that encompasses all aspects of trustworthiness within the FL process. Our proposed architecture comprises three main phases for achieving Trustworthy FL: Trust-aware interpretability phase covers all aspects related to data selection, data quality, feature selection, and trustworthy model selection at the local node. After carefully evaluating the local model performance and data verifiability, we proceed to the FL server-side phase. Upon receiving local data and requirements from clients, the server aggregates the models after thoroughly assessing their quality and trust level. To promote trustworthy client participation, a contribution evaluation method and incentive mechanism are implemented at the server side having fairness aspect in Trustworthy FL. Multiple parameters and strategies are used to compute incentives based on reputation and other factors, encouraging clients to participate in subsequent federated rounds. The verification module within the architecture is responsible for validating local models, local data, client and server interactions, and global model updates. The secure aggregation and verification module ensures privacy preservation in our proposed Trustworthy FL architecture.
The proposed architecture is visually represented in Fig. 3, illustrating the interaction of these components and their contribution to achieving a Trustworthy FL system.
\begin{table*}[]
\caption{Comparative Analysis of related Works with our Proposed Work}
\label{tab:my-table1}
\resizebox{\textwidth}{!}{%
\begin{tabular} {|ccccc|cccc|cccc|ccc|}
\hline
\multicolumn{1}{|c|}{\textbf{Related Work}} &
  \multicolumn{1}{c|}{\textbf{Year}} &
  \multicolumn{1}{c|}{\textbf{FL}} &
  \multicolumn{1}{c|}{\textbf{ML}} &
  \textbf{Trust} &
  \multicolumn{1}{c|}{\begin{tabular}[c]{@{}c@{}}F\&DS\end{tabular}} &
  \multicolumn{1}{c|}{\begin{tabular}[c]{@{}c@{}}DS\end{tabular}} &
  \multicolumn{1}{c|}{\begin{tabular}[c]{@{}c@{}}MS\end{tabular}} &
  Explainability &
  \multicolumn{1}{c|}{Fairness} &
  \multicolumn{1}{c|}{\begin{tabular}[c]{@{}c@{}}CS\end{tabular}} &
  \multicolumn{1}{c|}{\begin{tabular}[c]{@{}c@{}}IM\end{tabular}} &
  \begin{tabular}[c]{@{}c@{}}CE\end{tabular} &
  \multicolumn{1}{c|}{\begin{tabular}[c]{@{}c@{}}Verifiability\\ and Aud.\end{tabular}} &
  \multicolumn{1}{c|}{\begin{tabular}[c]{@{}c@{}}Secure \\ Agg\end{tabular}} &
  \begin{tabular}[c]{@{}c@{}}Privacy\\ and \\ Security\end{tabular} \\ \hline
\multicolumn{5}{|c|}{} &
  \multicolumn{4}{c|}{Interpretability } &
  \multicolumn{4}{c|}{\begin{tabular}[c]{@{}c@{}}Fairness \\ Aware\end{tabular}} &
  \multicolumn{3}{c|}{Security and privacy} \\ \hline
\multicolumn{1}{|c|}{\cite{rwml1}} &
  \multicolumn{1}{c|}{2020} &
  \multicolumn{1}{c|}{\begin{large}\textcolor{lightgray}{\xmark}\end{large}} &
  \multicolumn{1}{c|}{\quad\circledmark[cyan]} &
  \quad\circledmark[cyan] &
  \multicolumn{1}{c|}{\begin{large}\textcolor{lightgray}{\xmark}\end{large}} &
  \multicolumn{1}{c|}{\begin{large}\textcolor{lightgray}{\xmark}\end{large}} &
  \multicolumn{1}{c|}{\begin{large}\textcolor{lightgray}{\xmark}\end{large}} &
  \quad\circledmark[cyan] &
  \multicolumn{1}{c|}{\quad\circledmark[cyan]} &
  \multicolumn{1}{c|}{\begin{large}\textcolor{lightgray}{\xmark}\end{large}} &
  \multicolumn{1}{c|}{\begin{large}\textcolor{lightgray}{\xmark}\end{large}} &
  \begin{large}\textcolor{lightgray}{\xmark}\end{large} &
  \multicolumn{1}{c|}{\quad\circledmark[cyan]} &
  \multicolumn{1}{c|}{\begin{large}\textcolor{lightgray}{\xmark}\end{large}} &
  \begin{large}\textcolor{cyan}{\xmark}\end{large} \\ \hline
\multicolumn{1}{|c|}{\cite{rwml2}} &
  \multicolumn{1}{c|}{2022} &
  \multicolumn{1}{c|}{\begin{large}\textcolor{lightgray}{\xmark}\end{large}} &
  \multicolumn{1}{c|}{\quad\circledmark[cyan]} &
  \quad\circledmark[cyan] &
  \multicolumn{1}{c|}{\begin{large}\textcolor{lightgray}{\xmark}\end{large}} &
  \multicolumn{1}{c|}{\begin{large}\textcolor{lightgray}{\xmark}\end{large}} &
  \multicolumn{1}{c|}{\begin{large}\textcolor{lightgray}{\xmark}\end{large}} &
  \quad\circledmark[cyan] &
  \multicolumn{1}{c|}{\quad\circledmark[cyan]} &
  \multicolumn{1}{c|}{\begin{large}\textcolor{lightgray}{\xmark}\end{large}} &
  \multicolumn{1}{c|}{\begin{large}\textcolor{lightgray}{\xmark}\end{large}} &
  \quad\circledmark[cyan] &
  \multicolumn{1}{c|}{\begin{large}\textcolor{lightgray}{\xmark}\end{large}} &
  \multicolumn{1}{c|}{\begin{large}\textcolor{lightgray}{\xmark}\end{large}} &
  \quad\circledmark[cyan] \\ \hline
\multicolumn{1}{|c|}{\cite{rwml3}} &
  \multicolumn{1}{c|}{2023} &
  \multicolumn{1}{c|}{\begin{large}\textcolor{lightgray}{\xmark}\end{large}} &
  \multicolumn{1}{c|}{\quad\circledmark[cyan]} &
  \begin{large}\textcolor{lightgray}{\xmark}\end{large} &
  \multicolumn{1}{c|}{\begin{large}\textcolor{lightgray}{\xmark}\end{large}} &
  \multicolumn{1}{c|}{\begin{large}\textcolor{lightgray}{\xmark}\end{large}} &
  \multicolumn{1}{c|}{\begin{large}\textcolor{lightgray}{\xmark}\end{large}} &
  \quad\circledmark[cyan] &
  \multicolumn{1}{c|}{\quad\circledmark[cyan]} &
  \multicolumn{1}{c|}{\begin{large}\textcolor{lightgray}{\xmark}\end{large}} &
  \multicolumn{1}{c|}{\begin{large}\textcolor{lightgray}{\xmark}\end{large}} &
  \begin{large}\textcolor{lightgray}{\xmark}\end{large} &
  \multicolumn{1}{c|}{\quad\circledmark[cyan]} &
  \multicolumn{1}{c|}{\begin{large}\textcolor{lightgray}{\xmark}\end{large}} &
  \quad\circledmark[cyan] \\ \hline
\multicolumn{1}{|c|}{\cite{rwip1}} &
  \multicolumn{1}{c|}{2023} &
  \multicolumn{1}{c|}{\quad\circledmark[cyan]} &
  \multicolumn{1}{c|}{\begin{large}\textcolor{lightgray}{\xmark}\end{large}} &
  \begin{large}\textcolor{lightgray}{\xmark}\end{large} &
  \multicolumn{1}{c|}{\quad\circledmark[cyan]} &
  \multicolumn{1}{c|}{\begin{large}\textcolor{lightgray}{\xmark}\end{large}} &
  \multicolumn{1}{c|}{\quad\circledmark[cyan]} &
  \begin{large}\textcolor{lightgray}{\xmark}\end{large} &
  \multicolumn{1}{c|}{\begin{large}\textcolor{lightgray}{\xmark}\end{large}} &
  \multicolumn{1}{c|}{\quad\circledmark[cyan]} &
  \multicolumn{1}{c|}{\begin{large}\textcolor{lightgray}{\xmark}\end{large}} &
  \begin{large}\textcolor{lightgray}{\xmark}\end{large} &
  \multicolumn{1}{c|}{\begin{large}\textcolor{lightgray}{\xmark}\end{large}} &
  \multicolumn{1}{c|}{\begin{large}\textcolor{lightgray}{\xmark}\end{large}} &
  \begin{large}\textcolor{cyan}{\xmark}\end{large} \\ \hline
\multicolumn{1}{|c|}{\begin{tabular}[c]{@{}c@{}}\cite{rwim1}\\ \cite{rwim2}\\ \cite{rwim3}\\ \cite{rwim4}\end{tabular}} &
  \multicolumn{1}{c|}{\begin{tabular}[c]{@{}c@{}}2021\\ 2021\\ 2022\\ 2021\end{tabular}} &
  \multicolumn{1}{c|}{\quad\circledmark[cyan]} &
  \multicolumn{1}{c|}{\begin{large}\textcolor{lightgray}{\xmark}\end{large}} &
  \begin{large}\textcolor{lightgray}{\xmark}\end{large} &
  \multicolumn{1}{c|}{\begin{large}\textcolor{lightgray}{\xmark}\end{large}} &
  \multicolumn{1}{c|}{\begin{large}\textcolor{lightgray}{\xmark}\end{large}} &
  \multicolumn{1}{c|}{\begin{large}\textcolor{lightgray}{\xmark}\end{large}} &
  \begin{large}\textcolor{lightgray}{\xmark}\end{large} &
  \multicolumn{1}{c|}{\begin{large}\textcolor{lightgray}{\xmark}\end{large}} &
  \multicolumn{1}{c|}{\begin{large}\textcolor{lightgray}{\xmark}\end{large}} &
  \multicolumn{1}{c|}{\quad\circledmark[cyan]} &
  \begin{large}\textcolor{lightgray}{\xmark}\end{large} &
  \multicolumn{1}{c|}{\begin{large}\textcolor{lightgray}{\xmark}\end{large}} &
  \multicolumn{1}{c|}{\begin{large}\textcolor{lightgray}{\xmark}\end{large}} &
  \begin{large}\textcolor{lightgray}{\xmark}\end{large} \\ \hline
\multicolumn{1}{|c|}{\cite{rwim5}} &
  \multicolumn{1}{c|}{2022} &
  \multicolumn{1}{c|}{\quad\circledmark[cyan]} &
  \multicolumn{1}{c|}{\begin{large}\textcolor{lightgray}{\xmark}\end{large}} &
  \begin{large}\textcolor{lightgray}{\xmark}\end{large} &
  \multicolumn{1}{c|}{\begin{large}\textcolor{lightgray}{\xmark}\end{large}} &
  \multicolumn{1}{c|}{\begin{large}\textcolor{lightgray}{\xmark}\end{large}} &
  \multicolumn{1}{c|}{\begin{large}\textcolor{lightgray}{\xmark}\end{large}} &
  \begin{large}\textcolor{lightgray}{\xmark}\end{large} &
  \multicolumn{1}{c|}{\begin{large}\textcolor{lightgray}{\xmark}\end{large}} &
  \multicolumn{1}{c|}{\begin{large}\textcolor{lightgray}{\xmark}\end{large}} &
  \multicolumn{1}{c|}{\quad\circledmark[cyan]} &
  \begin{large}\textcolor{lightgray}{\xmark}\end{large} &
  \multicolumn{1}{c|}{\begin{large}\textcolor{lightgray}{\xmark}\end{large}} &
  \multicolumn{1}{c|}{\begin{large}\textcolor{lightgray}{\xmark}\end{large}} &
  \begin{large}\textcolor{lightgray}{\xmark}\end{large} \\ \hline
\multicolumn{1}{|c|}{\cite{rwce1}} &
  \multicolumn{1}{c|}{2020} &
  \multicolumn{1}{c|}{\quad\circledmark[cyan]} &
  \multicolumn{1}{c|}{\begin{large}\textcolor{lightgray}{\xmark}\end{large}} &
  \begin{large}\textcolor{lightgray}{\xmark}\end{large} &
  \multicolumn{1}{c|}{\begin{large}\textcolor{lightgray}{\xmark}\end{large}} &
  \multicolumn{1}{c|}{\begin{large}\textcolor{lightgray}{\xmark}\end{large}} &
  \multicolumn{1}{c|}{\begin{large}\textcolor{lightgray}{\xmark}\end{large}} &
  \begin{large}\textcolor{lightgray}{\xmark}\end{large} &
  \multicolumn{1}{c|}{\quad\circledmark[cyan]} &
  \multicolumn{1}{c|}{\begin{large}\textcolor{lightgray}{\xmark}\end{large}} &
  \multicolumn{1}{c|}{\begin{large}\textcolor{lightgray}{\xmark}\end{large}} &
  \quad\circledmark[cyan] &
  \multicolumn{1}{c|}{\begin{large}\textcolor{lightgray}{\xmark}\end{large}} &
  \multicolumn{1}{c|}{\begin{large}\textcolor{lightgray}{\xmark}\end{large}} &
  \quad\circledmark[cyan] \\ \hline
\multicolumn{1}{|c|}{\cite{rwf1}} &
  \multicolumn{1}{c|}{2021} &
  \multicolumn{1}{c|}{\quad\circledmark[cyan]} &
  \multicolumn{1}{c|}{\begin{large}\textcolor{lightgray}{\xmark}\end{large}} &
  \begin{large}\textcolor{lightgray}{\xmark}\end{large} &
  \multicolumn{1}{c|}{\begin{large}\textcolor{lightgray}{\xmark}\end{large}} &
  \multicolumn{1}{c|}{\begin{large}\textcolor{lightgray}{\xmark}\end{large}} &
  \multicolumn{1}{c|}{\begin{large}\textcolor{lightgray}{\xmark}\end{large}} &
  \begin{large}\textcolor{lightgray}{\xmark}\end{large} &
  \multicolumn{1}{c|}{\quad\circledmark[cyan]} &
  \multicolumn{1}{c|}{\quad\circledmark[cyan]} &
  \multicolumn{1}{c|}{\quad\circledmark[cyan]} &
  \quad\circledmark[cyan] &
  \multicolumn{1}{c|}{\begin{large}\textcolor{lightgray}{\xmark}\end{large}} &
  \multicolumn{1}{c|}{\begin{large}\textcolor{lightgray}{\xmark}\end{large}} &
  \begin{large}\textcolor{lightgray}{\xmark}\end{large} \\ \hline
\multicolumn{1}{|c|}{\begin{tabular}[c]{@{}c@{}}\cite{rwpr1}\\ \cite{rwpr2}\end{tabular}} &
  \multicolumn{1}{c|}{\begin{tabular}[c]{@{}c@{}}2021\\ 2021\end{tabular}} &
  \multicolumn{1}{c|}{\quad\circledmark[cyan]} &
  \multicolumn{1}{c|}{\begin{large}\textcolor{lightgray}{\xmark}\end{large}} &
  \begin{large}\textcolor{lightgray}{\xmark}\end{large} &
  \multicolumn{1}{c|}{\begin{large}\textcolor{lightgray}{\xmark}\end{large}} &
  \multicolumn{1}{c|}{\begin{large}\textcolor{lightgray}{\xmark}\end{large}} &
  \multicolumn{1}{c|}{\begin{large}\textcolor{lightgray}{\xmark}\end{large}} &
  \begin{large}\textcolor{lightgray}{\xmark}\end{large} &
  \multicolumn{1}{c|}{\begin{large}\textcolor{lightgray}{\xmark}\end{large}} &
  \multicolumn{1}{c|}{\begin{large}\textcolor{lightgray}{\xmark}\end{large}} &
  \multicolumn{1}{c|}{\begin{large}\textcolor{lightgray}{\xmark}\end{large}} &
  \begin{large}\textcolor{lightgray}{\xmark}\end{large} &
  \multicolumn{1}{c|}{\begin{large}\textcolor{lightgray}{\xmark}\end{large}} &
  \multicolumn{1}{c|}{\begin{large}\textcolor{lightgray}{\xmark}\end{large}} &
  \begin{large}\textcolor{lightgray}{\xmark}\end{large} \\ \hline
\multicolumn{1}{|c|}{\cite{rwpr3}} &
  \multicolumn{1}{c|}{2022} &
  \multicolumn{1}{c|}{\quad\circledmark[cyan]} &
  \multicolumn{1}{c|}{\begin{large}\textcolor{lightgray}{\xmark}\end{large}} &
  \begin{large}\textcolor{lightgray}{\xmark}\end{large} &
  \multicolumn{1}{c|}{\begin{large}\textcolor{lightgray}{\xmark}\end{large}} &
  \multicolumn{1}{c|}{\begin{large}\textcolor{lightgray}{\xmark}\end{large}} &
  \multicolumn{1}{c|}{\begin{large}\textcolor{lightgray}{\xmark}\end{large}} &
  \begin{large}\textcolor{lightgray}{\xmark}\end{large} &
  \multicolumn{1}{c|}{\begin{large}\textcolor{lightgray}{\xmark}\end{large}} &
  \multicolumn{1}{c|}{\begin{large}\textcolor{lightgray}{\xmark}\end{large}} &
  \multicolumn{1}{c|}{\begin{large}\textcolor{lightgray}{\xmark}\end{large}} &
  \begin{large}\textcolor{lightgray}{\xmark}\end{large} &
  \multicolumn{1}{c|}{\begin{large}\textcolor{lightgray}{\xmark}\end{large}} &
  \multicolumn{1}{c|}{\begin{large}\textcolor{lightgray}{\xmark}\end{large}} &
  \quad\circledmark[cyan] \\ \hline
\multicolumn{1}{|c|}{\cite{rwpr4}} &
  \multicolumn{1}{c|}{2022} &
  \multicolumn{1}{c|}{\quad\circledmark[cyan]} &
  \multicolumn{1}{c|}{\begin{large}\textcolor{lightgray}{\xmark}\end{large}} &
  \begin{large}\textcolor{lightgray}{\xmark}\end{large} &
  \multicolumn{1}{c|}{\begin{large}\textcolor{lightgray}{\xmark}\end{large}} &
  \multicolumn{1}{c|}{\begin{large}\textcolor{lightgray}{\xmark}\end{large}} &
  \multicolumn{1}{c|}{\begin{large}\textcolor{lightgray}{\xmark}\end{large}} &
  \begin{large}\textcolor{lightgray}{\xmark}\end{large} &
  \multicolumn{1}{c|}{\begin{large}\textcolor{lightgray}{\xmark}\end{large}} &
  \multicolumn{1}{c|}{\begin{large}\textcolor{lightgray}{\xmark}\end{large}} &
  \multicolumn{1}{c|}{\begin{large}\textcolor{lightgray}{\xmark}\end{large}} &
  \begin{large}\textcolor{lightgray}{\xmark}\end{large} &
  \multicolumn{1}{c|}{\begin{large}\textcolor{lightgray}{\xmark}\end{large}} &
  \multicolumn{1}{c|}{\begin{large}\textcolor{lightgray}{\xmark}\end{large}} &
  \quad\circledmark[cyan] \\ \hline
\multicolumn{1}{|c|}{\cite{rwms1}} &
  \multicolumn{1}{c|}{2020} &
  \multicolumn{1}{c|}{\quad\circledmark[cyan]} &
  \multicolumn{1}{c|}{\begin{large}\textcolor{lightgray}{\xmark}\end{large}} &
  \begin{large}\textcolor{lightgray}{\xmark}\end{large} &
  \multicolumn{1}{c|}{\begin{large}\textcolor{lightgray}{\xmark}\end{large}} &
  \multicolumn{1}{c|}{\begin{large}\textcolor{lightgray}{\xmark}\end{large}} &
  \multicolumn{1}{c|}{\begin{large}\textcolor{lightgray}{\xmark}\end{large}} &
  \begin{large}\textcolor{lightgray}{\xmark}\end{large} &
  \multicolumn{1}{c|}{\begin{large}\textcolor{lightgray}{\xmark}\end{large}} &
  \multicolumn{1}{c|}{\begin{large}\textcolor{lightgray}{\xmark}\end{large}} &
  \multicolumn{1}{c|}{\begin{large}\textcolor{lightgray}{\xmark}\end{large}} &
  \begin{large}\textcolor{lightgray}{\xmark}\end{large} &
  \multicolumn{1}{c|}{\begin{large}\textcolor{lightgray}{\xmark}\end{large}} &
  \multicolumn{1}{c|}{\quad\circledmark[cyan]} &
  \quad\circledmark[cyan] \\ \hline
\multicolumn{1}{|c|}{\cite{rwvf1}} &
  \multicolumn{1}{c|}{2022} &
  \multicolumn{1}{c|}{\quad\circledmark[cyan]} &
  \multicolumn{1}{c|}{\begin{large}\textcolor{lightgray}{\xmark}\end{large}} &
  \begin{large}\textcolor{lightgray}{\xmark}\end{large} &
  \multicolumn{1}{c|}{\begin{large}\textcolor{lightgray}{\xmark}\end{large}} &
  \multicolumn{1}{c|}{\begin{large}\textcolor{lightgray}{\xmark}\end{large}} &
  \multicolumn{1}{c|}{\quad\circledmark[cyan]} &
  \begin{large}\textcolor{lightgray}{\xmark}\end{large} &
  \multicolumn{1}{c|}{\begin{large}\textcolor{lightgray}{\xmark}\end{large}} &
  \multicolumn{1}{c|}{\begin{large}\textcolor{lightgray}{\xmark}\end{large}} &
  \multicolumn{1}{c|}{\begin{large}\textcolor{lightgray}{\xmark}\end{large}} &
  \begin{large}\textcolor{lightgray}{\xmark}\end{large} &
  \multicolumn{1}{c|}{\quad\circledmark[cyan]} &
  \multicolumn{1}{c|}{\quad\circledmark[cyan]} &
  \quad\circledmark[cyan] \\ \hline
\multicolumn{5}{|c|}{} &
  \multicolumn{1}{c|}{\begin{tabular}[c]{@{}c@{}}TW\\F\&DS\end{tabular}} &
  \multicolumn{1}{c|}{\begin{tabular}[c]{@{}c@{}}TW\\DS\end{tabular}} &
  \multicolumn{1}{c|}{\begin{tabular}[c]{@{}c@{}}TW\\MS\end{tabular}} &
  TW Explainability &
  \multicolumn{1}{c|}{TW Fairness} &
  \multicolumn{1}{c|}{\begin{tabular}[c]{@{}c@{}}TW\\CS\end{tabular}} &
  \multicolumn{1}{c|}{\begin{tabular}[c]{@{}c@{}}TW\\IM\end{tabular}} &
  \begin{tabular}[c]{@{}c@{}}TW\\CE\end{tabular} &
  \multicolumn{1}{c|}{\begin{tabular}[c]{@{}c@{}}TW\\Verifiability\\ and Aud.\end{tabular}} &
  \multicolumn{1}{c|}{\begin{tabular}[c]{@{}c@{}}TW\\Secure \\ Aggregation\end{tabular}} &
  \begin{tabular}[c]{@{}c@{}}TW Privacy\\ and \\ Security\end{tabular} \\ \hline
\multicolumn{2}{|c|}{\begin{tabular}[c]{@{}c@{}}Proposed \\ Trustworthy FL\end{tabular}} &
  \multicolumn{1}{c|}{\circledmark} &
  \multicolumn{1}{c|}{\begin{large}\textcolor{lightgray}{\xmark}\end{large}} &
  \circledmark &
  \multicolumn{1}{c|}{\circledmark} &
  \multicolumn{1}{c|}\circledmark &
  \multicolumn{1}{c|}\circledmark &
  \circledmark &
  \multicolumn{1}{c|}\circledmark &
  \multicolumn{1}{c|}\circledmark &
  \multicolumn{1}{c|}\circledmark &
 \circledmark &
  \multicolumn{1}{c|}\circledmark &
   \multicolumn{1}{c|}\circledmark &
  \circledmark \\ \hline \\
\multicolumn{16}{|l|}{*TW: Trustworthy, F\&DS: Feature and Data Selection, DS: Data Sharing, MS: Model Selection, CE: Contribution Evaluation, IM: Incentive Mechanism, CS: Client Selection} \\ \hline
\end{tabular}%
}

\end{table*}
\section{Literature review}
In this comparative analysis, we will discuss the focus of each paper concerning trustworthy aspects, including data selection, client selection, model selection, model aggregation, data aggregation, incentive mechanism, contribution evaluation, and privacy preservation.\\
The authors in \cite{rwml1} mainly discuss the relationship between trust in AI and the development of trustworthy machine learning technologies. This paper does not specifically address the trustworthy aspects mentioned above but provides a broader understanding of trust in AI systems. The research work in \cite{rwml2} present a survey on explainable, trustworthy, and ethical machine learning for healthcare. The paper focuses on ethical concerns and trustworthiness in healthcare applications, touching upon privacy preservation and model/data selection. moreover, \cite{rwml3} discuss trustworthy AI from principles to practices. The paper covers various aspects of trustworthiness, including privacy preservation, fairness-awareness, and interpretability but does not specifically focus on FL. Authors in \cite{rwip1} propose an interpretable FL approach. The paper emphasizes model selection and model aggregation, as well as privacy preservation, by providing interpretable and explainable models.\\
The articles \cite{rwim1, rwim2} both offer surveys on incentive mechanisms for FL. They focus on incentive mechanisms and contribution evaluation, discussing various economic and game-theoretic models to ensure effective and fair participation. Economic and game theoretic oriented FL incentive mechanisms have have been explored in \cite{rwim3}. This paper mainly targets the incentive mechanism and contribution evaluation. Authors in \cite{rwim4} presented a systematic review on incentive-driven FL and associated security challenges. The paper discusses incentive mechanisms, privacy preservation, and the challenges that arise from implementing incentives. however, \cite{rwim5} present a systemic survey on blockchain for FL, focusing on secure distributed Machine Learning (ML) systems. The paper addresses privacy preservation, data aggregation, and the use of blockchain for enhancing trust.
The authors in \cite{rwce1} investigate users' contributions and the factors influencing user participation and performance. Fairness-aware FL is examined in \cite{rwf1}, where client selection, incentive mechanisms, and contribution evaluations are explored with a taxonomy, but the paper does not focus on all aspects of trustworthiness in FL. Research works in \cite{rwpr1, rwpr2, rwpr3, rwpr4} discuss the security and privacy aspects in FL, exploring cryptographic techniques, secure multi-party computation, differential privacy, secure data aggregation, trust management, and secure model aggregation. These papers are not focusing on trust factor. Authors in\cite{rwms1}  conduct a systematic literature model quality perspective in FL. The paper examines various aspects of model quality, including model selection, but does not specifically address the other trustworthy aspects. Verifiablity in FL is discussed in \cite{rwvf1}, emphasizing model aggregation and contribution evaluation by introducing verification mechanisms. \\
In the realm of Trustworthy FL, existing research studies primarily focus on individual aspects and domains, such as Fairness-aware FL, Contribution Evaluation, Verifiability, Secure Aggregation, Model Selection, and Privacy Preservation. These studies, while valuable, do not holistically address the overarching trustworthiness of FL, leaving gaps in our understanding. Our research paper aims to bridge this gap by providing a comprehensive exploration of all facets of Trustworthy FL. With a strong emphasis on both client and server-side considerations, our work delves into crucial topics ranging from client and model selection to secure aggregation and privacy preservation. Moreover, we contribute significantly to the understanding of contribution evaluation and the development of incentive mechanisms that promote equitable reward distribution and verifiability. By synthesizing and expanding upon these diverse aspects, our research paper stands as a more complete and integrative approach to Trustworthy FL, ultimately enhancing the field's knowledge and paving the way for further advancements. A comprehensive comparison of the previously discussed related works and our proposed survey focusing on Trustworthy FL is provided in Table 1. 
In this paper, we establish a clear taxonomy for Trustworthy FL by examining trust evaluation methods and relevant research. This foundational analysis aims to enhance our understanding of Trustworthy FL. We have organized the concept into three primary pillars: Trust-aware Interpretability, Fairness-aware Trustworthiness, and Security and Privacy-aware Trustworthiness in FL. Each pillar contains subcategories that further explore their respective key aspects. The remainder of the paper presents an in-depth investigation of each pillar in separate sections. This research is the first of its kind to encompass all dimensions of Trustworthy FL. Fig. 4 presents an illustration of the taxonomy for Trustworthy FL. For clearer comprehension by our readers, we have further refined this taxonomy. We categorize the algorithms and methodologies based on their primary objectives and these are then elaborated in their respective sections throughout the remaining part of the document.

\begin{figure*}[!ht]
\centering
\includegraphics[width=17cm,height=14cm,keepaspectratio]{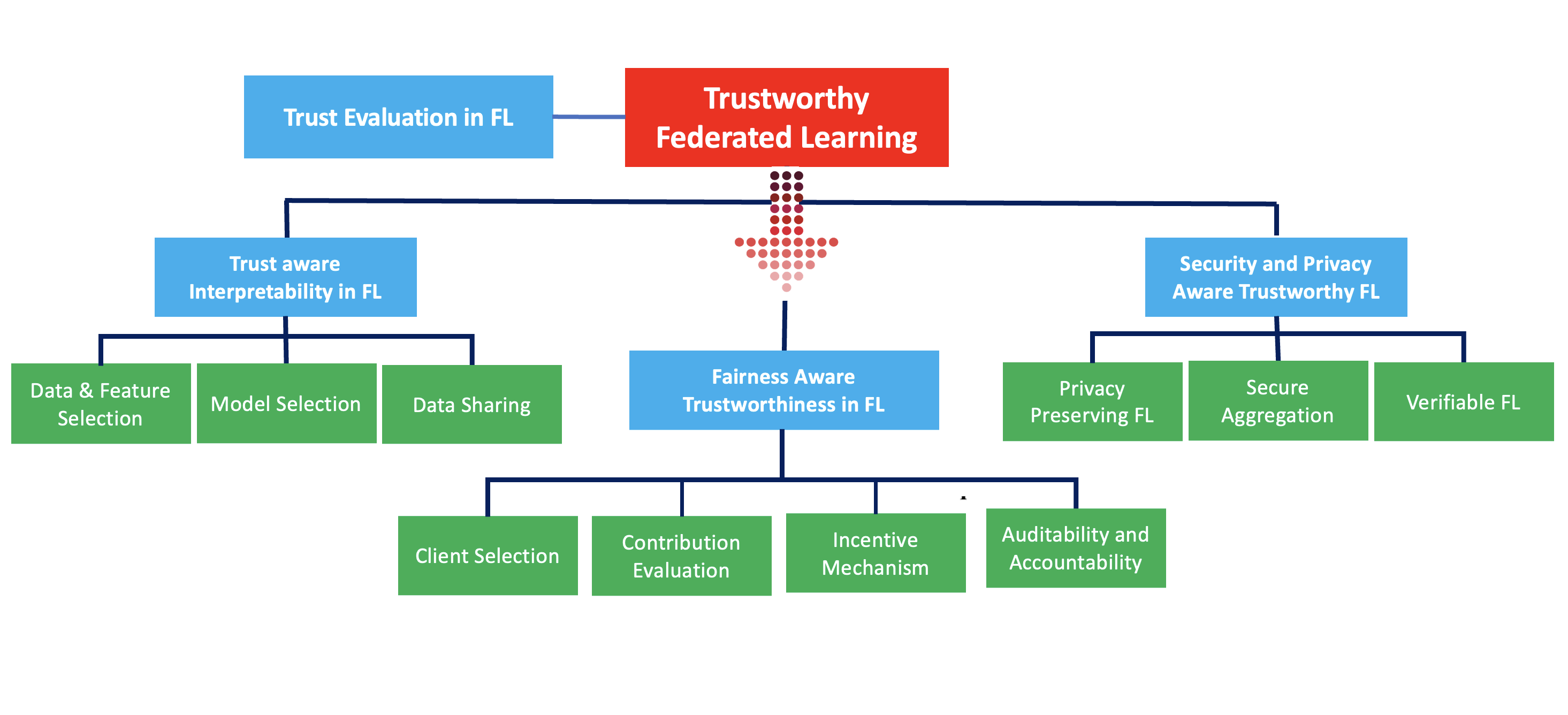}
\caption{An Illustration of Trustworthy FL taxonomy.}
\label{fog6}
\end{figure*}

\section{Trust Evaluation in FL}

In this section, we present a collection of criteria for assessing trust evaluation methods grounded in FL:

\subsubsection{Effectiveness} A vital aspect of trust evaluation is the accurate determination of a trustee's trust value. Trustworthy methods must ensure precision, demonstrated by metrics like recall, precision, accuracy, and F-score.

\subsubsection{Data and Algorithm Selection} Trust evaluation relies on two critical components: training data and model-building algorithms. Optimal data and algorithm choices lead to accurate evaluations, and methods should consider their impact on trust assessment.
\subsubsection{Robustness} Trust evaluation is vulnerable to attacks. Addressing these attacks enhances resistance to disruptions, ensuring robust evaluation methods.
\subsubsection{Privacy Protection} Trust evaluation data might include sensitive user information. It is crucial to protect this data from unauthorized disclosure, prioritizing user privacy and trust evidence protection in trust evaluation processes.
\subsubsection{Context-Awareness} Trust evaluation methods should be adaptable to changes in application scenarios, contexts, or environments, reflecting the fundamental characteristic of trust: context-awareness.
\subsubsection{Subjectivity} Trust evaluation must capture trust's subjective nature for a more authentic representation, emphasizing the importance of subjectivity as a key trust characteristic.
\subsubsection{Distributed Learning:} In FL, trust evaluation methods should account for distributed data storage and processing, ensuring that trust assessment is compatible with the decentralized nature of the learning process.
\subsubsection{Local Model Quality:} Trust evaluation should consider the quality of local models, as FL relies on combining local models to create a global model. Assessing the quality of local models can help identify unreliable participants.
\subsubsection{Incentive Mechanisms:} Implementing incentive mechanisms can encourage honest participation and cooperation, elevating trust assessment in FL by ensuring that participants are motivated to contribute high-quality data and models.
\subsubsection{Contribution Evaluation:} Trust evaluation methods should incorporate mechanisms to measure the value of each client's contributions to the global model. These mechanisms should consider factors such as data quality, data diversity, and the impact of the local model on the global model's performance.
\subsubsection{Client Selection:} Incorporating client selection strategies in trust evaluation helps identify and select reliable clients that can contribute effectively to the global model. By selecting trustworthy clients, the overall quality and trustworthiness of the FL system can be improved.
\subsubsection{Verifiability and Auditibility:} Trust evaluation methods should provide a means of verifying the accuracy and reliability of both local models and the global model. This could involve techniques such as cryptographic proofs, secure aggregation, or trusted execution environments, ensuring transparency and trustworthiness in the FL system.

In the field of network security, trust is considered a crucial aspect for ensuring the secure transmission of data. In the context of the vehicle-road-cloud collaborative system, trust evaluation becomes increasingly complex due to the heterogeneity of the network and its openness to attacks. To address this challenge, the authors in \cite{te1} proposed a trust evaluation scheme based on FL. The scheme is designed as a hierarchical trust evaluation model that takes into account the different trust indices at various layers and factors affecting trust among nodes. The proposed model updates trust values in real-time, providing a personalized trust evaluation for each device in the network. This allows for a more thorough assessment of trust than traditional trust evaluation mechanisms, while also reducing the energy consumption and increasing accuracy compared to previous schemes. By combining FL with the hierarchical trust evaluation model, the system solves the problem of limited edge node resources and reduces the overhead of trust evaluation. This innovative approach to trust evaluation in the vehicle-road-cloud collaborative system shows promising results in improving network security and reliability.

The authors in \cite{te2} proposed a solution to the problem of trust in group decision-making for FL systems. The key contribution of their work is the introduction of a trust-based consensus method, called Trust X-BFT (TX-BFT), which utilizes a consortium chain to reach a consensus among participants. The TX-BFT method evaluates the trust levels of participants based on their historical behaviors in previous consensus processes and stores this information in a public ledger on the blockchain. This information is used to incentivize participants with higher trust levels by rewarding them and punishing those with lower trust levels. This, in turn, helps to improve the overall trust perception performance of the FL network. The proposed method has three stages - preliminary, prepare, and commit - and utilizes a parliament of consensus nodes to communicate and reach a consensus. In each round, a leader collects block generation proposals and broadcasts the pre-prepare information, while the verifiers wait for the pre-prepare message. Once the verifiers receive 2/3 commit messages, they begin inserting the proposed block into the chain and marking their status as final committed. The simulation results and security analysis demonstrate that the TX-BFT method can effectively defend against malicious users and data, and enhance the trust and stability of FL networks. The authors' contribution provides a valuable solution to the problem of trust in group decision-making for FL systems and has the potential to be widely adopted in various applications.

A clustering-based and distance-based trust evaluation methods are proposed in \cite{te3}. The clustering-based method groups FL agents based on their trust scores, while the distance-based method calculates the trust scores based on the similarity of FL agents' behaviors. The authors introduce the trusted decentralized FL algorithm, which incorporates the trust concept into the FL process to enhance its security. This paper addresses the challenge of enhancing the security of FL by introducing trust as a metric to measure the trustworthiness of FL agents. The authors propose a mathematical framework for trust computation and aggregation in a multi-agent system, which is the main contribution of the paper. This framework enables the calculation of trust scores for each FL agent based on their behavior, which can then be used to assess the risk of malicious attacks. Furthermore, the authors propose an attack-tolerant consensus-based FL algorithm, which takes into account the trust scores of FL agents during the consensus step. This helps to mitigate the risk of malicious attacks and ensures the security of the FL training.

PRTrust in \cite{te4}, a trust model is proposed for a peer-to-peer federated cloud system. The authors aim to address the challenge of trust establishment among participating cloud service providers (CSPs) to enable resource sharing in a secured manner. The trust model considers both reputation-based trust and performance-based risk in evaluating the trustworthiness of CSPs. PRTrust provides a two-tier weighted performance evaluation mechanism, a risk evaluation mechanism, and a personalized reputation-based trust evaluation mechanism. It also provides a CSP selection mechanism based on the evaluated trust and risk. The authors intend to reduce the risk of sub-standard service delivery and improve the selection of appropriate CSPs for resource and service sharing.

The security of the Internet of Vehicles (IoV) relies heavily on trust management between various connected devices and nodes. With the increasing number of connected vehicles, it becomes imperative to establish trust and identify dishonest nodes. To improve the security of IoV, a new approach for trust management is proposed in \cite{te5}, which combines FL with blockchain technology (FBTM). This approach involves designing a vehicular trust evaluation to enhance the data acquired for the FL model and developing a blockchain-based reputation system to securely store and share the global FL models. The proof of reputation consensus is also proposed to evaluate the reliability of roadside units functioning as aggregators in the IoV network. Simulation results demonstrate the effectiveness of the proposed FBTM approach in ensuring the security of the IoV network.

The research work in \cite{te6} proposes a Federated Hierarchical Trust Interaction (FHTI) scheme for the Cross-Domain Industrial Internet of Things (IIoT) to address the challenge of multidomain trust interaction. To achieve this, the FHTI scheme integrates consortium blockchain and FL in a seamless manner. A blockchain-based trusted environment for the IIoT is established, followed by the development of a multidomain behavior detection mechanism using FL. The hierarchical trust system is then constructed by combining blockchain transaction performance, leading to unified trust management across multiple domains. Finally, a blockchain cross-chain interaction mechanism is proposed to ensure the credibility of trust values between parties. The main contributions of the article include a two-tier consortium blockchain security architecture and a hierarchical trust mechanism based on federated detection of blockchain nodes, which enables dynamic trust evaluation and hierarchical trust management, thereby improving trust between IIoT devices and breaking down trust barriers between domains.

The proposed system \cite{te7} combines FL with trust establishment mechanism and recommender selection strategy to address the challenge of cold-start items in recommendation systems. The cold-start problem occurs when a recommender system has limited or no information about a new user or item. To address this issue, the authors propose a trust establishment mechanism that enables the recommender system to build trust relationships with potential recommenders. The trust scores are derived from the devices' resource utilization data and the credibility scores of the recommenders. Additionally, the authors propose a recommender selection strategy based on double deep Q learning that considers the devices' trust scores and energy levels to choose the subset of IoT devices that will contribute to improving the accuracy of the recommendations. The authors demonstrate the value of FL for the cold-start item recommendation problem and provide insights into how to design intelligent algorithms that support the FL process while prioritizing trust management.

The authors presents a novel approach to address the challenges of trust management in cross-domain industrial IoT by introducing the FHTI architecture in \cite{te8}. It combines the power of consortium blockchain and FL to provide a safe and reliable network environment for users. The FHTI scheme is based on a behavior detection mechanism that uses FL to evaluate the trustworthiness of devices in a multidomain setting. The architecture also establishes a blockchain cross-chain interaction mechanism that ensures the credibility of the trust value of both parties. The results of the simulation indicate that the proposed scheme can improve the accuracy of abnormal behavior recognition, increase resource utilization, and enhance the stability of the system compared to traditional methods. The FHTI scheme presents a promising solution for trust management in the cross-domain industrial IoT.

\section{Trust aware interpretability in FL}
In this section, we have delved into the comprehensive research efforts undertaken to explore trust-aware interpretability in FL. Various factors are considered in the pursuit of trustworthy interpretability within FL models, including algorithm transparency, model selection, data and feature selection, sample selection, and data sharing evaluation. This objective seeks to provide a clear and comprehensive understanding of ML and DL models, with some being inherently interpretable and others necessitating additional exploration.

To foster trust-aware interpretability in FL, attention must be given to enhancing the quality of clients' local features, which affects the performance of both local and global FL models. Identifying and interpreting crucial features, while eliminating noisy and redundant ones, is vital for achieving trustworthy interpretability. Taking into account the diverse data held by clients in FL systems, it is imperative to acknowledge that not all training data carries equal importance for a specific FL task. By implementing trust-aware interpretability in sample selection, the server and clients can interpret the usefulness of local data, ultimately improving training efficiency and model performance.

Trustworthy interpretability in model optimization can be attained by designing intrinsically interpretable models or robust aggregation methods. Interpretable models incorporate interpretability directly into the model structures, while interpretable robust aggregation allows the FL server to assess the quality of clients' updates and perform quality-aware model aggregation. Thus, trust-aware interpretability in FL enhances the overall reliability and transparency of the system.

\subsection{Trustworthy Feature and Sample Selection}
The authors proposes a new approach called Federated Stochastic Dual-Gate based Feature Selection (FedSDG-FS) \cite{fds1} for feature selection in Vertical FL (VFL). Existing FS works for VFL assume prior knowledge on the number of noisy features or the threshold of useful features to be selected, making them unsuitable for practical applications. FedSDG-FS uses a Gaussian stochastic dual-gate to approximate the probability of a feature being selected with privacy protection through Partially Homomorphic Encryption without a trusted third-party. It also proposes a feature importance initialization method based on Gini impurity to reduce overhead. Experiments show that FedSDG-FS outperforms existing approaches in selecting high-quality features and building global models with higher performance. The proposed method solves the problem of efficient feature selection in VF.

A trustworthiness evaluation framework, TrustE-VC, is proposed in \cite{fds2} that combines criteria importance and performance rates to determine the service attributes of vertical FL that require more attention. It also suggests a three-level security feature to enhance effectiveness and trustworthiness in VC. The proposed framework comprises three interconnected components, including an aggregation of the security evaluation values, a fuzzy multicriteria decision-making algorithm, and a simple additive weight associated with importance-performance analysis and performance rate to visualize the framework findings. The proposed framework provides a useful tool for designers and industrial CV practices to evaluate and select industrial CV trust requirements. The framework addresses the challenges of developing effective and trustworthy VFL models.

In \cite{fds3}, authors present an XAI Federated Deep Reinforcement Learning model aimed at improving decision-making for new Autonomous Vehicles (AVs) in trajectory and motion planning. This model tackles appropriate AV selection for FL and guarantees explainability and trustworthiness. Using XAI, it determines each feature's importance and AV's trust value. A trust-based deep reinforcement learning model is introduced for selections, showing superior performance in real-world data experiments. The study highlights trust's role in AV selection and proposes an innovative XAI-based trust computation method, providing a sophisticated mechanism for new AVs' decision-making.

The main contribution of \cite{fds4} is a FL model named FedPARL, which aims to reduce the model size while performing sample-based pruning, avoiding misbehaved clients, and considering resource-availability for partial workloads. This is especially useful for resource-constrained IoT clients. FedPARL, a tri-level FL strategy, aids clients in conserving resources throughout training, eliminates unreliable or resource-deficient clients during selection, and allows for flexible local epochs based on client resource availability. An incentive-deterrent framework promotes effective clients and discourages poor-performing or malicious ones. This approach exhibits robustness in constrained FL-IoT environments, and results reveal that FedPARL outperforms existing methods, delivering an enhanced FL solution.

This authors proposes a new approach to optimize smart device sampling and data offloading in FL \cite{fds5}. The authors propose a joint sampling and data offloading optimization problem where devices are selected based on their expected contribution to model training. The non-selected devices can transfer data to selected ones based on estimated data dissimilarities between nodes. The proposed approach aims to improve the efficiency and accuracy of FedL by reducing the communication and computational costs. The approach is evaluated using real-world data, and the results demonstrate its effectiveness in improving the performance of FedL.

A new framework is proposed for Importance Sampling FL (ISFL) \cite{fds6}, which addresses the non-i.i.d. data distribution issue. The proposed framework mitigates the performance gap by introducing local importance sampling and formulating the weighting selection of each client as an independent optimization sub-problem. The paper presents theoretical solutions for optimal IS weights and an adaptive water-filling approach for numerical solutions. The model deviation bound is derived between ISFL and centralized training, relaxing the restriction of convexity of loss functions. The proposed framework offers a local importance sampling-based solution to the label-skewed non-i.i.d. data distribution problem in FL.

The Quality Inference (QI) method is proposed to recover the quality of the aggregated updates and participants' datasets \cite{fds7}. QI uses inferred information across multiple rounds and known per-round subset to evaluate the relative quality ordering. The method assigns scores to contributors based on The Good, The Bad, and The Ugly rules. QI successfully recovers the relative ordering of participant's dataset qualities when secure aggregation is in use, without requiring computational power or background information. The proposed method can be useful in ensuring trustworthy and high-quality FL in a decentralized setting.

TrustFL \cite{fds8}, a scheme that utilizes Trusted Execution Environments (TEEs) to ensure the integrity of participant's training executions in FL. FL faces new security challenges due to the lack of direct control over the training process. The proposed scheme randomly verifies a small proportion of the training processes using TEEs for a customizable level of assurance, while computations are executed on a faster but less secure processor for improved efficiency. TrustFL also employs a commitment-based method with specific data selection to counteract cheating behaviors. The proposed scheme provides high confidence in participants’ training executions in FL.

\subsection{Trustworthy Data Sharing}
Centralized data sharing mechanisms are vulnerable to several issues, including single point of failure, data privacy, authenticity, equitable income distribution, and low user engagement. Despite being simple and easy to implement, these limitations hinder their effectiveness. Alternative approaches are needed to ensure secure and trustworthy data sharing, promoting collaboration, and facilitating innovation in data-driven industries \cite{ds1, ds2}.

This authors in \cite{ds3} proposes a solution to the problem of data, model, and result trust in cross-domain data sharing by using blockchain and cryptography to establish an endogenous trusted architecture. The proposed reverse auction node incentive mechanism based on high credit preference addresses issues such as low user enthusiasm, unstable data quality, and unfair data sharing benefit distribution. The decentralized, tamper-proof, and traceable nature of blockchain ensures a trusted trading environment, while FL combined with differential privacy enhances user privacy and data security sharing. The proposed approach provides a potential solution for a more secure and efficient data sharing framework, enabling users to participate in data sharing and submit higher quality data.

The authors present a reliable framework combining FL and BC in \cite{ds4}, for enhanced security and trustworthiness in IoT network . The framework employs a trust evaluation mechanism and introduces a reinforcement-based FL (R-FL) system for managing IoT devices' training models. The study assesses network lifespan, energy usage, and trust, while addressing communication security and device mobility with the adaptive FL-based trustworthiness evaluation (AFL-TE) system.

The integration of blockchain and FL is a promising approach to achieve trusted data sharing with privacy protection. However, existing mechanisms overlook the supervision of the FL model and computing process. To address this issue, \cite{ds5} proposes a new paradigm for trusted sharing using the concepts of sandbox and state channel. The state channel creates a trusted sandbox to instantiate FL tasks in a trustless edge computing environment, while also solving data privacy and quality issues. An incentive mechanism based on smart contract encourages local devices and edge nodes to participate in FL tasks, and a DRL-based node selection algorithm selects different node sets with the reward of epoch delay and training accuracy. The proposed architecture uses PBFT consensus mechanism to ensure smooth generation of blocks and reduce communication delay for the blockchain platform. The DRL algorithm effectively deals with the node selection problem with a large action space, providing better accuracy and delay performance than traditional algorithms.

The study in \cite{ds6} presents a blockchain and FL-based architecture for secure data sharing, prioritizing user privacy and data value transmission. The architecture uses on-chain and off-chain storage, user identity authentication, and data integrity verification. It employs an ABAC-based fine-grained access control model with transaction attributes and node reputation. The paper also suggests a state channel-based FL training supervision mechanism, addressing trust challenges in cross-domain data sharing by utilizing a state channel-based trust supervision mechanism. This approach enhances system security and trust while minimally affecting FL efficiency.

FL has the potential to break down data silos and enhance the intelligence of the Industrial Internet of Things (IIoT). However, the principal-agent architecture used in this approach increases costs and fails to ensure privacy protection and trustworthiness in flexible data sharing. In response, the authors propose a secure and trusted federated data sharing (STFS) approach based on blockchain \cite{ds7}. They first construct an autonomous and reliable federated extreme gradient boosting learning algorithm to provide privacy protection, verifiability, and reliability. Then, they design a secure and trusted data sharing and trading mechanism that includes encryption for secure data storage, threshold aggregation signature to guarantee model ownership, and proxy re-encryption and retrieval for controllable and trusted data sharing. The proposed approach can enhance the connectivity of IIoT data and ensure secure and controlled sharing while protecting privacy and ownership.

This authors in \cite{ds8}, presents a blockchain-enabled FL model for the industrial Internet of Things (IIoT) to solve data sharing and model sharing security issues. The proposed data protection aggregation scheme ensures privacy and security in data sharing. Additionally, the paper proposes three distributed ML algorithms: K-means clustering based on differential privacy and homomorphic encryption, random forest with differential privacy, and AdaBoost with homomorphic encryption. These methods enable multiple data protection in complex IIoT scenarios.

The concept of Self-Driving Networks (SelfDN) and a framework to create distributed and trustworthy SelfDNs across multiple domains is presented in \cite{ds9} . The framework utilizes programmable data planes for real-time In-band telemetry, AI for automatic network device code rewriting, and blockchain and FL for decentralized knowledge sharing among domains. The effectiveness of this approach is demonstrated through a proof-of-concept experiment where INT, Deep Learning, and P4 were used to autonomously detect and mitigate application-layer DDoS attacks at the data plane level. The proposed framework provides a promising approach to building self-driving networks that are secure, trustworthy, and effective in detecting and responding to network threats.

In \cite{ds10}, the authors propose a blockchain-supported FL (BFL) marketplace, integrating social Internet of Things (SIoT) to enable FL in devices with computational constraints. The BFL marketplace allows devices to exchange FL services, with blockchain standardizing market operations and logging transactions. The paper introduces a trust-enhanced collaborative learning strategy (TCL) and a quality-focused task allocation algorithm (QTA) for handling trust relationships among heterogeneous IoT devices and directing FL task allocation to proficient devices for optimal training quality. To maintain long-term stability, an encrypted model training scheme (EMT) is designed to defend against malicious attacks, and a contribution-based delegated proof of stake (DPoS) consensus mechanism ensures equitable reward distribution. These algorithms successfully utilize data from computationally limited devices for FL through SIoT while safeguarding security and fairness in the BFL marketplace.

\subsection{Trustworthy Model Selection}

In FL, it is crucial to evaluate the contributions of participants to the performance of the final model while ensuring privacy. To achieve this, the widely adopted method is the use of Shapley Value (SV) techniques. However, existing SV-based approaches are computationally expensive and impractical for real-world applications. To tackle this issue, authors in \cite{ms1} introduced the Guided Truncation Gradient Kapley (GTG-Shapley) approach, which reduces the computation cost of SV-based FL participant contribution evaluation. Unlike traditional methods, GTG-Shapley does not require extra learning tasks from participants, as it reconstructs FL sub-models using their previous gradient updates instead of training them from scratch. Additionally, GTG-Shapley employs guided Monte Carlo sampling to further reduce the number of required model reconstructions and evaluations, thereby enhancing the efficiency of SV computation. GTG-Shapley offers a more practical and scalable solution for fair FL participant contribution evaluation. GTG-Shapley enables FL to be more practical and widely adopted in real-world applications.

The aggregation of local models from participating clients is a critical component in generating the final global model in FL. However, traditional aggregation methods can be susceptible to adversarial attacks and client failures. To mitigate this issue, the authors of this paper propose a truth inference approach to FL that incorporates the reliability of each client's local model into the aggregation process. The proposed approach in \cite{ms2} models the clients' reliability based on their submitted local model parameters and considers these parameters during the aggregation process to produce a robust estimate of the global model. The authors have further enhanced the method by considering the model parameters submitted by clients in previous rounds in addition to the current round, thus providing a more comprehensive evaluation of client reliability.The proposed truth inference approach provides a more robust estimate of the global model, protects against potential adversarial attacks, and considers client reliability in the aggregation process, thereby improving the robustness of FL.

In FL, the server aggregates the uploaded model parameters from participating clients to generate a global model. The common practice is to evenly weight the local models, assuming equal contribution from all nodes. However, the heterogeneous nature of devices and data leads to variations in contribution from users. To address this issue, authors in \cite{ms3} introduces a reputation-enabled aggregation method that adjusts the aggregation weights based on the reputation scores of users. The reputation score is computed based on the performance metrics of the local models during each training round. The proposed method showed an improvement of 17.175\% over the standard baseline in non-independent and identically distributed (non-IID) scenarios for a FL network of 100 participants. This work considers the mobile network of distributed computing nodes where the performance and reputation of individual nodes vary. The reputation-enabled weighted aggregation is hypothesized to lead to faster convergence and a higher accuracy level for FL in a mobile environment.

In an effort to improve privacy and reward mechanism, the research work in \cite{ms4} proposes a block-chained FL (BlockFL) architecture that uses blockchain technology instead of a central entity. With BlockFL, each device computes and uploads its local model update to a miner in the blockchain network and receives a data reward proportional to the number of its data samples. Miners exchange and verify all the local model updates, and run Proof-of-Work (PoW). Once a miner successfully completes PoW, it generates a block that stores the verified local model updates and receives a mining reward. The generated block is added to a blockchain, which is downloaded by the devices. The devices then compute the global model update from the latest block, which is used as input for the next local model update. This architecture allows for on-device ML without central coordination, even when each device lacks its own training data samples.

In \cite{ms5}, a new approach to FL is proposed, focusing on the improvement of learning speed and stability. The approach includes three key components: a node recognition-based local learning weighting method, a node selection method based on participation frequency and data amount, and a weighting method based on participation frequency. The performance of this proposed approach is compared to traditional FL and the results show that it outperforms the traditional method in terms of both learning speed and stability. The paper presents a unique solution for improving the performance of FL through the use of blockchain-based node recognition.

The authors in \cite{ms6}, presents a framework for secure and privacy-preserving deep learning (DL) services in the Industrial Internet of Things (IIoT) systems . This framework leverages FL to aggregate multiple locally trained models without sharing datasets among participants, thereby overcoming the privacy challenges of traditional collaborative learning. However, FL-based DL (FDL) models can be vulnerable to intermediate results and data structure leakage. The proposed framework comprises a service-oriented architecture that identifies key components and implements a service model for residual networks-based FDL with differential privacy (DP) to produce trustworthy locally trained models. The services in the framework ensure secure execution through privacy preservation, while the privacy-preserving local model aggregation mechanism ensures further privacy protection. The framework features DP-based residual networks (ResNet50) that use GR to guarantee privacy during federated training, and a trusted curator who adds random noise to the function output to ensure global privacy. Additionally, DP is leveraged in the DL model architecture to generate DP local model representations during training, providing extra protection against data leakage. The DP approach has two important properties: composition and postprocessing.

A new framework for robust FL is proposed in \cite{ms7} to address the vulnerability of FL systems to malicious clients. These clients can send malicious model updates to the central server, leading to a degradation in learning performance or targeted model poisoning attacks. To tackle this issue, the proposed framework uses spectral anomaly detection to detect and remove the malicious model updates. The framework is evaluated in image classification and sentiment analysis tasks, and results show that the low-dimensional embeddings used in the framework can easily differentiate the malicious updates from normal updates, leading to targeted defense. The proposed solution is an effective method to address the threat of adversarial attacks in FL systems.

Automated FL (AutoFL) in \cite{ms8} is proposed to enhance model accuracy and simplify the design process. The focus is on using Neural Architecture Search (NAS) for automation, leading to the development of a Federated NAS (FedNAS) algorithm. This algorithm enables the scattered workers to work together and find an architecture with improved accuracy. The implementation of FedNAS is also demonstrated through a system build.

The Federated Graph Convolution Network (FGC) in \cite{ms9} is a new approach to recommendation systems that combines privacy and accuracy. Unlike traditional methods that gather raw data, the FGC approach allows clients to keep their data locally and only upload model parameters to a central server. This protects privacy while improving prediction accuracy. The FGC approach also features a model segmentation method that adjusts to varying weight dimensions, ensuring global weight aggregation. Additionally, it improves the calculation of service node embeddings by focusing only on relevant data, reducing the impact of noise and increasing trustworthiness and accuracy. The goal of recommendation systems is to leverage historical behavior and knowledge, but the model accuracy is often limited by the risk of data leaking from multiple departments. The FGC approach addresses these limitations by having clients train locally and upload only model weights to the server, while also leveraging overlapping services to optimize local training results. Overall, the FGC approach offers a novel solution to recommendation systems by balancing privacy and accuracy, while avoiding the risk of data leaking.

A hierarchical framework of federated control for Industrial Internet of Things (IIoTs) is proposed in \cite{ms10}, to address the trustworthiness and privacy preservation of tracking systems. The framework consists of a federated control center, network layer, and a federated control node, and integrates a collaborative Cloud-Edge-End structure and a ML-oriented localization method based on Expectation Maximization (EM). A trustworthy localization model is built using the EM method, which iteratively solves for the latent variable of untrustworthiness probability. The EM-based federated control scheme offers a solution to the trustworthiness issue in IIoTs while preserving privacy.

Authors in \cite{ms11} introduces the Federated Trustworthy Artificial Intelligence (FTAI) Architecture, which combines TAI Architecture and FL to provide a secure platform for user data privacy. The proposed model aggregation strategy integrates FedCS and FedPSO, and employs AIF360 to guarantee fairness in the client-side training process by eliminating discrimination. The main contributions of this paper are: (1) a trustworthy and secure architecture for protecting user data privacy; (2) a client system that is tolerant to heterogeneity and low bandwidth; and (3) a fair and improved model.

The research study in \cite{ms12}, authors introduced the concept of Fine-Grained FL, aimed at decentralizing shared ML models on edge servers. This work outlines a comprehensive definition of Fine-Grained FL in Mobile Edge Computing systems and the key requirements of these systems, including personalization, decentralization, incentives, trust, and efficiency in communication and bandwidth. To ensure trustworthy collaboration, authors propose the use of a Blockchain-based Reputation-Aware Fine-Grained FL system that provides all participants with reputation information via Frontend DApps. This system leverages Ethereum's public blockchain and smart contract technologies to compute trustworthy reputation scores and aggregate them for model selection and aggregation. The reputation information about each device acts as a deterrent against malicious, faulty, and ghost devices in achieving the requirements of Fine-Grained FL.

In \cite{ms13}, authors propose a new FL system design that ensures the security of individual model updates during the learning process. The system enables clients to provide encrypted model updates while the cloud server performs aggregation. Our design differs from previous works by supporting lightweight encryption, aggregation and resilience against drop-out clients with no impact on future participation. Our system is designed to handle client drop-out while keeping their secret keys confidential. To improve communication efficiency, we employ quantization-based model compression and secure aggregation. Additionally, we present mechanisms to make our system more resilient against adversarial servers. Our experiments on real-world datasets show that our system achieves comparable accuracy to plaintext baselines with practical performance. By integrating a cherry-picked aggregation protocol, our system offers practical encryption of model updates at the client and aggregation of encrypted model updates at the cloud server.

In \cite{ms14}, the authors introduce a hybrid blockchain architecture, PermiDAG, to address the transmission load and privacy concerns in FL systems. The architecture combines a permissioned blockchain maintained by road-side units (RSUs) with a local Directed Acyclic Graph (DAG) run by vehicles for efficient data sharing in the Internet of Vehicles (IoV). The authors also propose an asynchronous FL approach using Deep Reinforcement Learning (DRL) for node selection to improve efficiency. The learned models are integrated into the blockchain and their quality is verified through two-stage validation, ensuring the reliability of shared data. The authors aim to improve the security and reliability of model parameters through the proposed hybrid architecture and FL approach.

In \cite{ms15}, the authors present a method for improving the performance of FL models by assigning a reputation score to individual models. This score is calculated based on various performance metrics and is used to select the best models for aggregation. The proposed scheme ensures that the final aggregated model is of higher quality as it is based on the performance of models with high reputation scores. The reputation score is a novel addition to the FL process that enhances the performance and reliability of the final model.

Authors proposes a reputation opinion-based PoL consensus protocol for edge blockchain in IIoT, which enhances the trustworthiness of edge intelligence \cite{ms16}. The protocol employs a smart contract to obtain reputation opinions, reducing the impact of malicious or accidental anomalies and minimizing reliance on trusted intermediaries. Trustworthy edge intelligence is achieved by adopting the winner's intelligence through a weight aggregation of the winner's learning model based on its reputation opinion, rather than completely discarding all local models. The proposed scheme is analyzed for performance in terms of security, latency, and throughput, and simulation results demonstrate its effectiveness. This approach offers a new way to handle model selection in FL, enabling more reliable and trustworthy model aggregation in a distributed setting.

\subsection{Discussion}
In the feature and sample selection domain (\cite{fds1}-\cite{fds8}), the studies propose various techniques for efficient and trustworthy feature selection and model optimization. However, they often overlook the trustworthiness of clients or devices participating in the FL process. To address this limitation, future research should focus on incorporating trust metrics and secure aggregation techniques to ensure the integrity of the selected features and data. In the model selection domain (\cite{ms1}-\cite{ms16}), the studies propose different mechanisms to evaluate clients' contributions and enhance the security of the FL process. However, they tend to focus on specific application scenarios or client behaviors, limiting their generalizability. To improve these studies, it is crucial to develop more comprehensive frameworks that consider different aspects of trustworthiness, such as data privacy, model robustness, and client reputation. Additionally, techniques should be adaptable to various application domains and client behavior patterns. In the data sharing domain (\cite{ds1}-\cite{ds10}), blockchain-based approaches have been widely adopted to ensure trustworthiness in data sharing for FL. Despite the innovative solutions provided, these studies face challenges in terms of scalability and performance. To overcome these limitations, future research should explore alternative techniques to blockchain, such as secure multiparty computation and homomorphic encryption, which can provide better scalability and efficiency while maintaining data privacy and integrity. Moreover, the integration of these techniques with FL should be investigated to create a seamless and more trustworthy data sharing process. Overall, the existing literature in trustworthy interpretability has made significant strides in addressing trustworthiness in FL. However, there is still room for improvement in terms of generalizability, comprehensiveness, and scalability. Future research should focus on developing frameworks that can efficiently address the limitations and pitfalls identified in the current studies, paving the way for more trustworthy and robust FL systems.
\begin{figure*}[!ht]
\centering
\includegraphics[width=11cm,height=11cm,keepaspectratio]{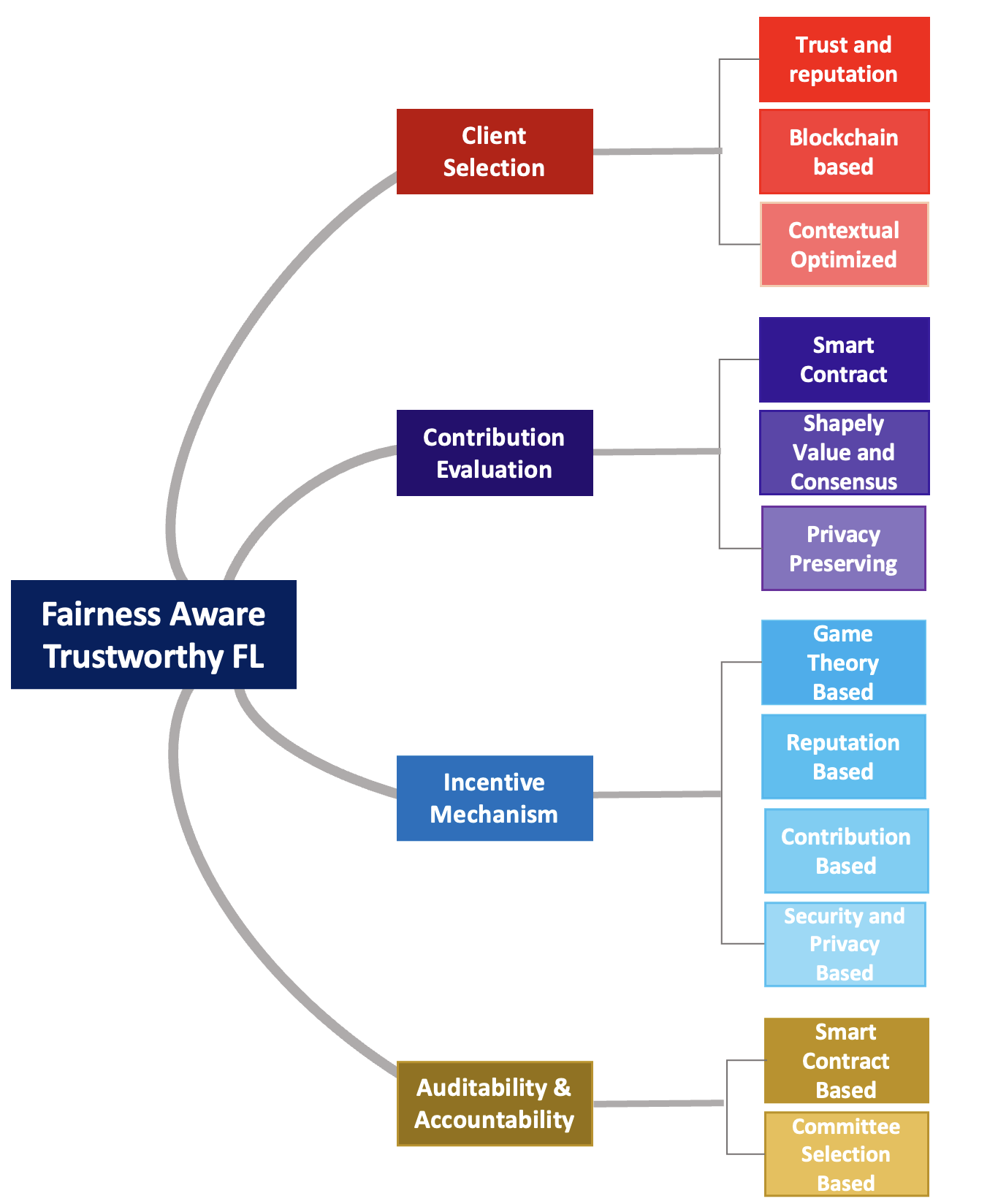}
\caption{A visual overview of the Categorization of Fairness aware Trustworthy FL.}
\label{fog7}
\end{figure*}

\section{Fairness aware Trustworthiness in FL}
Fairness is a critical aspect of Trustworthy FL since many parties provide data for training the model and eventually receive the same combined global model. Key ideas involve client selection that considers accuracy parity and selection fairness. These concepts measure how evenly performance is distributed across FL client devices, with the goal of reducing bias and decreasing under-representation or non-representation. Furthermore, contribution fairness and incentive mechanism fairness aim to fairly distribute rewards based on each client's input. In this section, we will explore existing research on fairness-aware Trustworthy FL. By examining these topics in more depth, we will develop a better understanding of the challenges and opportunities in creating a fair and Trustworthy FL framework that is easily accessible to readers. Fig. 5 provides a visual summary of the topics addressed in the context of fairness-aware Trustworthy FL. Moreover, we established detailed sub-categorization of all the aspects of Fairness aware Trustworthy FL. The Illustration of the refined taxonomy related to Fairness aware Trustworthy FL is presented in Fig. 6.

\subsection{Trustworthy Client Selection}
The selection of FL worker nodes using reputation is faced with the issue of the "cold start problem". In previous studies, the assumption is that there is historical interaction data available to evaluate the reputation of the worker nodes. However, this becomes a challenge when the worker node has no prior interaction with the master node and there is a risk of tampering with the reputation values. To address these uncertainties, contract theory is used to mitigate the cold start problem in reputation-based FL worker node selection.

\subsubsection{Trust and Reputation based Trustworthy Client Selection }
A trust-based deep reinforcement learning approach is proposed for client selection in FL \cite{cs2}. The solution involves a resource and trust-aware DRL strategy that considers the execution time and trust levels of clients to make an efficient selection decision. The proposed solution integrates multiple technologies such as federated transfer learning, IoT, edge computing, trust management and DRL to provide a holistic approach for a detection-driven scenario. A stochastic optimization problem is formulated to determine the selection of IoT devices to which the FL tasks will be sent, with the goal of minimizing execution time while maximizing trust. The optimization problem is then solved using a DRL algorithm that models the server's uncertainty about the resource and trust levels of IoT devices.

In \cite{cs3}, researchers address the challenge of selecting IoT devices to participate in distributed training within FL. Existing approaches primarily consider device resource features, but the authors emphasize that trust should also factor into decision-making. To tackle this, they create a trust-building mechanism between edge servers and IoT devices, introducing DDQN-Trust, a selection algorithm based on double deep Q learning that considers both trust scores and energy levels of IoT devices. This solution is integrated FedAvg, FedShare, FedProx and FedSGD. By accounting for resource characteristics and trust, the proposed method delivers a holistic solution for organizing IoT devices in FL scenarios.

The lack of profit motivates participants to provide low-quality data and prevents requesters from identifying reliable participants. To address this, authors proposes a horizontal FL incentive mechanism called RRAFL in \cite{cs6}, which combines reputation and reverse auction theory. Participants bid for tasks, and reputation indirectly reflects their reliability and data quality. his paper proposes a reputation mechanism as a critical component of the FL incentive mechanism. The reputation is a reflection of the requester's evaluation of the participants and is saved in the interaction blockchain, which is tamper-proof and open to transparency. The reputation is based on the participants' data quality and reliability, which are measured through a model quality detection method and a participant contribution measurement method. The reputation mechanism plays a crucial role in the reverse auction process, where participants bid for tasks and are selected based on their reputation and bid price. The selected participants with good reputation and low bid price are rewarded with the budget. The reputation mechanism is designed to incentivize participants to provide high-quality data and actively participate in the FL program. It also allows the requester to make informed decisions about selecting reliable participants with good data quality.

A Robust and Fair FL (RFFL) framework in \cite{cs7}, designed to address the challenges of achieving both collaborative fairness and adversarial robustness in FL. The framework relies on a reputation mechanism, which evaluates the contributions of each participant by comparing their uploaded gradients to the aggregated global gradients. By examining the cosine similarity between the two, RFFL can identify non-contributing or malicious participants and remove them. The authors emphasize that their approach does not require any auxiliary or validation dataset, setting it apart from other methods. The main contribution of this work is a framework for achieving both collaborative fairness and adversarial robustness in FL via a reputation mechanism, which is evaluated via the comparison of gradients.
\begin{figure*}[!ht]
\centering
\includegraphics[width=17cm,height=14cm,keepaspectratio]{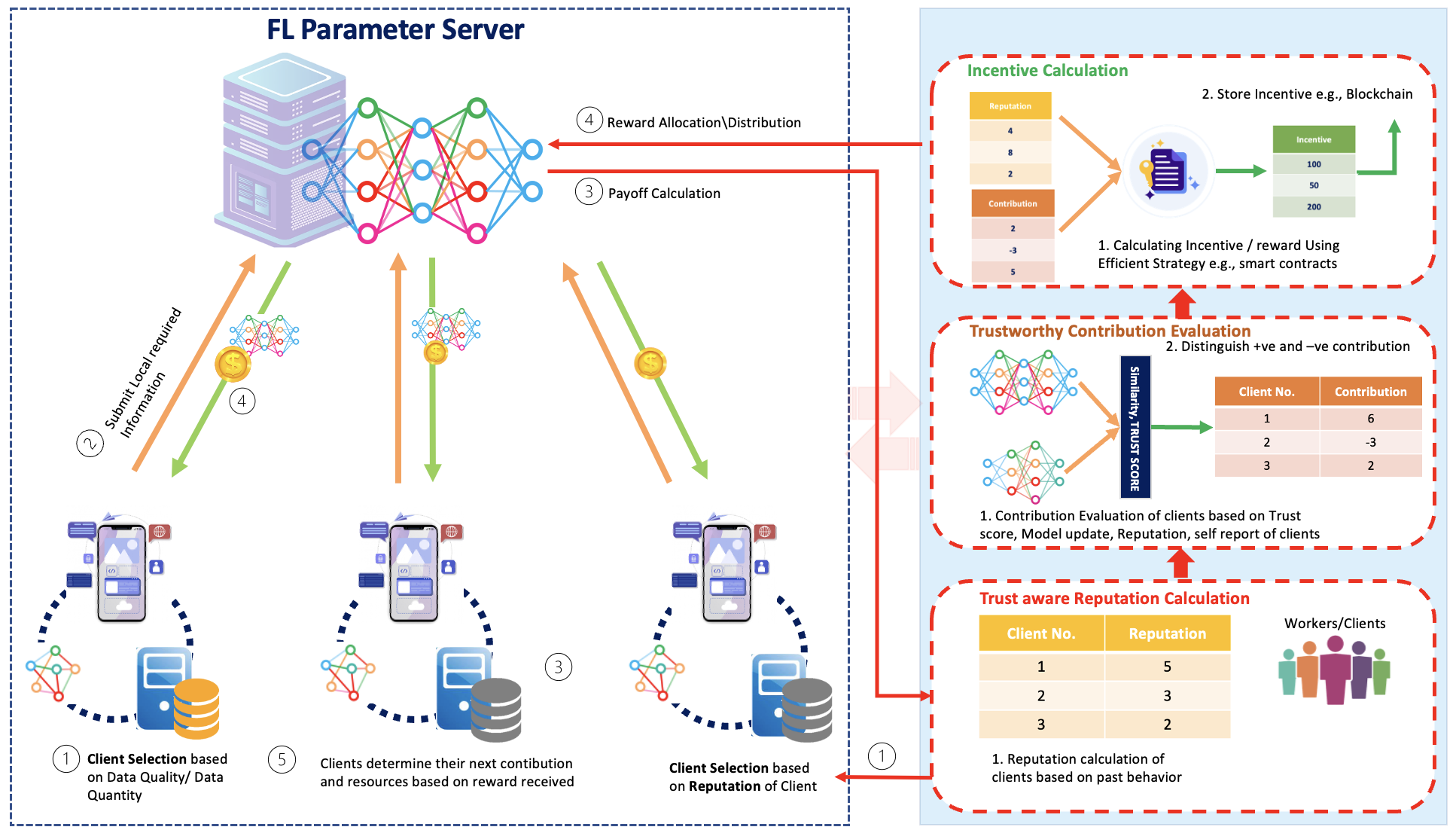}
\caption{An Illustration of fairness-aware trustworthy FL settings, showcasing the key elements and relationships that contribute to achieving fairness and trustworthiness in FL environments.}
\label{fog8}
\end{figure*}
\subsubsection{Blockchain based Trustworthy Client Selection }
FEDAR, A trust and resource-aware FL framework is proposed in \cite{cs10} to address the challenges posed by unreliable and resource-constrained FL environments. Specifically, the authors focus on distributed mobile robots that are often resource-limited and may exhibit inconsistent behavior. To address this issue, the authors introduce a trust score for each FL client, which is updated based on their previous performance and resource availability. The authors then use this trust score to select the most reliable and resource-efficient clients for FL training, while excluding those clients that are deemed untrustworthy. FoolsGold \cite{cs11}algorithm is used to indentify the unreliable participants in fedar. In addition, the authors also address the straggler effect by applying asynchronous FL, which enables the FL server to aggregate the updates from clients without waiting for the slowest client. The authors use this framework to evaluate the performance of FL on resource-constrained mobile robots in a real-world setting and show that the proposed approach is effective in improving the convergence time and ensuring reliable results.

TrustWorker, a worker selection scheme for blockchain-based crowdsensing is proposed in \cite{cs12}that focuses on trustworthiness and privacy protection. TrustWorker leverages the decentralization, transparency and immutability of blockchains to make the worker selection process trustworthy. Reputation privacy of workers is protected through the use of a deterministic encryption algorithm and a secret minimum heapsort scheme. Reputation comparison is carried using two-party comparison protocol\cite{cs13}.  The effectiveness and efficiency of TrustWorker is analyzed theoretically and through experiments. TrustWorker's limitation is its sole focus on worker reputation, without considering the influence of requester reputation on task selection. Future work should investigate this gap while balancing privacy and efficiency.

In \cite{cs15}, the authors aim to address the challenge of unreliable data being uploaded by mobile devices in FL, leading to frauds and low-quality training results. To solve this problem, they propose a new approach that leverages the concept of reputation as a reliable metric to select trusted workers for FL. To calculate the reputation efficiently, a multi-weight subjective logic model is applied, considering both the interaction histories of task publishers and recommended reputation opinions. To ensure secure and decentralized reputation management, the authors use consortium blockchain technology deployed at edge nodes. The consortium blockchain acts as a trusted ledger to record and manage the data owners' reputation, improving the reliability of FL tasks in mobile networks.

In \cite{cs16}, a blockchain-based reputation evaluation method is proposed to improve the security of FL against poisoning attacks. The method evaluates the reputation of each participant based on three factors: interaction reputation, data reputation, and resource reputation. By doing so, trusted participants can be selected and malicious participants can be identified. To enhance the detection of malicious behavior, a combination of multi-domain detection and a distributed knowledge base is proposed. A feature graph based on a knowledge graph is designed to store and manage multi-domain feature knowledge. The proposed solution aims to ensure the robustness of FL by identifying trustworthy participants.

TrustRe, A reputation evaluation scheme for improving worker selection in FL is proposed in \cite{cs17}. Our approach takes a step further from existing work, which relied on subjective judgment for reputation evaluation. Instead, we use the quality of model training as the basis of reputation evaluation and store the reputation values on a blockchain. This ensures privacy protection for workers and accurately records the history of reputation. Our contribution also includes the design of a new blockchain platform for decentralized reputation management. Additionally, we propose a consensus algorithm called proof of reputation (PoR) to aggregate FL models. PoR uses worker reputation as a factor for competition for the bookkeeper role in the blockchain ledger, improving the quality of the global model and encouraging worker participation. Based on the workers' reputation values, a leader worker is selected in the blockchain network, responsible for responding to requests and facilitating consensus.
\begin{table*}[]
\caption{Trustworthy Client Selection in FL}
\label{tab:my-table2}
\resizebox{\textwidth}{!}{%
\begin{tabular}{@{}ccclccccllc@{}}
\toprule
\textbf{\begin{tabular}[c]{@{}c@{}}Proposed \\ Technique\end{tabular}} &
  \textbf{YoP} &
  \textbf{\begin{tabular}[c]{@{}c@{}}Algorithms \\ Used\end{tabular}} &
  \multicolumn{1}{c}{\textbf{Description}} &
  \textbf{TR} &
  \textbf{IM} &
  \textbf{CE} &
  \textbf{RP} &
  \multicolumn{1}{c}{\textbf{\begin{tabular}[c]{@{}c@{}}Problem \\ Encountered\end{tabular}}} &
  \multicolumn{1}{c}{\textbf{Outcome}} &
  \textbf{Dataset} \\ \midrule
\begin{tabular}[c]{@{}c@{}}C2MAB,  RBCS-F \\ \cite{cs1}\end{tabular} &
  2020 &
  \begin{tabular}[c]{@{}c@{}}Online \\ Lyapunov \\ Optimization\end{tabular} &
  \begin{tabular}[c]{@{}l@{}}C2MAB estimating model exchange time based\\ on client contextual properties and historical\\ performance, RBCS-F, efficiently solves the\\ optimization problem with fairness guaranteed\end{tabular} &
  \quad\circledmark[cyan!60] &
  \quad\circledmark[cyan!60] &
   &
  \quad\circledmark[cyan!60] &
  \begin{tabular}[c]{@{}l@{}}Efficiency\\ boosting client\\ selection\end{tabular} &
  \begin{tabular}[c]{@{}l@{}}long-term fairness,\\ improving training\\ efficiency,\\ accuracy\end{tabular} &
  \begin{tabular}[c]{@{}c@{}}FMNIST\cite{d1}, \\ CIFAR-10\cite{dd1}\end{tabular} \\
\begin{tabular}[c]{@{}c@{}}Trust-augmented \\ DRL\\ \cite{cs2}\end{tabular} &
  2022 &
  \begin{tabular}[c]{@{}c@{}}Federated \\ Transfer Learning\end{tabular} &
  \begin{tabular}[c]{@{}l@{}}Trust-aware DRL considers the execution time\\ and trust levels of clients to make an efficient\\ selection decision.\end{tabular} &
  \quad\circledmark[cyan!60] &
   &
   &
   &
  \begin{tabular}[c]{@{}l@{}}A stochastic\\ Optimization\\ problem for client\\ selection\end{tabular} &
  \begin{tabular}[c]{@{}l@{}}Minimizing\\ execution time\\ while maximizing\\ trust\end{tabular} &
  \begin{tabular}[c]{@{}c@{}}COVID-19 \\ dataset \cite{d2}\end{tabular} \\
\begin{tabular}[c]{@{}c@{}}DDQN-Trust\\ \cite{cs3}\end{tabular} &
  2022 &
  CNN &
  \begin{tabular}[c]{@{}l@{}}DDQN-trust considers both trust scores and\\ energy levels of the IoT devices, Using FedAvg,\\ FedProx, FedShare, and FedSGD\end{tabular} &
  \quad\circledmark[cyan!60] &
   &
   &
   &
  \begin{tabular}[c]{@{}l@{}}Efficient Client\\ Selection\end{tabular} &
  \begin{tabular}[c]{@{}l@{}}Higher Accuracy\\ and higher reward\\ collection\end{tabular} &
  CIFAR-10 \\
\begin{tabular}[c]{@{}c@{}}Blockchain-based \\ FL aggregation \\ protocol\\ \cite{cs4}\end{tabular} &
  2022 &
  \begin{tabular}[c]{@{}c@{}}Ethereum smart \\ contract, \\ Contribution \\ assessment\end{tabular} &
  \begin{tabular}[c]{@{}l@{}}The proposed deposit mechanism differentiates\\ between trusted and untrusted clients by imposing\\ different deposit requirements\end{tabular} &
  \quad\circledmark[cyan!60] &
   &
  \quad\circledmark[cyan!60] &
   &
  \begin{tabular}[c]{@{}l@{}}Client dropouts,\\ model poisoning,\\ unauthorized\\ access\end{tabular} &
  \begin{tabular}[c]{@{}l@{}}improved FL\\ performance and\\ security\end{tabular} &
  \begin{tabular}[c]{@{}c@{}}Client \\ dataset\end{tabular} \\
\begin{tabular}[c]{@{}c@{}}SCS\\ \cite{cs5}\end{tabular} &
  2022 &
  \begin{tabular}[c]{@{}c@{}}Stochastic \\ Integer \\ Programming (SIP)\end{tabular} &
  \begin{tabular}[c]{@{}l@{}}SCS uses SIP together with reputation as a\\ metricto select as many reliable worker nodes as\\ possible\end{tabular} &
  \quad\circledmark[cyan!60] &
   &
   &
  \quad\circledmark[cyan!60] &
  \begin{tabular}[c]{@{}l@{}}uncertainty in the\\ estimated\\ reputation values\\ of worker nodes\end{tabular} &
  \begin{tabular}[c]{@{}l@{}}Improved\\ reliability of\\ worker nodes\end{tabular} &
  \begin{tabular}[c]{@{}c@{}}MNIST\cite{d3},\\ FMNIST\end{tabular} \\
\begin{tabular}[c]{@{}c@{}}RRAFL\\ \cite{cs6}\end{tabular} &
  2021 &
  \begin{tabular}[c]{@{}c@{}}Reputation and\\ Reverse\\ auction\end{tabular} &
  \begin{tabular}[c]{@{}l@{}}A horizontal FL incentive mechanism, which\\ combines reputation and reverse auction theory\end{tabular} &
  \quad\circledmark[cyan!60] &
  \quad\circledmark[cyan!60] &
   &
  \quad\circledmark[cyan!60] &
  \begin{tabular}[c]{@{}l@{}}Low data quality\\ and low quality\\ participants\end{tabular} &
  \begin{tabular}[c]{@{}l@{}}Computational efficiency,\\ Individual\\ rationality, budget\\ feasibility, and\\ truthfulness\end{tabular} &
  \begin{tabular}[c]{@{}c@{}}MNIST, \\ FMNIST, \\ IMDB \cite{d4}\end{tabular} \\
\begin{tabular}[c]{@{}c@{}}RFFL\\ \cite{cs7}\end{tabular} &
  2020 &
  \begin{tabular}[c]{@{}c@{}}Reputation\\ Aware\end{tabular} &
  \begin{tabular}[c]{@{}l@{}}A Framework relies on a reputation mechanism,\\ which evaluates the contributions of each\\ participant by comparing their uploaded gradients\\ to the aggregated global gradients\end{tabular} &
  \quad\circledmark[cyan!60] &
  \quad\circledmark[cyan!60] &
  \quad\circledmark[cyan!60] &
  \quad\circledmark[cyan!60] &
  \begin{tabular}[c]{@{}l@{}}Targeted and un\\ targeted poisoning\\ attack, free rider\\ problem\end{tabular} &
  \begin{tabular}[c]{@{}l@{}}High fairness,\\ robustness,\\ competitive\\ predictive\\ accuracy\end{tabular} &
  \begin{tabular}[c]{@{}c@{}}CIFAR-10, \\ MNIST\end{tabular} \\
\begin{tabular}[c]{@{}c@{}}MarS-FL\\ \cite{cs8}\end{tabular} &
  2022 &
  \begin{tabular}[c]{@{}c@{}}Market Share\\ Mechanism\end{tabular} &
  \begin{tabular}[c]{@{}l@{}}MarS-FL introduces two key measures, stable\\ market and friendliness, to evaluate the viability\\ and market acceptability of FL.\end{tabular} &
  \quad\circledmark[cyan!60] &
   &
   &
  \quad\circledmark[cyan!60] &
  \begin{tabular}[c]{@{}l@{}}Performance of\\ ML models to\\ provide services in\\ a competitive\\ market\end{tabular} &
  \begin{tabular}[c]{@{}l@{}}Viability of FL in\\ a wide range of\\ market condition\end{tabular} &
  \begin{tabular}[c]{@{}c@{}}FL-PT \\ local \\ dataset\end{tabular} \\
\begin{tabular}[c]{@{}c@{}}Blockchain based \\ PS\\ \cite{cs9}\end{tabular} &
  2020 &
  Blockchain &
  \begin{tabular}[c]{@{}l@{}}Numerical evaluation to prevent malicious\\ devices from participating in FL and a\\ participant-selection algorithm for FL server to\\ select the most suitable group\end{tabular} &
  \quad\circledmark[cyan!60] &
   &
   &
   &
  \begin{tabular}[c]{@{}l@{}}Malicious attackers,\\ Participant\\ selection\end{tabular} &
  \begin{tabular}[c]{@{}l@{}}Better accuracy,\\ estimated time and\\ convergence\\ speed\end{tabular} &
  Sim \\
\begin{tabular}[c]{@{}c@{}}Fedar\\ \cite{cs10}\end{tabular} &
  2020 &
  \begin{tabular}[c]{@{}c@{}}FL-SDG,\\ FoolsGold \\ \cite{cs11}\end{tabular} &
  \begin{tabular}[c]{@{}l@{}}Introduced a trust score for each FL client DMR\\ (distributed mobile robots), which is updated\\ based on their previous performance and resource\\ availability.\end{tabular} &
  \quad\circledmark[cyan!60] &
   &
   &
  \quad\circledmark[cyan!60] &
  \begin{tabular}[c]{@{}l@{}}Recourse\\ limitation and\\ reliability issues\end{tabular} &
  \begin{tabular}[c]{@{}l@{}}better model\\ accuracy, reduced\\ convergence time\end{tabular} &
  \begin{tabular}[c]{@{}c@{}}MNIST, \\ Digit images \\ using DMR\end{tabular} \\
\begin{tabular}[c]{@{}c@{}}TrustWorket\\ \cite{cs12}\end{tabular} &
  2021 &
  \begin{tabular}[c]{@{}c@{}}Deterministic\\ encryption,\\ secret\\ minimum\\ heapsort\end{tabular} &
  \begin{tabular}[c]{@{}l@{}}In blockchain based TrustWorker, Reputation\\ privacy of workers is protected using a\\ deterministic encryption algorithm and a secret\\ minimum heapsort scheme.\end{tabular} &
  \quad\circledmark[cyan!60] &
   &
   &
  \quad\circledmark[cyan!60] &
  \begin{tabular}[c]{@{}l@{}}Client selection\\ for crowdsensing\end{tabular} &
  \begin{tabular}[c]{@{}l@{}}high sensing\\ services quality,\\ protecting worker\\ reputation privacy\end{tabular} &
  \begin{tabular}[c]{@{}c@{}}distance \\ measurement \\ task (sim)\end{tabular} \\
\begin{tabular}[c]{@{}c@{}}TrustFed\\ \cite{cs14}\end{tabular} &
  2021 &
  \begin{tabular}[c]{@{}c@{}}Ethereum\\ Blockchain\\ Smart\\ contracts\end{tabular} &
  \begin{tabular}[c]{@{}l@{}}The proposed protocol for CDFL detects outliers\\ in the training distributions and removes them\\ before aggregating the model updates ensuring\\ fairness and maintains the reputation of\\ participants using blockchain\end{tabular} &
  \quad\circledmark[cyan!60] &
   &
   &
  \quad\circledmark[cyan!60] &
  \begin{tabular}[c]{@{}l@{}}model poisoning\\ attacks and the\\ lack of fairness in\\ the training\\ process\end{tabular} &
  \begin{tabular}[c]{@{}l@{}}detecting the\\ outliers within the\\ FL, Fairness\end{tabular} &
  \begin{tabular}[c]{@{}c@{}}Turbofan \\ Engine \\ Degradation \\ (sim)\end{tabular} \\
\begin{tabular}[c]{@{}c@{}}Reputation based \\ MSL\\ \cite{cs15}\end{tabular} &
  2020 &
  \begin{tabular}[c]{@{}c@{}}multi-weight\\ subjective\\ logic model\end{tabular} &
  \begin{tabular}[c]{@{}l@{}}Reputation used as a reliable metric to select\\ trusted workers. To ensure secure and\\ decentralized reputation management,\\ consortium blockchain technology deployed at\\ edge nodes.\end{tabular} &
  \quad\circledmark[cyan!60] &
   &
  \quad\circledmark[cyan!60] &
  \quad\circledmark[cyan!60] &
  \begin{tabular}[c]{@{}l@{}}unreliable data\\ and clients\end{tabular} &
  High accuracy &
  MNIST \\
\begin{tabular}[c]{@{}c@{}}A multi-domain \\ DDoS detection\\ \cite{cs16}\end{tabular} &
  2022 &
  HomoCNN &
  \begin{tabular}[c]{@{}l@{}}Based on interaction reputation, data reputation,\\ and resource reputation, trusted participants can\\ be selected, and malicious participants can be\\ identified.\end{tabular} &
  \quad\circledmark[cyan!60] &
   &
   &
  \quad\circledmark[cyan!60] &
  DDoS attack &
  High accuracy &
  \begin{tabular}[c]{@{}c@{}}Simulated \\ using \\ CICFlowMeter\end{tabular} \\
\begin{tabular}[c]{@{}c@{}}TrustRE\\ \cite{cs17}\end{tabular} &
  2021 &
  \begin{tabular}[c]{@{}c@{}}Proof of\\ Reputation\\ (poR)\end{tabular} &
  \begin{tabular}[c]{@{}l@{}}Use quality of model training as the basis of\\ reputation evaluation and store the reputation\\ values on a blockchain.  consensus algorithm\\ used to aggregate FL models.\end{tabular} &
  \quad\circledmark[cyan!60] &
   &
   &
  \quad\circledmark[cyan!60] &
  \begin{tabular}[c]{@{}l@{}}Subjective\\ judgmental factors\\ and unfair profit\\ distribution in\\ client selection\end{tabular} &
  \begin{tabular}[c]{@{}l@{}}prevent the\\ privacy leakage,\\ enhance the\\ mutual trust,\\ better prediction\end{tabular} &
  \begin{tabular}[c]{@{}c@{}}MNIST, \\ CIFAR-10\end{tabular} \\
\begin{tabular}[c]{@{}c@{}}RCSRM \\ \cite{cs18}\end{tabular} &
  2021 &
  \begin{tabular}[c]{@{}c@{}}DRL-D3QN,\\ CNN\end{tabular} &
  \begin{tabular}[c]{@{}l@{}}Accurately assessing client truthfulness, using\\ reputation mechanism and DRL-powered reverse\\ auction. Improves efficiency and economic\\ benefits.\end{tabular} &
  \quad\circledmark[cyan!60] &
  \quad\circledmark[cyan!60] &
   &
  \quad\circledmark[cyan!60] &
  \begin{tabular}[c]{@{}l@{}}Malicious\\ behavior of\\ Clients\end{tabular} &
  \begin{tabular}[c]{@{}l@{}}Improves\\ economic benefit\\ and accuracy\end{tabular} &
  MNIST \\ \\ 
\multicolumn{11}{l}{\textit{*TR: Trust, CE: Contribution Evaluation, IM: Incentive Mechanism, RP: Reputation}} \\ \bottomrule
\end{tabular}%
}
\end{table*}

In \cite{cs4}, the authors address several limitations of traditional FL that hinder its use in untrusted environments. These limitations include low motivation among clients, high rate of client dropouts, potential model poisoning and stealing, and unauthorized access to the model. To overcome these issues, the authors propose a blockchain-based FL aggregation protocol that is divided into seven stages. This protocol has been implemented in the Ethereum smart contract and is designed to incentivize clients to participate in the training process. The proposed deposit mechanism differentiates between trusted and untrusted clients by imposing different deposit requirements. Trusted clients are required to pay lower deposits compared to untrusted clients, who are required to pay up to three times more. The authors believe that this approach will encourage clients to be more responsible and trustworthy in their participation in the FL process, ultimately leading to improved FL performance and security.

The authors in \cite{cs9} aims to address the challenges of secure and efficient participant selection in FL. The proposed blockchain enabled FL architecture consists of two phases. The first phase is a numerical evaluation, which serves to prevent malicious devices from participating in FL. The second phase involves a participant-selection algorithm that allows the FL server to select the most suitable group of devices for each round of FL training. The numerical evaluation and participant-selection algorithm work together to ensure that only trustworthy devices participate in FL, thereby mitigating the risk of malicious attacks and ensuring the privacy and security of the training data.

The research work in \cite{cs14}presents the TrustFed framework, a decentralized and trustworthy framework for Crowdsourced FL (CDFL) systems. The use of CDFL in ML has been hindered by issues such as model poisoning attacks and the lack of fairness in the training process. The TrustFed framework addresses these issues by incorporating blockchain technology and smart contracts. CDFL removes detected outliers during training distributions before aggregating the model updates, thereby ensuring fairness in the training process. Additionally, the framework maintains the reputation of participating devices by recording their activities on the blockchain, encouraging honest model contributions. In conclusion, the TrustFed framework offers a solution to the problems of fairness and trust in CDFL systems, ensuring that the training process is secure and reliable.

\subsubsection{Contextual Optimization based Trustworthy Client Selection }
The authors in \cite{cs1} studies client selection in FL with a focus on minimizing the average model exchange time while satisfying long-term fairness and system constraints. The authors transform the offline problem into an online Lyapunov optimization problem and quantify the long-term fairness of client participation rate using dynamic queues. They propose a Contextual Combinatorial Multi-Arm Bandit (C2MAB) model for estimating model exchange time based on client contextual properties and historical performance. The proposed fairness-guaranteed selection algorithm, RBCS-F, efficiently solves the optimization problem in FL. Theoretical evaluation and real-data experiments show that RBCS-F can ensure long-term fairness while improving training efficiency while maintaining close accuracy to the random client selection scheme of FL.

A risk-aware Stochastic Integer Programming (SIP) based Client Selection method (SCS) for FL has been proposed in \cite{cs5}, to tackle uncertainty in worker nodes' reputation values. This method aims to minimize operational costs while considering reputation uncertainty. SCS selects worker nodes from both high and lower reputation pools, reducing herding issues. SIP addresses uncertainty in worker node reliability, and SCS combines SIP with reputation to select reliable nodes cost-effectively. Defence mechanisms like Foolsgold can be employed to protect FL models from damage, particularly near convergence.

The research work in \cite{cs8} proposes the MarS-FL framework, a decision support framework for participation in FL, based on market share. MarS-FL introduces two key measures, stable market and friendliness, to evaluate the viability and market acceptability of FL. Using game theoretic tools, the framework predicts the optimal strategies of FL participants and quantifies the impact of market conditions on the final model performance. The results show that the FL performance improvement is bounded by the market conditions, and the friendliness of the market conditions to FL can be quantified using k. The MarS-FL framework provides a systematic approach for evaluating and optimizing FL participation.

The proposed mechanism in \cite{cs18} aims to enhance the quality of FL by accurately assessing the truthfulness and reliability of clients, and selecting the best set of clients that maximize the social surplus in the FL service trading market. The reputation mechanism evaluates clients based on multiple reputation evaluations and is designed to detect malicious or negative behaviors during the FL training process. The mechanism also employs a reverse auction, powered by the deep reinforcement learning (DRL) algorithm D3QN, to select the optimal clients while satisfying weak budget balance, incentive compatibility, and individual rationality. The results of the proposed incentive mechanism demonstrate improved efficiency and economic benefit for FL tasks, as it motivates more number clients having data with high-quality and high reputations to participate in FL at lower costs.
Table 2 provides an extensive overview of the research conducted on trustworthy client selection in FL. This table highlights various methods, techniques, and approaches employed to ensure trustworthiness during the client selection process.

\subsection{Trustworthy Contribution Evaluation}
The contribution measurement strategies used in FL systems can be classified into three main categories: self-report based evaluation, Shapley value based evaluation, and influence and reputation based evaluation. Self-report based evaluation involves the data owner directly reporting their level of contribution to the model owner. Metrics used to evaluate the contribution include computational resources and data size. The Shapley value based evaluation takes into account the participation order of the data owners and assesses their contributions in a fair manner, typically used in cooperative games. Influence and reputation based evaluation considers the client's impact on the FL model's loss function and their reliability, and may involve techniques like Fed-Influence and reputation mechanisms that can be combined with blockchain. A client's reputation is generally a combination of their internal reputation in a task and their accumulated reputation based on historical records. The authors in \cite{ce1} focuses on the critical aspect of measuring the contributions of users in FL systems. The authors conduct a comprehensive analysis of the different strategies for evaluating user contributions, examining the factors that can impact the effectiveness of these methods. The paper emphasizes the need for fair and accurate evaluation techniques to assess the contributions of data owners in FL and provides valuable insights into the current state of research in this area. This study highlights the importance of considering user contributions in FL systems and provides a solid foundation for future research in this field. The categorization and most relevant research work are presented in following sub sections.
\begin{table*}[]
\caption{Trustworthy Contribution Evaluation in FL}
\label{tab:my-table3}
\resizebox{\textwidth}{!}{%
\begin{tabular}{@{}lcclllllllll@{}}
\toprule
\multicolumn{1}{c}{\textbf{\begin{tabular}[c]{@{}c@{}}Proposed \\ Technique\end{tabular}}} &
  \textbf{YoP} &
  \textbf{\begin{tabular}[c]{@{}c@{}}Accompanied \\ Algorithm\end{tabular}} &
  \multicolumn{1}{c}{\textbf{Proposed Methodology}} &
  \multicolumn{1}{c}{\textbf{TR}} &
  \multicolumn{1}{c}{\textbf{IM}} &
  \multicolumn{1}{c}{\textbf{CS}} &
  \multicolumn{1}{c}{\textbf{RP}} &
  \multicolumn{1}{c}{\textbf{\begin{tabular}[c]{@{}c@{}}AD\\ VD\end{tabular}}} &
  \multicolumn{1}{c}{\textbf{\begin{tabular}[c]{@{}c@{}}Problem \\ Encountered\end{tabular}}} &
  \multicolumn{1}{c}{\textbf{Benefits}} &
  \multicolumn{1}{c}{\textbf{Data Set}} \\ \midrule
\begin{tabular}[c]{@{}l@{}}Comparativ\\ study for CE\\ \cite{ce1}\end{tabular} &
  2020 &
  \begin{tabular}[c]{@{}c@{}}VGG-type\\ CNN\end{tabular} &
  \begin{tabular}[c]{@{}l@{}}Examining the factors that can impact the\\ effectiveness of evaluating user contributions\end{tabular} &
  \circledmark &
  \circledmark &
  \circledmark &
  \circledmark &
  \multicolumn{1}{c}{} &
  \begin{tabular}[c]{@{}l@{}}Low quality FL\\ models from\\ malicious\\ Clients\end{tabular} &
  \begin{tabular}[c]{@{}l@{}}Computational\\ Efficiency\\ And\\ effectiveness\end{tabular} &
  CIFAR10 \\
\begin{tabular}[c]{@{}l@{}}BC based\\ PoIS\\ \cite{ce2}\end{tabular} &
  2022 &
  \begin{tabular}[c]{@{}c@{}}PoIS\\ Consensus,\\ Shapely value\end{tabular} &
  \begin{tabular}[c]{@{}l@{}}The framework employs a new consensus\\ mechanism, which evaluates the contributions of\\ individual clients by analyzing model\\ interpretation results.\end{tabular} &
  \circledmark &
  \circledmark &
  \multicolumn{1}{c}{} &
  \circledmark &
  \multicolumn{1}{c}{} &
  \begin{tabular}[c]{@{}l@{}}Unfair\\ Contributions\\ and malicious\\ participants\end{tabular} &
  \begin{tabular}[c]{@{}l@{}}Provides\\ guarantee for\\ byzantine\\ robust\\ aggregation\end{tabular} &
  \begin{tabular}[c]{@{}l@{}}MNIST, \\ FMNIST, \\ CIFAR10\end{tabular} \\
\begin{tabular}[c]{@{}l@{}}CSVES\\ \cite{ce3}\end{tabular} &
  2019 &
  \begin{tabular}[c]{@{}c@{}}EoS\\ Blockchain,\\ Smart Contract\end{tabular} &
  \begin{tabular}[c]{@{}l@{}}This scheme tailors the validation set to a\\ device's data breakdown and checks the validity\\ of the data cost claimed by the device\end{tabular} &
  \circledmark &
  \circledmark &
   &
   &
  \circledmark &
  \begin{tabular}[c]{@{}l@{}}Data validity\\ and quality\end{tabular} &
  \begin{tabular}[c]{@{}l@{}}Accuracy of\\ the model but\\ overfitting\end{tabular} &
  MNIST \\
\begin{tabular}[c]{@{}l@{}}SMPC based\\ ensemble FL\\ \cite{ce4}\end{tabular} &
  2020 &
  \begin{tabular}[c]{@{}c@{}}SMPC protocol,\\ CNN\end{tabular} &
  \begin{tabular}[c]{@{}l@{}}Proposed architecture prioritizes general model\\ evaluation over incentivization for fair\\ contribution assessment.\end{tabular} &
  \circledmark &
   &
   &
   &
  \circledmark &
  \begin{tabular}[c]{@{}l@{}}Data integrity,\\ Model\\ versioning and\\ privacy\\ preservation\end{tabular} &
  \begin{tabular}[c]{@{}l@{}}Improved\\ Privacy and\\ security of\\ medical data\end{tabular} &
  \begin{tabular}[c]{@{}l@{}}2D Colon \\ Pathology \cite{d5}, \\ Breast Tumor \cite{d6}\end{tabular} \\
\begin{tabular}[c]{@{}l@{}}FPPDL\\ \cite{ce5}\end{tabular} &
  2020 &
  \begin{tabular}[c]{@{}c@{}}Three layered\\ onion styled\\ encryption, HE\end{tabular} &
  \begin{tabular}[c]{@{}l@{}}The proposed framework is designed to create a\\ healthy FL ecosystem where each participant's\\ contribution is valued and accurately reflected in\\ the final FL model.\end{tabular} &
  \circledmark &
   &
   &
   &
  \circledmark &
  \begin{tabular}[c]{@{}l@{}}Unfair FL\\ Model\\ evaluation\end{tabular} &
  \begin{tabular}[c]{@{}l@{}}Guaranteed\\ fairness,\\ privacy and\\ accuracy\end{tabular} &
  \begin{tabular}[c]{@{}l@{}}SVHN \cite{d7},\\ MNIST\end{tabular} \\
\begin{tabular}[c]{@{}l@{}}BESIFL\\ \cite{ce6}\end{tabular} &
  2021 &
  \begin{tabular}[c]{@{}c@{}}Consensus\\ Algorithm\end{tabular} &
  \begin{tabular}[c]{@{}l@{}}Three essential components are there in system:\\ an accuracy-based malicious node detection\\ mechanism, a contribution-based incentive\\ mechanism, and an algorithm having token\\ based incentive mechanism\end{tabular} &
  \circledmark &
  \circledmark &
   &
   &
   &
  \begin{tabular}[c]{@{}l@{}}Malicious\\ nodes and\\ unfair\\ contribution\\ evaluation\end{tabular} &
  \begin{tabular}[c]{@{}l@{}}Reduced\\ Training\\ rounds,\\ increased\\ accuracy\end{tabular} &
  MNIST \\
\begin{tabular}[c]{@{}l@{}}CAreFL\\ \cite{ce7}\end{tabular} &
  2022 &
  Shapely Value &
  \begin{tabular}[c]{@{}l@{}}Model aggregation approach in framework\\ selects the best performing sub-model for the\\ next round.  historical contribution evaluation\\ records are further converted into reputation\\ values for the FL participants\end{tabular} &
  \circledmark &
   &
   &
  \circledmark &
   &
  heterogeneity &
  \begin{tabular}[c]{@{}l@{}}Improved\\ Fairness and\\ privacy and\\ accuracy\end{tabular} &
  CIFAR10 \\
\begin{tabular}[c]{@{}l@{}}BFL\\ \cite{ce7}\end{tabular} &
  2022 &
  \begin{tabular}[c]{@{}c@{}}Smart\\ contract\end{tabular} &
  \begin{tabular}[c]{@{}l@{}}Reputation mechanism evaluates model quality\\ and contribution, which encourages data owners\\ to join the BFL and contribute high-quality data\\ for local and weighted global model aggregation\\ and reward allocation.\end{tabular} &
  \circledmark &
   &
   &
  \circledmark &
   &
  \begin{tabular}[c]{@{}l@{}}High quality\\ data and\\ participation\end{tabular} &
  \begin{tabular}[c]{@{}l@{}}High quality\\ model, Ideal\\ accuracy\end{tabular} &
  \begin{tabular}[c]{@{}l@{}}MNSIT,\\ CIFAR10\end{tabular} \\ \\
\multicolumn{11}{l}{\textit{*TR: Trust, CS: Client Selection, IM: Incentive Mechanism, RP: Reputation, AD/VD: Auditibility and Validity}} \\ \bottomrule
\end{tabular}%
}
\end{table*}

\subsubsection{Smart Contract based Trustworthy Contribution Evaluation}

The authors in \cite{ce3} proposes a solution to the issue of data validity and quality in FL systems by utilizing the EOS blockchain and IPFS. The system records uploaded updates in a scalable manner and rewards users based on the cost of their training data. To ensure that only valuable updates are validated and rewarded, the authors propose the Class-Sampled Validation Error Scheme (CSVES). This scheme tailors the validation set to a device's data breakdown and checks the validity of the data cost claimed by the device. By implementing CSVES, the system can ensure that the data quality is maintained while incentivizing users to contribute high-quality updates to the FL model.

A BC-based FL scheme with a reputation mechanism to motivate data owners in high-quality data contribution has been proposed in \cite{ce8}. By integrating smart contract technology and blockchain, the authors aim to create a decentralized and trustworthy environment for conducting FL tasks transparently and fairly. The proposed reputation mechanism evaluates model quality and contribution, which encourages data owners to join the BFL and contribute high-quality data for local and weighted global model aggregation and reward allocation. The noncooperative game theory with equilibrium point make sure the highest rewrard to the contributer with high quality data. Moreover, an optional grouping mechanism is proposed to address the high complexity of a large number of participants. The BFL mechanism is expected to improve the credibility and reliability of FL and guarantee the model quality.
\subsubsection{Shapely Value and Consensus based Trustworthy Contribution Evaluation}
In \cite{ce2}, a blockchain-assisted FL framework is presented to encourage honest participation and reward fair contributions. The framework employs a new consensus mechanism known as "Proof of Interpretation and Selection" (PoIS), which evaluates the contributions of individual clients by analyzing model interpretation results. PoIS aggregates feature attributions to distinguish prominent contributors and outliers and uses a credit function that considers contribution, relevance, and past performance to determine incentives. The proposed framework has been tested against various types of adversaries, datasets, and attack strategies and found to be robust. This framework offers a promising solution for addressing the issues of traditional FL and ensuring fairness in contribution-based incentivization.

The authors propose a decentralized FL system named BESIFL (Blockchain Empowered Secure and Incentive FL) that leverages blockchain technology to remove the central FL server in \cite{ce6}. Three essential components are there in system: an accuracy-based malicious node detection mechanism, a contribution-based incentive mechanism, and an algorithm that coordinates both mechanisms. The malicious node detection mechanism identifies and removes malicious nodes during the training process, while the contribution-based incentive mechanism motivates nodes to participate in FL and rewards them based on their contribution to model training. BESIFL is designed to address the security and incentive issues in FL while also stimulating credible nodes to contribute to the model training. The proposed system applies the consensus algorithm and identity authentication of blockchain to ensure the security of model training and employs mechanisms for accuracy-based malicious node detection, contribution-based node selection, and token-based node incentive.

The varying data quality and heterogeneous data distributions across multiple healthcare institutions pose significant challenges to existing FL frameworks. To address these issues, a Contribution-Aware FL (CAreFL) framework \cite{ce7}, has been proposed, which focuses on fair and efficient contribution evaluation of FL participants. The proposed GTG-Shapley approach allows for fast and accurate evaluation of participant contributions. Moreover, the framework introduces a novel FL model aggregation approach, which selects the best performing sub-model for the next round of local training instead of always aggregating all received local models. This approach addresses the heterogeneity issues of data distribution in the healthcare sector. The historical contribution evaluation records are further converted into reputation values for the FL participants, which can serve as a basis for stakeholder management decision support. The CAreFL framework offers a promising solution to the challenges faced by existing FL frameworks in the healthcare sector, improving FL model performance while ensuring privacy and fairness.

\subsubsection{Privacy Preserving Trustworthy Contribution Evaluation}

Authors presents a novel architecture for healthcare institutions to collaborate in improving the performance of a global ML model in \cite{ce4}. The proposed system utilizes a combination of blockchain and secure multi-party computation (SMPC) to ensure data integrity, model versioning and privacy preservation during model training and ensemble. Unlike traditional methods, proposed architecture prioritizes general model evaluation over incentivization for fair contribution assessment. This evaluation process is carried out by the blockchain nodes and recorded on tamper-proof storage. The final contribution of each participant's ML model is determined based on their performance on unforeseen data. Additionally, the architecture enables each participant to define their own model structure, taking into account the varying computing power among participants. The proposed hierarchical ensemble FL method promises to advance the field of collaborative ML in healthcare while maintaining the privacy and security of sensitive medical data.

The authors present a decentralized Fair and Privacy-Preserving Deep Learning (FPPDL) framework in \cite{ce5}, that aims to address the issue of fairness in FL models. Unlike traditional FL solutions that provide all parties with the same model regardless of their contribution, FPPDL aims to provide each participant with a final FL model that reflects their individual contributions. To achieve fairness, the authors propose a local credibility mutual evaluation mechanism, and for privacy preservation, a three-layer onion-style encryption scheme is proposed. The framework operates by recording all transactions, including uploading and downloading, through blockchain technology. This eliminates the need for participants to trust each other or a third party. The proposed framework is designed to create a healthy FL ecosystem where each participant's contribution is valued and accurately reflected in the final FL model.

Table 3 provides an extensive overview of the research conducted on trustworthy contribution evaluation in FL. 

\subsection{Trustworthy Incentive Mechanism}
In this section, we categorize trustworthy incentive mechanism algorithms and methodologies into four sub-categories based on their primary objectives: Game theory, Reputation based, Contribution based and Privacy and Security based Trustworthy Incentive Mechanisms.

\subsubsection{Game Theory based Trustworthy Incentive Mechanism}

In \cite{im4}, the authors propose a novel approach to designing incentives for a blockchain-enabled FL platform using mechanism design, an economic approach to realizing desired objectives in situations where participants act rationally. The main idea behind the incentive mechanism is to introduce a repeated competition for model updates, so that any rational worker follows the protocol and maximizes their profits. During each round, selected workers choose the best k model updates from previous round and update their own model based on them. The reward to workers in the previous round is decided by the vote of the next round workers. The model updates of the next round workers are also competed and voted by workers in the subsequent round, ensuring that they cannot sabotage the system. The authors provide a rigorous theoretical analysis of the incentive compatibility based on contest theory and clarify the optimal conditions for reward policy in a blockchain-enabled FL platform. The contribution of the paper includes a competitive incentive mechanism design, a full-fledged protocol that can be implemented on existing public blockchains, and a theoretical analysis to clarify incentive compatibility based on contest theory.

Authors proposed a novel FL incentive mechanism called Fair-VCG (FVCG) in \cite{im11}, which is based on the well-known Vickrey-Clarke-Groves (VCG) mechanism \cite{im12}. FVCG incentivizes fair treatment of FL participants and can easily be integrated into existing FL platforms as a standard module. Our mechanism aims to optimally share revenues with data owners while encouraging full data contribution and truthful cost reporting. Through the use of a neural network method, FVCG optimizes for social surplus and minimizes unfairness, making it individually rational and weakly budget balanced. The practical applications of FVCG's economic concepts make it a promising solution for ensuring fairness in FL.

InFEDge \cite{im17}, is a blockchain-based incentive mechanism proposed to address challenges related to multi-dimensional individual properties, incomplete information, and unreliable participants in edge computing. The mechanism models rationality of clients, edge servers, and the cloud, and proposes a hierarchical contract solution to obtain the optimal solution under incomplete information. The existence and uniqueness of Nash equilibrium is proved with closed-form solution under complete information. Blockchain is introduced to implement the incentive mechanism in the smart contract and provides a credible, faster, and transparent environment for the system. InFEDge ensures privacy, prevents unreliable participants' disturbance, and provides a credible and transparent environment to effectively manage the incentive mechanism.

Another incentive mechanism to encourage mobile users to participate in FL is presented in \cite{im18}. The winner selection problem in the auction game is formulated as a social welfare maximization problem and solved with a primal-dual greedy algorithm. The proposed auction mechanism is guaranteed to be truthful, individually rational, and computationally efficient. The wireless resource limitation makes the winner selection problem an NP-hard problem. The critical value-based payment is proposed to deal with the NP-hard problem of selecting the winning users.

A Social Federated Edge Learning (SFEL) framework is proposed for wireless networks, which leverages social relationships to address the trustworthiness and incentive mechanisms in Federated Edge Learning (FEL) \cite{im20}. The SFEL framework uses a social graph model to find trustworthy learning partners with similar learning task interests, and a social-effect-based incentive mechanism to encourage better personal learning behaviors. The proposed incentive mechanism is a Stackelberg game-based model, which can handle both complete and incomplete information to encourage active participation from learners. BFEL solves the issue of malicious or inactive learners that result in low quality and untrustworthy model updates, which undermine the viability and stability of FEL.

The authors presents FAIR \cite{im23}, a distributed learning system that integrates three major technical components to ensure quality-aware FL. The system estimates individual learning quality to provide precise user incentives and model aggregation. It also allocates learning tasks and corresponding payments, and conducts model aggregation in real-time. The quality estimation is done using loss reduction during the learning process and leveraging historical quality records. The system uses a reverse auction case to motivate user participation, where mobile users submit their bids and the platform acts as the auctioneer. A greedy algorithm determines learning task allocation and reward distribution based on Myerson’s theorem to maximize collective learning quality within the recruiting budget. The model aggregation algorithm integrates model quality and filters non-ideal model updates to enhance the global learning model. The proposed FAIR system is truthful, individually rational, and computationally efficient. The paper's contributions lie in providing a system that ensures quality-aware FL, which is crucial in practical distributed learning scenarios but rarely seen in the literature.

Ensuring stable user participation necessitates a robust, fair, and reciprocal incentive mechanism. FedAB in \cite{im24}, a groundbreaking incentive strategy and client selection method based on a multi-attribute reverse auction mechanism and a combinatorial multi-armed bandit (CMAB) algorithm. FedAB contributes a local contribution evaluation technique tailored for FL, a payment mechanism fostering individual rationality and truthfulness, and a UCB-based winner selection algorithm to maximize utility, fairness, and reciprocity.

\subsubsection{Reputation based Trustworthy Incentive Mechanism}

The proposed SRB\-FL framework \cite{im1} is designed to address the challenges faced by FL and provide a secure and trustworthy solution. The framework focuses on improving the reliability of FL devices through an incentive mechanism that uses subjective multi-weight logic. This mechanism provides a reputation mechanism that incites FL devices to provide reliable model updates. The results show that the framework is both efficient and scalable, making it a promising solution for FL. The framework also uses blockchain sharding to ensure data reliability, scalability, and trustworthiness, enabling parallel model training and improving the scalability of blockchain FL. lightweight sharding consensus (LWSC), and secure sharding using subjective logic (SSSL) is also used to improve the reliability and security of proposed mechanism respectively.

FRFL scheme in \cite{im2} addresses the challenges of FL in battery-constrained UAVs for UAV-supported crowdsensing. The scheme focuses on fairness and robustness, utilizing an optimal contract theory-based incentive mechanism that fosters UAV participation under information asymmetry, with proven truthfulness, contractual feasibility, and computational efficiency. Additionally, the FRFL method leverages edge computing-powered 5G heterogeneous networks for high-speed, low-latency FL services. It employs Byzantine-robust aggregation rules, equitable model profit distribution, and a reputation mechanism to recruit reliable UAVs while deterring free-riding. Simulations confirm the FRFL scheme's efficacy in user utility, communication efficiency, and robustness, highlighting the importance of the incentive mechanism for fostering fair and robust FL in UAV-assisted crowdsensing.

The adoption of FL has been limited by certain challenges, including the lack of a comprehensive incentive mechanism for worker (mobile device) participation and the lack of a reliable worker selection method. To tackle these challenges, authors presents a novel approach that combines reputation and contract theory to incentivize high-reputation mobile devices to participate in the model training process in \cite{im3}. Firstly, we introduce reputation as a metric to evaluate the reliability and trustworthiness of the mobile devices. Secondly, a multi-weight subjective logic model is used to determine the worker's reputation, which is securely managed through the consortium blockchain. Incentivizing participation is achieved through a contract theory-based mechanism, which stimulates workers with high-quality data to join the learning process. The mechanism provides an incentive for high-reputation workers to maintain the accuracy and reliability of their local training data. Additionally, the subjective logic model is used to generate composite reputation values for worker candidates, allowing for the selection of credible and trustworthy participants. This approach introduces third-party miners which can lead to model leakage. The blockchain also has scalability issues that could increase communication delay for FL.The authors aim to create an incentive-aware platform that ensures the participation of devices in model training while taking into account the communication delays caused by the blockchain's scalability problems.

The authors in \cite{im5} propose a reputation-based selection methodology and an auction-driven incentive scheme to improve the participation of data owners while maintaining the desired level of platform utility and aggregated model performance. The reputation score is based on the performance of the local models, and the incentive mechanism is designed to reward participants fairly for their contributions. The compensation of the users is dynamically adjusted to distribute the benefits more fairly, while ensuring positive user utility and maintaining the aggregated model accuracy. Building on the Trustworthy Sensing for Crowd Management (TSCM) approach discused in \cite{im6, im7}, the authors propose a reputation score-based incentive scheme that assigns higher rewards to data owners with higher quality data and better-performing local models. The scheme adopts a reverse auction procedure to adjust the compensation of the users dynamically. Numerical evaluations have shown that the proposed scheme can improve the user utility while maintaining the platform utility and test accuracy of the global FL model. The ongoing study involves adjusting the frequency of local updates to improve efficiency under limited resources and implementing the adjustable FL algorithm during the training process to improve model performance.

A Blockchain Empowered Secure and Incentive FL (BESIFL) paradigm is proposed in \cite{im19}, to enhance security and performance in FL. The proposed BESIFL system is fully decentralized, leveraging blockchain technology to enable effective mechanisms for malicious node detection and incentive management. An accuracy-based malicious node detection mechanism is developed to identify and remove malicious nodes, while a contribution-based incentive mechanism with a token-based reward scheme motivates credible nodes to participate in the learning process. An algorithm is designed to coordinate the mechanisms for malicious node detection and contributing node incentive/selection, enabling the BESIFL system to efficiently address these issues. The proposed paradigm presents an innovative approach to enhancing the security and performance of FL.
\begin{table*}[]
\caption{Trustworthy Incentive Mechanism in FL}
\label{tab:my-table4}
\resizebox{\textwidth}{!}{%
\begin{tabular}{@{}ccclcccccllcc@{}}
\toprule
\textbf{\begin{tabular}[c]{@{}c@{}}Proposed \\ Technique\end{tabular}} &
  \textbf{YoP} &
  \textbf{\begin{tabular}[c]{@{}c@{}}Accompanied \\ Algorithm\end{tabular}} &
  \multicolumn{1}{c}{\textbf{Description}} &
  \textbf{TR} &
  \textbf{CE} &
  \textbf{CS} &
  \textbf{RP} &
  \textbf{AG} &
  \multicolumn{1}{c}{\textbf{\begin{tabular}[c]{@{}c@{}}Problem \\ Encountered\end{tabular}}} &
  \multicolumn{1}{c}{\textbf{Outocmes}} &
  \textbf{Data Set} \\ \midrule
\begin{tabular}[c]{@{}c@{}}SRB-FL\\ \cite{im1}\end{tabular} &
  2021 &
  \begin{tabular}[c]{@{}c@{}}Lightweight\\ Sharding\\ Consensus\\ (LWSC),\\ SSSL\end{tabular} &
  \begin{tabular}[c]{@{}l@{}}An incentive mechanism that uses subjective\\ multi-weight logic and a reputation mechanism\\ that incites FL devices for reliable model\\ updates\end{tabular} &
  \quad\circledmark[cyan!60] &
   &
   &
  \quad\circledmark[cyan!60] &
   &
  \begin{tabular}[c]{@{}l@{}}Reliability\\ tractability, and\\ anonymity\end{tabular} &
  \begin{tabular}[c]{@{}l@{}}Ensure data\\ reliability,\\ trustworthiness\\ scalability\end{tabular} &
  MNIST \\
\begin{tabular}[c]{@{}c@{}}FRFL\\ \cite{im2}\end{tabular} &
  2021 &
  \begin{tabular}[c]{@{}c@{}}Contract\\ theory\end{tabular} &
  \begin{tabular}[c]{@{}l@{}}A fair incentive reputation mechanism is\\ employed to recruit credible UAVs and\\ prevent free riding\end{tabular} &
  \quad\circledmark[cyan!60] &
   &
   &
   &
  \quad\circledmark[cyan!60] &
  \begin{tabular}[c]{@{}l@{}}heterogeneity of\\ UAVs, free-riders\\ and Byzantine\\ UAVs\end{tabular} &
  \begin{tabular}[c]{@{}l@{}}Improved user\\ utility,\\ communication\\ efficiency, and\\ robustness\end{tabular} &
  MNIST \\
\begin{tabular}[c]{@{}c@{}}RCIM\\  \\ \cite{im3}\end{tabular} &
  2019 &
  \begin{tabular}[c]{@{}c@{}}Contract\\ theory\end{tabular} &
  \begin{tabular}[c]{@{}l@{}}Reputation used to evaluate the reliability and\\ trustworthiness. Multi-weight subjective logic\\ model is used for worker's reputation\end{tabular} &
  \quad\circledmark[cyan!60] &
   &
  \quad\circledmark[cyan!60] &
  \quad\circledmark[cyan!60] &
   &
  \begin{tabular}[c]{@{}l@{}}Unfair incentive\\ mechanism and\\ reliable worker\\ selection\end{tabular} &
  \begin{tabular}[c]{@{}l@{}}Improved\\ accuracy and\\ reliability\end{tabular} &
  MNIST \\
\begin{tabular}[c]{@{}c@{}}IBFL\\ \cite{im4}\end{tabular} &
  2019 &
  \begin{tabular}[c]{@{}c@{}}Mechanism\\ design\end{tabular} &
  \begin{tabular}[c]{@{}l@{}}Mechanism design enabled incentives for a\\ BC-based FL platform, an economic approach\\ in situations where participants act rationally\end{tabular} &
  \quad\circledmark[cyan!60] &
   &
  \quad\circledmark[cyan!60] &
   &
   &
  \begin{tabular}[c]{@{}l@{}}Communication\\ delay\end{tabular} &
  \begin{tabular}[c]{@{}l@{}}Fair incentive\\ mechanism\end{tabular} &
  \begin{tabular}[c]{@{}c@{}}Network\\ Trace\\ CICIDS2017\end{tabular} \\
\begin{tabular}[c]{@{}c@{}}Modified TSCM\\ \cite{im5}\end{tabular} &
  2021 &
  TSCM &
  \begin{tabular}[c]{@{}l@{}}A reputation score-based incentive scheme that\\ assigns higher rewards to data owners with\\ higher quality data and better-performing local\\ models\end{tabular} &
  \quad\circledmark[cyan!60] &
   &
   &
  \quad\circledmark[cyan!60] &
   &
  \begin{tabular}[c]{@{}l@{}}Heterogeneity,\\ quantity and\\ quality of sensed\\ data\end{tabular} &
  \begin{tabular}[c]{@{}l@{}}Improved\\ Participant\\ utility, test\\ accuracy, and \\ model efficiency\end{tabular} &
  MNIST \\
\begin{tabular}[c]{@{}c@{}}Deep Chain\\ \cite{im8}\end{tabular} &
  2019 &
  \begin{tabular}[c]{@{}c@{}}Threshold\\ Paillier\\ algorithm,\\ UVCDN\end{tabular} &
  \begin{tabular}[c]{@{}l@{}}Incentive mechanism in DeepChain consists of\\ a trusted time clock mechanism and a secure\\ monetary penalty mechanism.\end{tabular} &
  \quad\circledmark[cyan!60] &
   &
  \quad\circledmark[cyan!60] &
   &
  \quad\circledmark[cyan!60] &
  \begin{tabular}[c]{@{}l@{}}Security and\\ unfairness\end{tabular} &
  \begin{tabular}[c]{@{}l@{}}Ensure privacy\\ and correctness\\ of training\\ process\end{tabular} &
  MNIST \\
\begin{tabular}[c]{@{}c@{}}RSTM-FL\\  \\ \cite{im10}\end{tabular} &
  2022 &
  \begin{tabular}[c]{@{}c@{}}Shapely\\ value\end{tabular} &
  \begin{tabular}[c]{@{}l@{}}Determine the contributions of FL institutions\\ to bias using Shapley value approximation\\ method\end{tabular} &
  \quad\circledmark[cyan!60] &
  \quad\circledmark[cyan!60] &
   &
   &
   &
  FL model bias &
  \begin{tabular}[c]{@{}l@{}}Better\\ Contribution\\ evaluation,\\ Improved bias\\ handling\end{tabular} &
  \begin{tabular}[c]{@{}c@{}}NIH, CXP,\\ and CXR \cite{d8}\end{tabular} \\
\begin{tabular}[c]{@{}c@{}}Fair-VCG\\  \\ \cite{im11}\end{tabular} &
  2020 &
  \begin{tabular}[c]{@{}c@{}}Vickrey\\ Clarke\\ Groves\\ (VCG)\end{tabular} &
  \begin{tabular}[c]{@{}l@{}}Using a neural network method, FVCG\\ optimizes for social surplus and minimizes\\ unfairness,\end{tabular} &
  \quad\circledmark[cyan!60] &
  \quad\circledmark[cyan!60] &
   &
   &
   &
  \begin{tabular}[c]{@{}l@{}}Functional\\ Optimization\\ problem\end{tabular} &
  \begin{tabular}[c]{@{}l@{}}Minimizing\\ unfairness,\\ maximizes\\ social surplus\end{tabular} &
  Sim \\
\begin{tabular}[c]{@{}c@{}}Fed-Token\\  \\ \cite{im13}\end{tabular} &
  2022 &
  \begin{tabular}[c]{@{}c@{}}Shapely\\ value\end{tabular} &
  \begin{tabular}[c]{@{}l@{}}The proposed tokenized incentive design\\ accommodates clients with diverse profiles,\\ making it a robust solution for collaborative FL\end{tabular} &
  \quad\circledmark[cyan!60] &
  \quad\circledmark[cyan!60] &
   &
   &
  \quad\circledmark[cyan!60] &
  \begin{tabular}[c]{@{}l@{}}Participation of\\ quality clients\end{tabular} &
  \begin{tabular}[c]{@{}l@{}}Communication\\ efficient, better\\ convergence,\\ effective\\ detection of\\ poisoning\\ attack\end{tabular} &
  CIFAR10 \\
\begin{tabular}[c]{@{}c@{}}FIFL\\  \\ \cite{im14}\end{tabular} &
  2021 &
  \begin{tabular}[c]{@{}c@{}}Subjective\\ Logic, Shapely\\ Incentive\end{tabular} &
  \begin{tabular}[c]{@{}l@{}}FIFL adopts a blockchain-based audit method\\ and a reputation-based server selection method\\ to prevent malicious nodes\end{tabular} &
  \quad\circledmark[cyan!60] &
  \quad\circledmark[cyan!60] &
  \quad\circledmark[cyan!60] &
  \quad\circledmark[cyan!60] &
   &
  \begin{tabular}[c]{@{}l@{}}Worker’s utility\\ with high\\ computation\\ overhead, fair-\\ ness and reliability\end{tabular} &
  \begin{tabular}[c]{@{}l@{}}Increased\\ system revenue,\\ Improved\\ reward system\end{tabular} &
  \begin{tabular}[c]{@{}c@{}}MNIST,\\ CIFAR10\end{tabular} \\
\begin{tabular}[c]{@{}c@{}}FGFL\\  \\ \cite{im15}\end{tabular} &
  2022 &
  \begin{tabular}[c]{@{}c@{}}time-decay\\ SLM, LW\\ Gradients\\ similarity\end{tabular} &
  \begin{tabular}[c]{@{}l@{}}Achieves trusted management of incentives\\ through an audit method and a reputation\\ based server selection method\end{tabular} &
  \quad\circledmark[cyan!60] &
  \quad\circledmark[cyan!60] &
   &
  \quad\circledmark[cyan!60] &
   &
  \begin{tabular}[c]{@{}l@{}}stable and\\ efficient training\\ issue, \\ Participation of\\ quality clients\end{tabular} &
  \begin{tabular}[c]{@{}l@{}}Increased\\ system revenue,\\ fair and reliable\\ incentive\\ mechanism and\\ management\end{tabular} &
  \begin{tabular}[c]{@{}c@{}}MNIST, \\ CIFAR10\end{tabular} \\
\begin{tabular}[c]{@{}c@{}}FLI\\ \cite{im16}\end{tabular} &
  2020 &
  \begin{tabular}[c]{@{}c@{}}Payoff\\ Sharing\\ scheme\end{tabular} &
  \begin{tabular}[c]{@{}l@{}}FLI dynamically adjusts data owners' shares to\\ distribute benefits fairly and sacrifice among\\ them\end{tabular} &
  \quad\circledmark[cyan!60] &
  \quad\circledmark[cyan!60] &
   &
   &
   &
  \begin{tabular}[c]{@{}l@{}}Temporary\\ mismatch and cost\\ between\\ contributions and\\ rewards\end{tabular} &
  \begin{tabular}[c]{@{}l@{}}Ensures\\ Contribution\\ fairness, regret\\ distribution\\ fairness\end{tabular} &
  \begin{tabular}[c]{@{}c@{}}Sim-based\\ 6- payoff\\ sharing \\ schemes\end{tabular} \\
\begin{tabular}[c]{@{}c@{}}InFEDge \\ \cite{im17}\end{tabular} &
  2022 &
  \begin{tabular}[c]{@{}c@{}}Deep\\ Reinforcement \\ Learning\\ (DRL)\end{tabular} &
  \begin{tabular}[c]{@{}l@{}}InFEDge, a game-based incentive mechanism\\ in the smart contract and provides a credible,\\ faster, and transparent environment\end{tabular} &
  \quad\circledmark[cyan!60] &
   &
  \quad\circledmark[cyan!60] &
   &
   &
  \begin{tabular}[c]{@{}l@{}}multi-dimensional\\ individual\\ properties,\\ unreliable\\ participants\end{tabular} &
  \begin{tabular}[c]{@{}l@{}}ensures privacy,\\ transparency\\ and prevents\\ from unreliable\\ participants\end{tabular} &
  \begin{tabular}[c]{@{}c@{}}MNIST,\\ FMNIST,\\ CIFAR10\end{tabular} \\
\begin{tabular}[c]{@{}c@{}}PDGA\\  \\ \cite{im18}\end{tabular} &
  2021 &
  \begin{tabular}[c]{@{}c@{}}Winner\\ Selection\\ scheme\end{tabular} &
  \begin{tabular}[c]{@{}l@{}}The critical value-based payment is proposed\\ to deal with the NP-hard problem of selecting\\ the winning users\end{tabular} &
  \quad\circledmark[cyan!60] &
   &
  \quad\circledmark[cyan!60] &
   &
   &
  \begin{tabular}[c]{@{}l@{}}Small data\\ samples, resource\\ and client\\ limitations\end{tabular} &
  \begin{tabular}[c]{@{}l@{}}Improved\\ accuracy and\\ outperformed in\\ social welfare\end{tabular} &
  sim \\
\begin{tabular}[c]{@{}c@{}}BESIFL \\ \cite{im19}\end{tabular} &
  2021 &
  \begin{tabular}[c]{@{}c@{}}Credit based\\ Node\\ selection\end{tabular} &
  \begin{tabular}[c]{@{}l@{}}A credit-based node selection accuracy-based\\ malicious node detection mechanism along\\ with contribution-based incentive mechanism\end{tabular} &
  \quad\circledmark[cyan!60] &
  \quad\circledmark[cyan!60] &
  \quad\circledmark[cyan!60] &
   &
   &
  \begin{tabular}[c]{@{}l@{}}Malicious nodes\\ and contribution\\ evaluation\end{tabular} &
  \begin{tabular}[c]{@{}l@{}}Protection\\ Against\\ Malicious\\ nodes, credible\\ node selection\end{tabular} &
  MNIST \\
\begin{tabular}[c]{@{}c@{}}SFEL \\ \cite{im20}\end{tabular} &
  2021 &
  \begin{tabular}[c]{@{}c@{}}Stackelberg\\ game-based model\end{tabular} &
  \begin{tabular}[c]{@{}l@{}}SFEL uses a social graph model to find\\ trustworthy learning partners with similar\\ learning task interests\end{tabular} &
  \quad\circledmark[cyan!60] &
   &
   &
   &
   &
  \begin{tabular}[c]{@{}l@{}}Low quality\\ learning params,\\ viability and\\ stability\end{tabular} &
  \begin{tabular}[c]{@{}l@{}}Win-win\\ situation,\\ improved\\ accuracy and\\ performance\end{tabular} &
  \begin{tabular}[c]{@{}c@{}}MNIST,\\ CIFAR10\end{tabular} \\
\begin{tabular}[c]{@{}c@{}}FedFAIM \\ \cite{im21}\end{tabular} &
  2022 &
  \begin{tabular}[c]{@{}c@{}}Shapely\\ value\end{tabular} &
  \begin{tabular}[c]{@{}l@{}}Aggregation fairness and reward fairness are\\ achieved through efficient gradient\\ aggregation and shapely value-based\\ contribution assessment\end{tabular} &
  \quad\circledmark[cyan!60] &
  \quad\circledmark[cyan!60] &
   &
  \quad\circledmark[cyan!60] &
    \quad\circledmark[cyan!60]  &
  Unfairness &
  \begin{tabular}[c]{@{}l@{}}Guaranteed\\ reward fairness\end{tabular} &
  \begin{tabular}[c]{@{}c@{}}MNIST,\\ CIFAR10\end{tabular} \\
\begin{tabular}[c]{@{}c@{}}SBFL FL\\ \cite{im22}\end{tabular} &
  2022 &
  \begin{tabular}[c]{@{}c@{}}Local\\ differential privacy (LDP)\end{tabular} &
  \begin{tabular}[c]{@{}l@{}}LDP ensures clients' model updates cannot\\ leak within their private data and non\\ interactive ZKPs ensure the integrity of the FL\\ system\end{tabular} &
  \quad\circledmark[cyan!60] &
   &
   &
   &
  \quad\circledmark[cyan!60] &
  \begin{tabular}[c]{@{}l@{}}Unfair incentive,\\ privacy and\\ integrity\end{tabular} &
  \begin{tabular}[c]{@{}l@{}}Improved trust,\\ Economic\\ incentives, and\\ confidentiality\end{tabular} &
  \begin{tabular}[c]{@{}c@{}}California\\ Housing \cite{d9}\\ prices\end{tabular} \\
\begin{tabular}[c]{@{}c@{}}FAIR\cite{im23}\end{tabular} &
  2021 &
  \begin{tabular}[c]{@{}c@{}}Myerson’s\\ theorem\end{tabular} &
  \begin{tabular}[c]{@{}l@{}}This quality aware system uses a reverse\\ auction case to motivate user participation.\end{tabular} &
  \quad\circledmark[cyan!60] &
   &
   &
   &
  \quad\circledmark[cyan!60] &
  \begin{tabular}[c]{@{}l@{}}Low quality\\ model updates and\\ low-quality client\\ participation\end{tabular} &
  \begin{tabular}[c]{@{}l@{}}truthful,\\ computationally\\ efficient\end{tabular} &
  \begin{tabular}[c]{@{}c@{}}SVHN,\\ MNIST,\\ FMNIST,\\ CIFAR10\end{tabular} \\
\begin{tabular}[c]{@{}c@{}}FedAB\\ \cite{im24}\end{tabular} &
  2023 &
  \begin{tabular}[c]{@{}c@{}}combinatorial multi-armed\\ bandit (CMAB)\end{tabular} &
  \begin{tabular}[c]{@{}l@{}}A groundbreaking incentive strategy and client\\ selection method based on a multi-attribute\\ reverse auction mechanism\end{tabular} &
  \quad\circledmark[cyan!60] &
  \quad\circledmark[cyan!60] &
  \quad\circledmark[cyan!60] &
   &
  \quad\circledmark[cyan!60] &
  Client selection issue &
  \begin{tabular}[c]{@{}l@{}}Fairness,\\ reciprocity and\\ server’s utility\\ maximization\end{tabular} &
  \begin{tabular}[c]{@{}c@{}}MNIST,\\ FMNIST,\\ CIFAR10\end{tabular} \\ \\
\multicolumn{9}{l}{\textit{*TR: Trust, CS: Client Selection, CE: Contribution Evaluation, RP: Reputation, AG: Secure Aggregation}} \\ \bottomrule
\end{tabular}%
}
\end{table*}

\begin{table*}[]
\caption{Trustworthy Audibility and Accountability in FL}
\label{tab:my-table5}
\resizebox{\textwidth}{!}{%
\begin{tabular}{@{}llllccccclll@{}}
\toprule
\multicolumn{1}{c}{\textbf{\begin{tabular}[c]{@{}c@{}}Proposed \\ Technique\end{tabular}}} &
  \multicolumn{1}{c}{\textbf{YoP}} &
  \multicolumn{1}{c}{\textbf{\begin{tabular}[c]{@{}c@{}}Accompanied \\ Algorithm\end{tabular}}} &
  \multicolumn{1}{c}{\textbf{Proposed Methodology}} &
  TR &
  IM &
  CE &
  VF &
  AG &
  \multicolumn{1}{c}{\textbf{Problem Encountered}} &
  \multicolumn{1}{c}{\textbf{Benefits}} &
  \multicolumn{1}{c}{\textbf{Data Set}} \\ \midrule
\begin{tabular}[c]{@{}l@{}}VFChain\\ \cite{ab1}\end{tabular} &
  2021 &
  \begin{tabular}[c]{@{}l@{}}Committee\\ selection \\ Scheme (DSC)\end{tabular} &
  \begin{tabular}[c]{@{}l@{}}Achieves auditability by proposing a novel\\ authenticated data structure and search\\ efficiency with an optimization scheme to \\ support multiple model learning tasks\end{tabular} &
  \circledmark &
  \circledmark &
   &
  \circledmark &
  \circledmark &
  \begin{tabular}[c]{@{}l@{}}Security attacks,\\ correctness of training\end{tabular} &
  \begin{tabular}[c]{@{}l@{}}Improved\\ execution time,\\ accuracy and\\ throughput\end{tabular} &
  MNIST \\
\begin{tabular}[c]{@{}l@{}}BlockFlow\\ \cite{ab2}\end{tabular} &
  2022 &
  \begin{tabular}[c]{@{}l@{}}Ethereum smart\\ contracts\end{tabular} &
  \begin{tabular}[c]{@{}l@{}}It utilizes differential privacy to protect\\ datasets and introduces a unique auditing\\ mechanism for model contribution\end{tabular} &
  \circledmark &
  \circledmark &
  \circledmark &
   &
   &
  \begin{tabular}[c]{@{}l@{}}Malicious\\ agents\end{tabular} &
  \begin{tabular}[c]{@{}l@{}}Guaranteed\\ accountability,\\ privacy and\\ trustworthiness\end{tabular} &
  \begin{tabular}[c]{@{}l@{}}Adult\\ Census\\ Income,\\ KDD\end{tabular} \\
\begin{tabular}[c]{@{}l@{}}FLChain\\ \cite{ab4}\end{tabular} &
  2019 &
  \begin{tabular}[c]{@{}l@{}}Federation\\ Establishment\\ algorithm\end{tabular} &
  \begin{tabular}[c]{@{}l@{}}Honest trainers can receive a fairly\\ partitioned profit from well-trained models\\ based on their contribution, while malicious\\ actors can be detected and heavily punished\end{tabular} &
  \circledmark &
  \circledmark &
  \circledmark &
  \multicolumn{1}{l}{} &
  \multicolumn{1}{l}{} &
  \begin{tabular}[c]{@{}l@{}}Single point of\\ failure,\\ malformed\\ messages\end{tabular} &
  \begin{tabular}[c]{@{}l@{}}Guaranteed\\ Confidentiality\\ and auditability\end{tabular} &
  MNIST \\
\begin{tabular}[c]{@{}l@{}}Blockchain\\ based\\ TrustworthyFL\\ \cite{ab5}\end{tabular} &
  2021 &
  Smart Contract &
  \begin{tabular}[c]{@{}l@{}}A weighted fair training dataset sampler\\ algorithm is introduced to improve fairness\\ affected by data class distribution\\ heterogeneity\end{tabular} &
  \circledmark &
  \multicolumn{1}{l}{} &
  \multicolumn{1}{l}{} &
  \circledmark &
  \multicolumn{1}{l}{} &
  \begin{tabular}[c]{@{}l@{}}Accountability\\ and fairness\\ issue due to\\ heterogeneity\\ and multi\\ stakeholder\\ involvement\end{tabular} &
  \begin{tabular}[c]{@{}l@{}}Improve\\ Accountability,\\ fairness,\\ accuracy, and\\ model’s\\ generalisation\end{tabular} &
  \begin{tabular}[c]{@{}l@{}}COVID-19-\\ Radiography,\\ Chest\\ X-ray\\ Dataset\end{tabular} \\ \\
\multicolumn{11}{l}{\textit{*TR: Trust, CE: Contribution Evaluation, IM: Incentive Mechanism, RP: Reputation, VF: Verification, AG: Secure Aggregation}} \\ \bottomrule
\end{tabular}%
}
\end{table*}

\subsubsection{Contribution based Trustworthy Incentive Mechanism}

FL in healthcare has gained considerable attention for training ML models. Ensuring the accuracy and fairness of these models is crucial, especially considering the potential for bias, which can lead to disparities in predictive performance across different patient subgroups. To address this issue, research work in \cite{im10} focuses on the prevention of excessive bias through the use of reward systems. Firstly, the researchers determine the contributions of each institution towards predictive performance and bias using an approximation method based on Shapley values. Subsequently, various reward systems are designed to incentivize high predictive performance or low bias, with a combined reward system incentivizing both. The effectiveness of these reward systems is evaluated using medical chest X-ray datasets and the results show that they successfully incentivize contributions towards a well-performing model with low bias. The study highlights the need for further research on developing practical reward distribution strategies, considering the challenge of incentivizing low bias contributions. Ultimately, the goal is to design a reward system that balances high predictive performance and low bias for fair and Trustworthy FL models in healthcare. Authors aims to quantify the bias in FL and design reward systems that encourage institutions to make contributions towards high predictive performance and low bias. The first step involves determining the contributions of FL institutions to bias through the use of Shapley value approximation method. The second step focuses on transforming these contributions into a reward system that incentivizes both high predictive performance and low bias. This requires further research on viable reward distribution strategies to incentivize institutions effectively. The ultimate goal is to develop a reward scheme that incentivizes both high performance and low bias in FL.

A novel approach to incentivize high-quality contributions in FL through tokenization is presented in \cite{im13}. Our method allocates tokens based on the value of each client's contribution during the aggregation phase. This incentivizes quality participation and discourages poor updates or attacks in a decentralized network environment, such as in Web 3.0. The proposed tokenized incentive design accommodates clients with diverse profiles, making it a robust solution for collaborative FL training. This approach is efficient and focuses on the allocation of tokens within budget constraints, known as quota, to ensure the optimal selection of local model parameters.

The authors proposes a fairness-aware incentive mechanism for FL called FedFAIM \cite{im21}, which addresses the problem of unfairness in the existing FL systems. The proposed mechanism achieves two types of fairness: aggregation fairness and reward fairness. Aggregation fairness is achieved through an efficient gradient aggregation method that filters out low-quality local gradients and aggregates high-quality ones based on data quality. Reward fairness is achieved through a Shapley value-based contribution assessment method and a novel reward allocation method based on reputation and distribution of local and global gradients. The proposed method ensures that participants are assigned a unique model with performance reflecting their contribution. The reward allocation mechanism incorporates reputation to determine the model's performance level assigned to each participant. The proposed mechanism is evaluated, and the results show that it outperforms existing FL systems.

FL Incentivizer (FLI), to incentivize high-quality data owners to contribute their data to a FL system is proposed in \cite{im16}. FLI is a real-time algorithm that ensures contribution fairness, regret distribution fairness, and expectation fairness while accounting for the interests of both the federation and data owners. Data owners who have contributed high-quality data will receive a higher share of subsequent revenues generated by the federation. FLI dynamically adjusts data owners' shares to distribute benefits fairly and sacrifice among them. FLI produces near-optimal collective utility while limiting data owners' regret and accounting for the temporary mismatch between contributions and rewards. By incentivizing data owners, FLI enables a healthy FL ecosystem to emerge over time.

The authors proposes FIFL \cite{im14}, a fair incentive mechanism for FL that combines workers' contribution and reputation indicators to reward honest workers and punish attackers. FIFL adopts a blockchain-based audit method and a reputation-based server selection method to prevent malicious nodes from manipulating assessment results. FIFL consists of four components: attack detection, reputation, contribution, and incentive modules. The attack detection module removes potential malicious updates to prevent model damage, and the reputation module measures workers' trustworthiness based on historical events. The contribution module measures workers' current utility to the system, and the incentive module determines workers' rewards based on both reputation and contribution indicators. FIFL is the first profit-sharing solution that works in unreliable federations containing unstable attackers and maintains a constant system revenue in such scenarios. The paper conducts comprehensive experiments to evaluate FIFL's effectiveness, and the results show that FIFL outperforms the baseline models. The study contributes to the development of FL systems by proposing a fair and effective incentive mechanism that can incentivize honest participation in FL while mitigating the negative effects of malicious attacks. The use of a blockchain-based audit method ensures transparency and prevents fraud.

\subsubsection{Security and Privacy based Trustworthy Incentive Mechanism}

Authors presents a novel deep learning framework named DeepChain in \cite{im8}, which aims to tackle the security and fairness problems that are present in traditional FL. The framework is based on blockchain technology and provides a value-driven incentive mechanism to encourage honest behavior from participants. The incentive mechanism in DeepChain consists of two security mechanisms in blockchain: a trusted time clock mechanism and a secure monetary penalty mechanism. These mechanisms work together to ensure fairness during the collaborative training process and to prevent participants from behaving incorrectly. In addition to the incentive mechanism, DeepChain also guarantees data privacy and provides auditability for the training process. The confidentiality of local gradients is ensured through the use of the Threshold Paillier algorithm, which provides additive homomorphic properties. The auditability is achieved through the universally verifiable CDN (UVCDN) protocol \cite{im9}. By incorporating these features, DeepChain provides a secure and fair collaborative training environment for deep learning models. The framework not only protects the privacy of local gradients and guarantees the correctness of the training process, but also encourages honest behavior from participants through the use of incentives.

FGFL \cite{im15}, a novel incentive governor for FL that assesses workers' contributions and reputations to reward efficient workers fairly and eliminate malicious ones. FGFL contains two main parts: a fair incentive mechanism and a reliable incentive management system. The incentive mechanism measures workers' utility and reliability, using the product of the two indicators to determine rewards and punishments. FGFL also includes a blockchain-based incentive management method, FG-chain, which achieves trusted management of incentives through an audit method and a reputation-based server selection method. The paper evaluates the effectiveness and fairness of FGFL through theoretical analysis and comprehensive experiments. The results show that FGFL effectively assesses workers' trustworthiness and utilities, prevents the decline of system revenue caused by attackers, and achieves secure management of FL. The use of a time-decay SLM algorithm and a lightweight method based on gradients similarity ensures real-time assessment of workers' contributions and reputations. FGFL contributes to the development of FL systems by proposing a fair and reliable incentive mechanism and management system.

A FL system that leverages blockchain technology, local differential privacy (LDP), and zero-knowledge proofs (ZKPs) to achieve fairness, integrity, and privacy has been proposed in \cite{im22}. The proposed FL architecture provides fair incentives to clients by measuring their individual contribution to the global model performance based on actual parameters and incentivizing them accordingly through a smart contract on the blockchain. Non-interactive ZKPs ensure the integrity of the FL system by enabling clients to validate fellow clients' model updates without revealing any private data. The blockchain-based design ensures neutrality, immutability, and transparency of the architecture, while LDP ensures clients' model updates cannot leak information on patterns within their private data. Overall, the proposed FL system offers a novel and smart FL-based architecture that achieves fairness, integrity, and privacy in a practical and scalable manner.

Table 4 provides an extensive overview of the research conducted on trustworthy Incentive mechanism in FL. This table highlights various methods, techniques, and approaches employed to ensure trustworthiness during the reward allocation process in FL.

\subsection{Accountability and Auditibility in FL}
In this section, we categorize trustworthy incentive mechanism algorithms and methodologies into two sub-categories based on their primary objectives: Smart Contract and Committee Selection based Trustworthy Auditibility and Accountability Mechanisms.
\subsubsection{Smart Contract based Trustworthy Accountability and Auditibility}

BlockFLow \cite{ab2, ab3}, is a fully PP decentralized FL system that aims to data privacy and attck resilience with contributions evaluation based on quality of data. Differential privacy and a unique auditing mechanism is utilized to protect datasets and evaluate model contribution respectively. Ethereum smart contracts is used in BlockFLow to incentivize good behavior resulting in more accountable and transparent approach.

A blockchain-based architecture for FL systems that improves accountability and fairness is proposed in \cite{ab5, ab6}. To achieve this, a smart contract-driven data-model provenance registry is designed to track and record local data used for model training, enabling auditing. Additionally, a weighted fair training dataset sampler algorithm is introduced to improve fairness affected by data class distribution heterogeneity. In this architecture, each client and central server should have a blockchain node to form a network. The blockchain is utilized for its immutability and transparency, and the smart contract improves accountability. The use of blockchain technology can improve accountability and has been evaluated previously. The proposed integration of blockchain and FL is feasible as both systems are decentralized. The effectiveness of the approach is demonstrated using a COVID-19 detection scenario with X-rays.
\subsubsection{Committee Selection based Trustworthy Accountability and Auditibility}
The VFChain framework \cite{ab1} offers verifiable and auditable FL through the use of blockchain technology. It guarantees verifiability by choosing a committee via the blockchain to jointly aggregate models and create verifiable proofs. To achieve auditability, a unique authenticated data structure is introduced to boost the search efficiency of verifiable proofs and facilitate secure committee rotation. VFChain enhances search efficiency for multi-model learning tasks while penalizing dishonest participants through the withdrawal of pre-frozen deposits. The paper introduces Dual skip chain (DSC), a practical committee selection method and novel authenticated data structure for blockchain. DSC augments verifiable proof search efficiency, enables secure committee rotation, and allows clients to securely traverse the blockchain. Furthermore, a comprehensive audit layer combines independent audit processes to improve model verification and audit performance.

The authors in \cite{ab4}, proposes FLChain, a decentralized and auditable framework for FL that addresses the issues of misbehavior, lack of auditability, and incentives. FLChain replaces the traditional parameter server with a consensus computation result, ensuring that the ecosystem is healthy and public auditable. The proposed framework provides sufficient incentive and deterrence to distributed trainers, allowing for a healthy marketplace for collaborative-training models. Honest trainers can receive a fairly partitioned profit from well-trained models based on their contribution, while malicious actors can be detected and heavily punished. Additionally, the paper introduces DDCBF, which accelerates the query of blockchain-documented information to reduce the time cost of misbehavior detection and model query.

Table 5 provides an extensive overview of the research conducted on accountability and auditibility aspect Trustworthy FL.

\subsection{Discussion}
In our discussion of the selected papers, we will focus on their pitfalls and potential improvements concerning trustworthiness in fairness aware FL. 

Many research works are incorporating client Selection using Reputation Mechanisms \cite {cs1, cs2, cs3, cs5, cs6, cs7, cs11, cs14, cs15, cs17, cs18}. There are some common limitations like Incomplete trustworthiness assessment, scalability and overhead, holistic trustworthiness assessment. Many papers focus on a single aspect of trustworthiness, such as client reputation or resource availability, without considering the interplay of various factors. Some proposed techniques may introduce high computational or communication overhead, limiting their applicability in large-scale FL deployments. Future research should focus on comprehensive trust models that incorporate multiple aspects, such as reliability, robustness, and privacy. Developing scalable client selection and reputation mechanisms with minimal overhead will help in practical FL deployments. Blockchain-based trustworthiness solutions \cite {cs4, cs9, cs12, cs13} often face issues such as latency, energy consumption, and storage constraints, which may hinder their use in FL systems. Combining blockchain with other trust-enhancing technologies (e.g., secure hardware, zero-knowledge proofs) may help overcome some limitations and provide a more robust trust foundation. Investigating new consensus protocols with a focus on low-latency, energy-efficient, and lightweight solutions will benefit blockchain-based FL trustworthiness. Application-specific trustworthy client selection \cite {cs10, cs16} address trustworthiness in specific application domains, which may limit the generalizability of their findings to broader FL settings. There is often insufficient benchmarking and comparison with existing methods, making it difficult to assess the real-world impact of the proposed solutions. Developing trust models and mechanisms that can be easily adapted to various application domains will improve the overall trustworthiness of FL systems. Future research should focus on evaluating and comparing proposed solutions using standardized metrics and real-world datasets, helping to assess their practical impact on trustworthiness in FL.

Trust models and assessments primarily concern the evaluation of users' contributions and the establishment of trust among participants. However, the studies \cite {ce1, ce2, ce3} provide incomplete trustworthiness assessments, lacking consideration of aspects like reliability and robustness. Moreover, their trust models are narrow in focus, tailored to specific application domains. Incentive mechanisms aim to encourage participant cooperation and ensure fair reward distribution. Several works employ auction or contract-based mechanisms (\cite{im3}, \cite{im11}, \cite{im18}). While these approaches can provide fair incentives, they may not adequately address the users' privacy concerns. In addition, the use of reputation systems in some studies (\cite{im6}, \cite{im7}, \cite{im9}) may lead to unfair treatment of new users who have not yet established a reputation. Thus, a more balanced approach, combining both privacy preservation and fairness, should be investigated. Privacy-preserving techniques are essential for maintaining data confidentiality and user privacy in FL. Techniques like \cite {im8, im9, ce5} however, face limitations in scalability and generalizability, requiring improvements in lightweight privacy-preserving methods.

In conclusion, future research should prioritize developing comprehensive trust models, exploring lightweight and scalable trust and incentive mechanisms, and enhancing privacy preservation. Furthermore, evaluations using standardized metrics and real-world datasets are necessary to assess the practical impact of trustworthiness solutions in FL. By addressing these pitfalls and critics, we can significantly improve the learning process and trustworthiness in FL.

\section{Security and Privacy Aware Trustworthy FL}

FL has emerged as a promising approach to train models on distributed data while maintaining data privacy. However, achieving Trustworthy FL necessitates addressing various security and privacy challenges, including verifiability, privacy preservation, and secure aggregation. Data privacy serves as the driving force behind FL's development, and it is crucial that FL models effectively safeguard data privacy throughout their lifecycle to foster trust among participants. Although FL inherently provides a certain level of data privacy, assumptions about the integrity of the various actors and parties within the federation must be made. Preventing information leakage, whether by ensuring secure communication among honest-but-curious federation members or defending against external malicious attacks, must be prioritized.
\begin{figure*}[!ht]
\centering
\includegraphics[width=17cm,height=14cm,keepaspectratio]{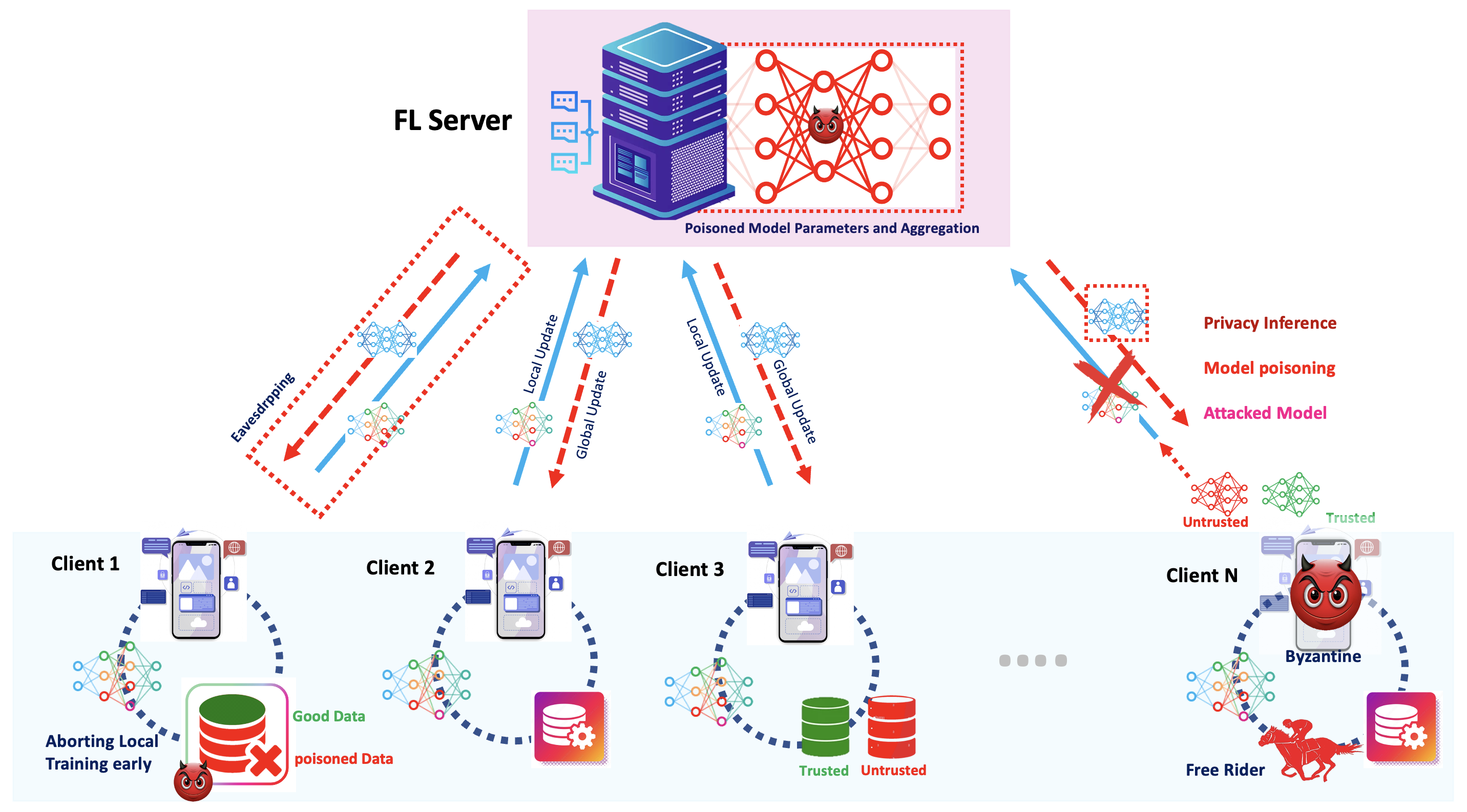}
\caption{A visual overview of the attack landscape in the context of security and privacy preservation for FL, highlighting various threats and challenges faced in maintaining a secure and privacy-preserving FL environment. .}
\label{fog9}
\end{figure*}
To address data privacy in FL, techniques can be grouped into three main categories: protection against poisoning attacks, preserving privacy in the face of data and model inversion, and ensuring verifiability and trustworthiness. Poisoning attacks can compromise the integrity of training data or manipulate the training process, making it essential to evaluate FL models for their defense mechanisms and effectiveness against such attacks. Due to FL's decentralized nature, it is vulnerable to numerous security threats, including data poisoning, model inversion, and membership inference attacks. The lack of a central authority further complicates matters, potentially leading to communication bottlenecks and impacting the quality of the trained models. As a result, designing a Trustworthy FL framework that ensures verifiability, secure aggregation, and privacy preservation is critical. Furthermore, privacy concerns arise in FL since global models and local updates can expose data information. Adversaries can accurately recover raw data from local updates or exploit differences between consecutive global models to reveal data properties and membership, posing a significant risk to user privacy. Training integrity is often neglected, as participants might not fully execute the protocol due to limited computational resources, which can reduce model accuracy. Trustworthiness of centralized FL centers can be difficult to assess, and communication can become a bottleneck. Therefore, it is crucial to develop a privacy-preserving, verifiable, and decentralized FL framework that safeguards data privacy, protects global models and local updates, ensures training integrity, and alleviates trust and communication concerns arising from centralized centers. The challenge of creating such a framework that balances participant rights and interests while achieving high-performance models has not yet been sufficiently explored. Moreover, dishonest aggregation from malicious servers can result in harmful aggregated gradients for some or all clients, further emphasizing the importance of addressing these issues. We established detailed sub-categorization of all the aspects of Security \& Privacy aware Trustworthy FL. The Illustration of the refined taxonomy related to Security \& Privacy aware Trustworthy FL is presented in Fig. 7. However, Fig. 8, depicts the various security and privacy challenges present within FL environments.

\subsection{Varifiablity in Trustworthy FL}

\subsubsection{Conensus and Smart Contract based Varifiability}

Authors in \cite{v1} proposes a blockchain-empowered FL framework for the Internet of Vehicles (IoV). The framework offers secure, privacy-preserving, and verifiable FL services by mitigating potential anomalies that may occur at the FL server or during communication. The proposed approach limits access to the blockchain network to only necessary participants, reducing the attack surface and offering secure and trustworthy services in the IoV.

Authors proposed a privacy-preserving and verifiable FL method based on blockchain in \cite{v7}. The method consists of three main components: Secure aggregation protocol: This protocol is based on gradient masking and verifiable secret sharing (VSS). It provides protection against potential malicious miners in the blockchain and is robust to clients dropping out. Blockchain structure with global gradient verification: The blockchain structure integrates gradient verification into the consensus algorithm using polynomial commitment. This design effectively defends against tampering attacks and ensures reliable FL. Gradient compression method reduces communication overhead while maintaining the accuracy of the trained model. Overall, the proposed privacy-preserving and verifiable FL method based on blockchain aims to enhance the reliability and privacy of the FL process by combining cryptographic techniques, consensus algorithms, and communication optimization methods. The problem discussed in the paper is the privacy and reliability concerns in FL. FL allows for a model to be trained without access to the raw data, but the sharing of gradients in the process still poses privacy concerns. Additionally, there is a risk of malicious parties manipulating the aggregated gradients and affecting the accuracy of the model.

BytoChain in \cite{v10}, introduces verifiers to execute heavy verification workflows in parallel and byzantine attacks can be detected through a Proof-of-Accuracy (PoA) consensus mechanism. BytoChain reduces the verification overhead for miners and prevents the loss of accuracy by tolerating models with limited non-positive gains. The Proof-of-Accuracy consensus effectively detects inferior models and provides a secure and efficient solution for FL.

A BC-based FL approach for device failure detection in IIoT has been proposed in \cite{v11}. A Merkle tree is used for verifiable integrity of each client's data. The impact of data heterogeneity can be reduce in device failure detection by implementing a centroid distance weighted federated averaging (CDW\_FedAvg) algorithm. A smart contract-based incentive mechanism is designed to motivate the participation of clients. The authors evaluate the feasibility, accuracy, and performance of the proposed approach using a real industry use case of detecting failures in water-cooled magnetic levitation chillers in air-conditioners.

\begin{figure*}[!ht]
\centering
\includegraphics[width=10cm,height=10cm,keepaspectratio]{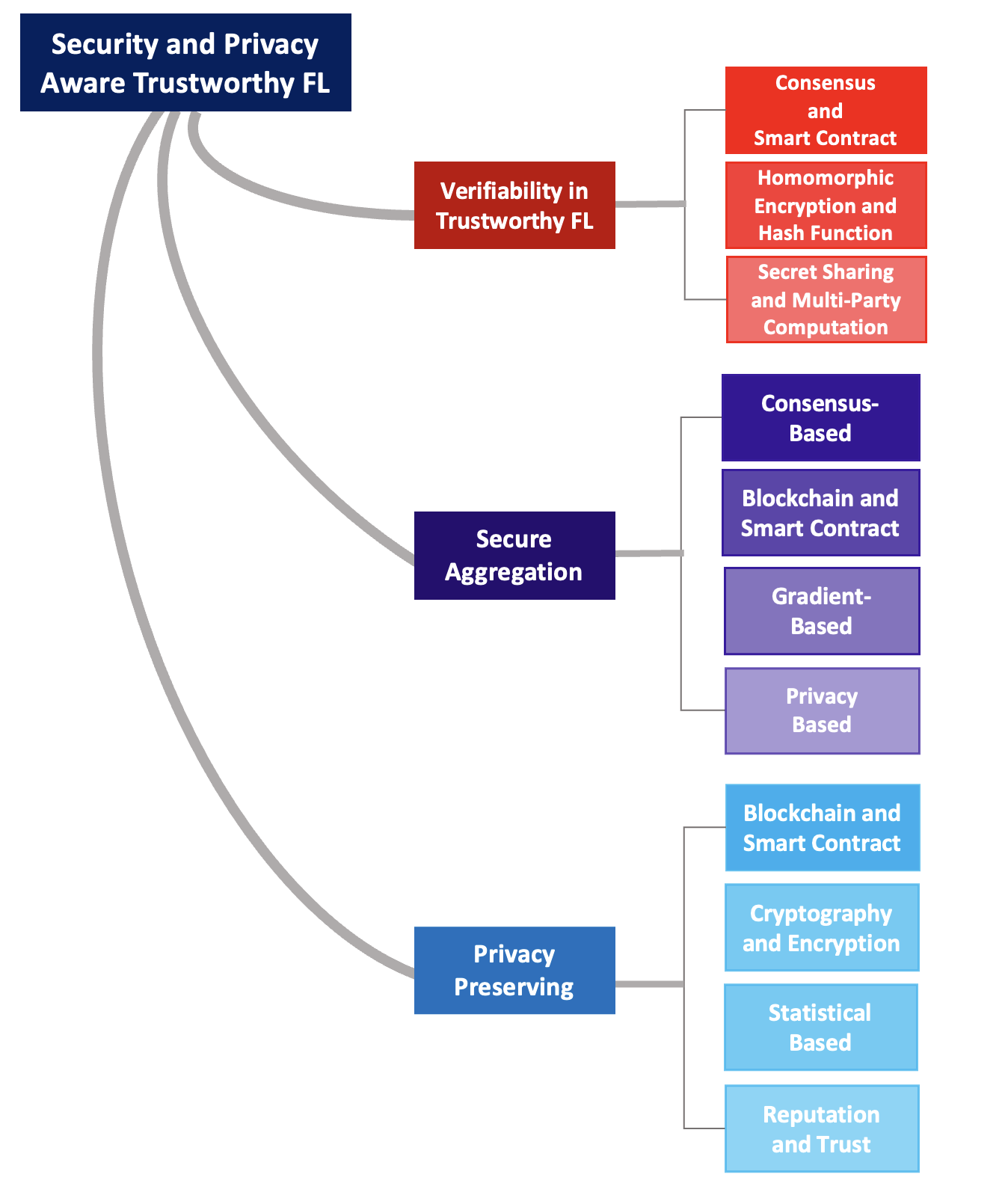}
\caption{A visual overview of the Categorization of Security \& Privacy aware Trustworthy FL.}
\label{fog10}
\end{figure*}

\subsubsection{Homomorphic Encryption and Hash Function based Varifiability}

Authors in \cite{v3}, introduces PVD-FL, a secure, verifiable, and privacy-centric decentralized FL structure aimed at mitigating security threats such as privacy invasion and integrity breaches. PVD-FL's objective is to facilitate safe deep learning model training in a decentralized setting, eliminating the need for a trusted central entity and circumventing communication restrictions. The work presents an algorithm called Efficient and Verifiable Cipher-based Matrix Multiplication (EVCM), used for basic deep learning computations. This algorithm, combined with a set of decentralized methods, builds the PVD-FL structure, ensuring confidentiality of the global model and local updates, and offering training step verification. Security analysis indicates PVD-FL's capability to counter different inference attacks while upholding training integrity. With no need for a centralized body, this framework enables secure construction of a global deep learning model. The EVCM algorithm, grounded on lightweight symmetric homomorphic encryption, safeguards the training process. A range of decentralized algorithms are also designed for precise deep learning model training. The system confirms confidentiality, training step verifiability, high accuracy, and efficient computational and communication cost. Experimentation on real-world datasets verifies PVD-FL's efficacy and efficiency

The referenced work in \cite{v4} presents VerifyNet, a groundbreaking framework that addresses verifiability and privacy concerns in FL. It employs a homomorphic hash function and pseudorandom mechanisms to allow user verification, while a double-masking protocol safeguards local gradient privacy. This secure framework manages user dropouts during training and preserves privacy. VerifyNet's objective is to enable clients to validate cloud server accuracy and protect user privacy throughout the training. The framework consists of two main elements: a double-masking protocol for maintaining gradient confidentiality, and a verification method that combines a homomorphic hash function and pseudorandom approaches to facilitate user confirmation of server results with limited overhead.

The problem with FL is that while it allows for training a model without accessing training sets, the server can extract information from the shared gradients or falsify the calculated result, affecting the accuracy of the model. To address these issues, a privacy-preserving and verifiable FL scheme is proposed in \cite{v6}. The scheme processes shared gradients using a combination of the Chinese Remainder Theorem and Paillier homomorphic encryption, providing privacy preservation with low computation and communication costs. Additionally, the bilinear aggregate signature technology is introduced to verify the correctness of the aggregated gradient, ensuring the accuracy of the trained model.

A privacy-preserving decentralized workflow for trusted FL is proposed in \cite{v8}. This proof-of-concept uses decentralised identity technologies, such as Hyperledger Aries/Indy/Ursa, to establish secure communication channels for entities with verified credentials to participate in a FL process related to mental health data. The trust model is defined and enforced through decentralised identity standards, and the authentication system is extended by separating Hyperledger Aries agents and controllers into isolated entities. A decentralized peer-to-peer infrastructure, TFL, is presented which uses Decentralized Identifiers (DIDs) and Verifiable Credentials (VCs) for mutual authentication and federated ML within a healthcare trust infrastructure. The performance of the TFL approach is improved compared to the previous state-of-the-art while maintaining the privacy guarantees of the authentication techniques and privacy-preserving workflows. The problem discussed in this paper is the need for a privacy-preserving and trusted FL workflow for mental health data.

\begin{table*}[]
\caption{Verifiability in Trustworthy FL}
\label{tab:my-table6}
\resizebox{\textwidth}{!}{%
\begin{tabular}{@{}ccllcccclll@{}}
\toprule
\textbf{\begin{tabular}[c]{@{}c@{}}Proposed \\ Technique\end{tabular}} &
  \textbf{YoP} &
  \multicolumn{1}{c}{\textbf{\begin{tabular}[c]{@{}c@{}}Accompanied \\ Algorithm\end{tabular}}} &
  \multicolumn{1}{c}{\textbf{Proposed Methodology}} &
  TR &
  PP &
  \begin{tabular}[c]{@{}c@{}}CF/\\ CR\end{tabular} &
  AG &
  \multicolumn{1}{c}{\textbf{Problem Encountered}} &
  \multicolumn{1}{c}{\textbf{Benefits}} &
  \multicolumn{1}{c}{\textbf{Data Set}} \\ \midrule
\begin{tabular}[c]{@{}c@{}}BC \\ empowered \\ FL for IoV\\   \cite{v1}\end{tabular} &
  2021 &
  Blockchain &
  \begin{tabular}[c]{@{}l@{}}Mitigating potential anomalies that may occur\\ at the FL server or during communication.\end{tabular} &
  \quad\circledmark[cyan!60] &
  \quad\circledmark[cyan!60] &
   &
   &
  \begin{tabular}[c]{@{}l@{}}Poisoning and\\ Reverse\\ Engineering\\ attacks\end{tabular} &
  \begin{tabular}[c]{@{}l@{}}Reduction of\\ attack surface,\\ Robustness\end{tabular} &
  \multicolumn{1}{c}{NA} \\
\begin{tabular}[c]{@{}c@{}}AV-FL\\   \cite{v2}\end{tabular} &
  2022 &
  \begin{tabular}[c]{@{}l@{}}SMPC protocol \\ based on HPRA \\ and HPRE\end{tabular} &
  \begin{tabular}[c]{@{}l@{}}Protects client confidentiality and data\\ verifiability while being flexible and\\ dynamically adjustable for participant\\ dropouts.\end{tabular} &
  \quad\circledmark[cyan!60] &
  \quad\circledmark[cyan!60] &
   &
  \quad\circledmark[cyan!60] &
  \begin{tabular}[c]{@{}l@{}}Malicious\\ clients and\\ client dropout\end{tabular} &
  \begin{tabular}[c]{@{}l@{}}data provenance\\ verifiability and\\ client\\ confidentiality\end{tabular} &
  \multicolumn{1}{c}{MNIST} \\
\begin{tabular}[c]{@{}c@{}}PVD-FL\\   \cite{v3}\end{tabular} &
  \multicolumn{1}{l}{2022} &
  EVCM &
  \begin{tabular}[c]{@{}l@{}}Framework ensures the confidentiality of both\\ the global model and local updates and verifies\\ every training step\end{tabular} &
  \quad\circledmark[cyan!60] &
  \quad\circledmark[cyan!60] &
  \quad\circledmark[cyan!60] &
  \quad\circledmark[cyan!60] &
  \begin{tabular}[c]{@{}l@{}}Integrity\\ verification and\\ communicational\\ bottleneck\end{tabular} &
  \begin{tabular}[c]{@{}l@{}}Guarantee\\ confidentiality and\\ integrity\\ verification, high\\ training accuracy\end{tabular} &
  \begin{tabular}[c]{@{}l@{}}MNIST,\\ Thyroid\\ \cite{d10}, Breast\\ cancer \cite{d11},\\ and\\ German\\ credit \cite{d12}\end{tabular} \\
\begin{tabular}[c]{@{}c@{}}VerifyNet\\ \cite{v4}\end{tabular} &
  2019 &
  \begin{tabular}[c]{@{}l@{}}Homeomorphic Hash \\ function (HHF), Double \\ Masking Protocol\end{tabular} &
  \begin{tabular}[c]{@{}l@{}}Uses homomorphic hash function and\\ pseudorandom for verifiability and a double\\ masking protocol to guarantee the\\ confidentiality of users' local gradients\end{tabular} &
  \quad\circledmark[cyan!60] &
  \quad\circledmark[cyan!60] &
  \quad\circledmark[cyan!60] &
   &
  \begin{tabular}[c]{@{}l@{}}Integrity\\ verification and\\ privacy of local\\ gradient\end{tabular} &
  \begin{tabular}[c]{@{}l@{}}Better verification\\ accuracy and\\ dropout handling\end{tabular} &
  MNIST \\
\begin{tabular}[c]{@{}c@{}}VFL\\  \cite{v5}\end{tabular} &
  2020 &
  \begin{tabular}[c]{@{}l@{}}Lagrange interpolation \\ and Blinding Technology\end{tabular} &
  \begin{tabular}[c]{@{}l@{}}VFL framework uses Lagrange interpolation to\\ set interpolation points for verifying the\\ correctness of the aggregated gradients\end{tabular} &
  \quad\circledmark[cyan!60] &
  \quad\circledmark[cyan!60] &
   &
  \quad\circledmark[cyan!60] &
  \begin{tabular}[c]{@{}l@{}}Forged\\ Aggregated\\ gradients\end{tabular} &
  \begin{tabular}[c]{@{}l@{}}Efficient detection\\ of forged results\end{tabular} &
  MNIST \\
\begin{tabular}[c]{@{}c@{}}Verifiable\\  and PP-FL\\  \cite{v6}\end{tabular} &
  2020 &
  \begin{tabular}[c]{@{}l@{}}Bilinear Aggregate \\ Signature, Homomorphic \\ Encryption (HE)\end{tabular} &
  \begin{tabular}[c]{@{}l@{}}Processes shared gradients using a\\ combination of the Chinese Remainder\\ Theorem and Paillier homomorphic encryption\end{tabular} &
  \quad\circledmark[cyan!60] &
  \quad\circledmark[cyan!60] &
   &
  \quad\circledmark[cyan!60] &
  \begin{tabular}[c]{@{}l@{}}Malicious and\\ falsified server\\ aggregation\end{tabular} &
  \begin{tabular}[c]{@{}l@{}}Ensures accuracy\\ of trained model\end{tabular} &
  MNIST \\
\begin{tabular}[c]{@{}c@{}}PPV-BC\\  \cite{v7}\end{tabular} &
  2022 &
  \begin{tabular}[c]{@{}l@{}}BC consensus algorithm, \\ commitment protocol\end{tabular} &
  \begin{tabular}[c]{@{}l@{}}A secure aggregation protocol based on\\ verifiable secret sharing and gradient masking\\ along with global gradient verification using\\ BC\end{tabular} &
  \quad\circledmark[cyan!60] &
  \quad\circledmark[cyan!60] &
   &
  \quad\circledmark[cyan!60] &
  \begin{tabular}[c]{@{}l@{}}Model\\ Inversion\\ attack, Single\\ point of Failure\end{tabular} &
  \begin{tabular}[c]{@{}l@{}}Robust of client\\ dropout, protects\\ model privacy and\\ tempering attacks\end{tabular} &
  \begin{tabular}[c]{@{}l@{}}CIFAR10,\\ 20newsgr\\ up \cite{d13}\end{tabular} \\
\begin{tabular}[c]{@{}c@{}}TFL\\  \cite{v8}\end{tabular} &
  2021 &
  HE &
  \begin{tabular}[c]{@{}l@{}}A trust model is defined and enforced through\\ decentralized identity standards, and the\\ authentication system is extended\end{tabular} &
  \quad\circledmark[cyan!60] &
  \quad\circledmark[cyan!60] &
  \quad\circledmark[cyan!60] &
  \quad\circledmark[cyan!60] &
  \begin{tabular}[c]{@{}l@{}}Malicious\\ attacks\end{tabular} &
  \begin{tabular}[c]{@{}l@{}}Improved trust in\\ mental health data\\ evaluation\end{tabular} &
  \begin{tabular}[c]{@{}l@{}}Mental\\ Health\\ dataset \cite{d14}\end{tabular} \\
\begin{tabular}[c]{@{}c@{}}FedTrust and VAP\\ \cite{v9}\end{tabular} &
  2022 &
  \begin{tabular}[c]{@{}l@{}}Shamir’s \\ secret sharing \\ and BGW\end{tabular} &
  \begin{tabular}[c]{@{}l@{}}There are three scenarios SVM, SSVM and\\ MSVM to address the security and privacy\\ requirements in FL\end{tabular} &
  \quad\circledmark[cyan!60] &
  \quad\circledmark[cyan!60] &
  \quad\circledmark[cyan!60] &
  \quad\circledmark[cyan!60] &
  \begin{tabular}[c]{@{}l@{}}Lack of trust,\\ reliability and\\ robustness\end{tabular} &
  \begin{tabular}[c]{@{}l@{}}Better training\\ accuracy,\\ trustworthiness\end{tabular} &
  CIFAR-10 \\
\begin{tabular}[c]{@{}c@{}}BFL based \\ BytoChain\\ \cite{v10}\end{tabular} &
  2021 &
  PoA consensus &
  \begin{tabular}[c]{@{}l@{}}improves the efficiency of model verification\\ by introducing verifiers and detects byzantine\\ attacks through a PoA\end{tabular} &
  \quad\circledmark[cyan!60] &
  \quad\circledmark[cyan!60] &
   &
   &
  \begin{tabular}[c]{@{}l@{}}Byzantine\\ attack\end{tabular} &
  \begin{tabular}[c]{@{}l@{}}Improved model\\ verification and\\ attack detection\end{tabular} &
  MNIST \\
\begin{tabular}[c]{@{}c@{}}CDW\_FedAvg\\  \cite{v11}\end{tabular} &
  2020 &
  \begin{tabular}[c]{@{}l@{}}Smart Contract, \\ Proof-of-work PoW\end{tabular} &
  \begin{tabular}[c]{@{}l@{}}Ensures verifiable integrity of client data by\\ creating a Merkle tree for each client\end{tabular} &
  \quad\circledmark[cyan!60] &
   &
   &
   &
  \begin{tabular}[c]{@{}l@{}}Data\\ heterogeneity,\\ Device failure\end{tabular} &
  \begin{tabular}[c]{@{}l@{}}Improved\\ accuracy and\\ feasibility\end{tabular} &
  Sim dataset \\
\begin{tabular}[c]{@{}c@{}}BC-based \\ PPFL\\  \cite{v12}\end{tabular} &
  2019 &
  HE, MPC &
  \begin{tabular}[c]{@{}l@{}}Verification mechanism and contribution\\ aware incentive mechanism using blockchain\\ ledger\end{tabular} &
  \quad\circledmark[cyan!60] &
  \quad\circledmark[cyan!60] &
   &
  \quad\circledmark[cyan!60] &
  \begin{tabular}[c]{@{}l@{}}Unreliable\\ Network\\ Connectivity\\ and data\\ privacy\end{tabular} &
  \begin{tabular}[c]{@{}l@{}}Better verifiability\\ and transparency\end{tabular} &
  \begin{tabular}[c]{@{}l@{}}Breast\\ Cancer\\ Dataset\end{tabular} \\
\begin{tabular}[c]{@{}c@{}}FedIPR\\  \cite{v13}\end{tabular} &
  2023 &
  MPC &
  \begin{tabular}[c]{@{}l@{}}Embeds watermarks into the FL models to\\ verify ownership\end{tabular} &
  \quad\circledmark[cyan!60] &
  \quad\circledmark[cyan!60] &
   &
   &
  robustness &
  \begin{tabular}[c]{@{}l@{}}Better attack\\ handling\end{tabular} &
  MNIST \\
\begin{tabular}[c]{@{}c@{}}SVeriFL\\  \cite{v14}\end{tabular} &
  2023 &
  \begin{tabular}[c]{@{}l@{}}BLS signature and \\ multi-party security, \\ HE\end{tabular} &
  \begin{tabular}[c]{@{}l@{}}Secure integrity verification of parameters\\ uploaded by participants and secure\\ correctness verification of aggregated results\\ by the server\end{tabular} &
  \quad\circledmark[cyan!60] &
  \quad\circledmark[cyan!60] &
   &
  \quad\circledmark[cyan!60] &
  \begin{tabular}[c]{@{}l@{}}Malicious\\ server conduct,\\ dishonest data\\ aggregation\end{tabular} &
  \begin{tabular}[c]{@{}l@{}}Improved practical\\ performance and\\ computational\\ efficiency\end{tabular} &
  \begin{tabular}[c]{@{}l@{}}MNIST,\\ FMNIST\end{tabular} \\
\begin{tabular}[c]{@{}c@{}}VERIFL\\ \cite{v15}\end{tabular} &
  2020 &
  HHF &
  \begin{tabular}[c]{@{}l@{}}Protocol uses a combination of a linearly HHF\\ and a commitment scheme to prevent the\\ aggregation server from forging results\end{tabular} &
  \quad\circledmark[cyan!60] &
  \quad\circledmark[cyan!60] &
   &
  \quad\circledmark[cyan!60] &
  \begin{tabular}[c]{@{}l@{}}Verification of\\ Data\\ aggregation,\\ High dropout\\ rate\end{tabular} &
  \begin{tabular}[c]{@{}l@{}}Improved\\ verification cost\\ and communication\\ efficiency\end{tabular} &
  Sim \\
\begin{tabular}[c]{@{}c@{}}VERSA\\ \cite{v16}\end{tabular} &
  2021 &
  HHF &
  \begin{tabular}[c]{@{}l@{}}It uses a lightweight primitive such as\\ pseudorandom generators to achieve\\ verifiability of model aggregation\end{tabular} &
   &
  \quad\circledmark[cyan!60] &
   &
  \quad\circledmark[cyan!60] &
  \begin{tabular}[c]{@{}l@{}}Untrusted\\ central server\end{tabular} &
  \begin{tabular}[c]{@{}l@{}}Verify the\\ correctness of\\ model, Improved\\ accuracy\end{tabular} &
  \begin{tabular}[c]{@{}l@{}}MNIST,\\ SVHN,\\ CIFAR100\end{tabular} \\
\begin{tabular}[c]{@{}c@{}}SFCV-FL\\ \cite{v17}\end{tabular} &
  2021 &
  \begin{tabular}[c]{@{}l@{}}HE, \\ Verifiable \\ Computation (VC)\end{tabular} &
  \begin{tabular}[c]{@{}l@{}}A concrete implementation of framework\\ which allows for authentication of the\\ computation of the global training model over\\ encrypted data\end{tabular} &
  \quad\circledmark[cyan!60] &
  \quad\circledmark[cyan!60] &
  \quad\circledmark[cyan!60] &
   &
  \begin{tabular}[c]{@{}l@{}}Integrity and\\ Confidentiality\\ threat\end{tabular} &
  \begin{tabular}[c]{@{}l@{}}Better practical\\ performance and\\ decreased letancy\end{tabular} &
  FEMNIST \\
\begin{tabular}[c]{@{}c@{}}CATFL\\ \cite{v18}\end{tabular} &
  2023 &
  \begin{tabular}[c]{@{}l@{}}Public Key\\ Infrastructure (PKI) \\ based authentication\end{tabular} &
  \begin{tabular}[c]{@{}l@{}}A pseudonym generation strategy is proposed\\ to balance the trade-off between user\\ anonymity and identity traceability\end{tabular} &
  \quad\circledmark[cyan!60] &
  \quad\circledmark[cyan!60] &
   &
  \quad\circledmark[cyan!60] &
  \begin{tabular}[c]{@{}l@{}}Model\\ Poisoning\\ attacks, privacy\\ leakage\end{tabular} &
  \begin{tabular}[c]{@{}l@{}}Lower\\ Communication\\ cost and higher\\ security\end{tabular} &
  Sim \\
\begin{tabular}[c]{@{}c@{}}DP-BFL\\ \cite{v19}\end{tabular} &
  2021 &
  GDP-FedAvg &
  \begin{tabular}[c]{@{}l@{}}It adopts a Local Differential Privacy (LDP)\\ technique to prevent inference attacks and an\\ incentive mechanism to prevent free rider\\ attack\end{tabular} &
  \quad\circledmark[cyan!60] &
  \quad\circledmark[cyan!60] &
   &
   &
  \begin{tabular}[c]{@{}l@{}}Free riding\\ attack,\\ Inference attack\end{tabular} &
  \begin{tabular}[c]{@{}l@{}}Enhanced\\ verifiability and\\ trust, high utility\end{tabular} &
  MNIST \\ \\
\multicolumn{11}{l}{\textit{*TR: Trust, PP: Privacy Preserving, CR: Correctness, AG: Secure Aggregation}} \\ \bottomrule
\end{tabular}%
}
\end{table*}

The authors propose a BC-based PPFL framework in \cite{v12}, that uses immutable and decentralized features of BC to record information flows such as client's local and  global updates, Fl tasks, and data provenance. This provides a verifiable mechanism for FL tasks, which was previously lacking in existing FL approaches. The verification mechanism can extend the semi-honest client assumption to a more realistic malicious client assumption, allowing for a secure FL process. Additionally, the blockchain ledger allows for the tracking of each client's contribution to the globally optimized model, enabling contribution-based incentive mechanisms and the rewarding of miners who own the improved model. The encrypted model updates are recorded on the blockchain, which allows for the tracking and verification of client contributions to the global model. In a preliminary experiment, the authors implemented a basic verification function where the server acts as a verifier to evaluate the gradients after the aggregate is recovered in each round. The updated global model's performance was then compared to the initial model of that round using the loss function.

Authors in \cite{v15} introduced the VERIFL protocol, which is designed to ensure the integrity of data aggregation in FL. The protocol uses a combination of a linearly homomorphic hash and a commitment scheme to prevent the aggregation server from forging results. The collision resistance of the hash scheme ensures that the server cannot have an honest client accept its false result. The protocol is also secure, as it only reveals the aggregation result computed by the server. The authors demonstrate through experiments that the VERIFL protocol can handle large amounts of data, a high number of clients, and high dropout rates with practical performance. This paper aims to address the challenge of verifying data aggregation in FL and provides a solution that is both efficient and effective.

In \cite{v16}, authors presented a security vulnerability in VerifyNet and a proposed solution to address this issue. The authors show that an attacker can recover a victim's model updates in a reasonable amount of time in VerifyNet. To address this security concern, the authors propose a new scheme called VERSA, which is a verifiable and privacy-preserving model aggregation method. The main contribution of this paper is the development of VERSA, which uses a lightweight primitive such as pseudorandom generators to achieve verifiability of model aggregation in cross-device FL. The authors aim to provide a secure solution for model aggregation in FL.

A secure framework for privacy-preserving and verifiable FL is introduced in \cite{v17}. The framework uses Homomorphic Encryption (HE) and Verifiable Computation (VC) techniques to ensure both the confidentiality and integrity of the data and models. The authors provide a concrete implementation of the framework using the Paillier additive homomorphic scheme and the LePCoV primitive, which allows for authentication of the computation of the global training model over encrypted data. The authors present several practical use cases for their secure FL solution, which addresses threats to both the training data and model from the aggregation server. The aim of this paper is to provide a secure and verifiable solution for FL, which balances the privacy concerns of individual participants and the need for accurate results.

\subsubsection{Secret Sharing and Multi-Party Computation based Varifiability}
In \cite{v2} authors present a novel framework for accountable and verifiable secure aggregation in FL. Our approach utilizes SMPC, secure multi-party computation protocols based on homomorphic proxy re-authenticators (HPRA) and homomorphic proxy re-encryption (HPRE) for secure aggregation, and integrates blockchain technology to enforce penalties for malicious behavior. The proposed framework protects client confidentiality and data verifiability while being flexible and dynamically adjustable for participant dropouts. To demonstrate its feasibility, we conduct performance tests using a blockchain prototype system in the context of Internet of Things networks. Our framework guarantees data provenance verifiability and holds malicious clients accountable, offering a robust solution for secure aggregation in FL.

The VFL framework in \cite{v5} is designed to address the privacy and security concerns in FL for industrial IoT applications. The problem with traditional FL approaches is that the shared gradient can still retain sensitive information of the training set, and a malicious aggregation server may return forged aggregated gradients. To address these concerns, the VFL framework uses Lagrange interpolation to set interpolation points for verifying the correctness of the aggregated gradients. This allows participants to independently and efficiently detect forged results with an overwhelming probability, and the computational overhead remains constant regardless of the number of participants. Additionally, the VFL framework also uses the blinding technology and Lagrange interpolation to achieve secure gradient aggregation. The joint model and gradients are protected from the aggregation server, and if no more than n-2 of n participants collude with the aggregation server, the gradients of other participants will not be leaked. The problem addressed in the paper is the security and privacy concerns in FL, where a shared model is trained by aggregating gradients from multiple parties without accessing the training data.

The problem addressed in \cite{v9} paper is the lack of trust in FL due to several critical security requirements such as reliability and robustness for moderators and privacy for clients. The authors aim to provide solutions for three specific scenarios in FL to address the security requirements. The first scenario is Single Verifiable Moderator (SVM) where the goal is to verify the correctness of the aggregated result at the client. The solution proposed is Verifiable Aggregation Protocol (VAP) which can be used to achieve robust FL for non-private gradients. The second scenario is Single Secure and Verifiable Moderator (SSVM) where the clients' gradients are private and should be protected. The solution proposed is still Verifiable Aggregation Protocol (VAP) which can achieve a verifiable and private aggregation. The third scenario is Multiple Secure and Verifiable Moderators (MSVM) where the focus is on the robustness of moderators. To achieve this goal, the authors decentralize a single moderator into multiple moderators and use the classic BGW protocol with Shamir's secret sharing to prevent disruption caused by moderator failure.

FedIPR in \cite{v13}, a novel ownership verification scheme is introduced to secure FL models and protect their intellectual property rights. This is done by embedding watermarks into the FL models to verify ownership. The authors emphasize the importance of security in FL and differentiate it from traditional distributed ML with no security guarantees. They propose a concept of Secure FL (SFL) with a goal of building trustworthy and safe AI systems with strong privacy and IP-right preservation.

The authors in \cite{v18}, introduces a new framework called Certificateless Authentication-based Trustworthy FL (CATFL) for 6G semantic communications. The aim of this framework is to provide a secure and trustworthy environment for FL while protecting user privacy. The CATFL framework provides mutual authentication between clients and servers to ensure the trustworthiness of the local client's gradients and the global model. Additionally, a pseudonym generation strategy is proposed to balance the trade-off between user anonymity and identity traceability. This strategy allows for tracing the original real identity of each CATFL client by trusted third-parties, enabling the identification of malicious devices. The CATFL framework can resist both poisoning attacks (including server-side and client-side) and privacy leakage (including gradient leakage and semantic representation leakage). Overall, the proposed CATFL framework provides a powerful deterrent against malicious threats to FL-based 6G semantic communication systems.

DP-BFL framework is proposed in \cite{v19},to address vulnerabilities in FL in IoT-based SM 3.0 networks. DP-BFL improves upon existing solutions by adopting a Local Differential Privacy (LDP) technique to prevent inference attacks, using decentralized verification and validation mechanisms to mitigate poisoning attacks, and employing a GDP-FedAvg algorithm to tackle membership attacks. Furthermore, the framework incorporates an incentive mechanism to discourage free-riding attacks and encourage more participation, and a QBC mechanism to nominate the consensus leader. DP-BFL's benefits include wiser privacy budget allocation, secure global model aggregation, and a more qualified leader for block and global model construction. Overall, DP-BFL enhances the verifiability and trust of FL in IoT-based SM 3.0 networks.
Table 6 presents an extensive overview of the research conducted on trustworthy verifiability in FL.

SVeriFL in \cite{v14}, a secure and verifiable FL framework that aims to address the issues of malicious server conduct and to protect the rights and interests of participants. The proposed system is based on a protocol designed with BLS signature and multi-party security, which enables secure integrity verification of parameters uploaded by participants and secure correctness verification of aggregated results conducted by the server. The system also ensures the consistency of aggregation results received by multiple participants and the dynamic parameters in the protocol enhance the security of the algorithm. The authors believe that the secure and verifiable FL framework will help to obtain a high performance FL model while protecting the rights and interests of participants.\\

\subsection{Secure Aggregation}

\subsubsection{Consensus based Secure Aggregation}

The article \cite{sa2} proposes a novel serverless FL framework called Committee Mechanism based FL (CMFL) to address the vulnerability of the typical FL system to Byzantine attacks from malicious clients and servers. The CMFL framework utilizes a committee system to monitor the gradient aggregation process and ensure the robustness of the algorithm with convergence guarantee. The committee members are responsible for filtering the uploaded local gradients and selecting the local gradients rated by elected members for the aggregation procedure through the selection strategy. Two opposite selection strategies are designed for accuracy and robustness considerations. The proposed framework distinguishes between honest and malicious gradients through a scoring system based on Euclidean distance. The proposed election strategy guarantees the honesty of committee clients. The experiments show that CMFL achieves faster convergence and better accuracy than typical FL while obtaining better robustness than traditional Byzantine-tolerant algorithms in a decentralized approach. The proposed framework ensures the secure aggregation of local gradients while preserving data privacy.

The authors proposes a decentralized blockchain-based FL (B-FL) architecture in \cite{sa4}, with secure global aggregation and Byzantine fault tolerance consensus protocol to ensure security and privacy against attacks from malicious devices and servers. The PBFT consensus protocol is utilized in B-FL to achieve high effectiveness and low energy consumption for Trustworthy FL. The proposed PBFT-based wireless B-FL architecture combines permissioned blockchain and wireless FL system to create a trustworthy AI model training environment that resists failures and attacks from malicious edge servers and devices. The article presents the detailed procedures of the PBFT-based wireless B-FL system and characterizes the training latency by considering the communication and computation processes. The combination of secure global aggregation and Byzantine fault tolerance consensus protocol ensures the integrity and robustness of the FL system.

A decentralized FL framework called BFLC is presented in \cite{sa14}, which uses BC for updates of local model exchange and storage of global model, to address security issues in centralized systems. An innovative committee consensus mechanism is also devised to reduce consensus computing and malicious attacks. The paper further discusses BFLC's scalability in terms of theoretical security, storage optimization, and incentives.

The proposed ensemble FL method \cite{sa12} provides a tight security guarantee against malicious clients, and an algorithm is introduced to compute certified security levels. The approach builds on existing base FL algorithms by training multiple global models, each learned using a randomly selected subset of clients. The ensemble method improves security by taking a majority vote among the models when predicting labels for testing data. Unlike single-global-model FL, the proposed approach uses a subsample of k clients chosen uniformly at random without replacement from the n clients.
\subsubsection{Blockchain and Smart Contract based Secure Aggregation}

The authors presents a novel cloud intrusion detection scheme called BFL-CIDS \cite{sa1} based on Blockchained-FL. The proposed scheme aims to improve the accuracy of a FL model for detecting distributed malicious attacks in the IoT environment while ensuring data privacy and security. The scheme sends local training parameters to a cloud computing center for global prediction and uses blockchain to store model training process information and behavior. To address the issue of false alerts, an alert filter identification module is introduced to filter out such alerts, reduce cloud workload, and improve the quality of the FL model. The proposed erasure code-based blockchain storage solution improves the storage performance of the blockchain and reduces storage pressure. The scheme uses Hyperledger Fabric expansion based on erasure codes to meet the storage needs of large amounts of alert training data in real scenarios. The proposed scheme provides an efficient and secure approach for detecting malicious attacks in distributed environments.

MiTFed \cite{sa15}, is a decentralized and secure framework based on FL, blockchain, and software-defined networking (SDN) for building efficient intrusion detection models that can detect new and sophisticated security attacks while maintaining the privacy of each SDN domain. SMPC based secure aggregation scheme is used by MiTFed to combine both local model updates and a BC-based scheme that uses Ethereum's smart contracts for maintaining a trustworthy, decentralized, and efficient collaboration. The proposed approach is flexible, transparent, tamper-proof, and scalable.

Fully decentralized FL presents challenges for securing the privacy of Internet of Health Things (IoHT) data due to the lack of training capabilities at all federated nodes, scarcity of high-quality training datasets, and the need for authentication of participating FL nodes. The author proposes a lightweight hybrid FL framework in \cite{sa18}, that combines blockchain smart contracts with edge training to manage trust, authentication, and distribution of global or locally trained models. The framework supports encryption of the dataset, model training, and inferencing process, and provides lightweight differential privacy (DP) for IoHT data. To ensure trust in the provenance of training data and models, blockchain and off-chain mechanisms track data lineage and secure the deep learning process. Secure aggregation of models mitigates malicious poisoning attacks by FL nodes. A provenance collection and management graph tracks the lineage of the data, model, and transaction history, and uses supervised learning to detect malicious intruder nodes.

FL has significantly improved the performance of automatic modulation recognition (AMC), but secure sharing of local model parameters remains an issue. To address this, a Blockchain-FL (BFL) framework is proposed for AMC \cite{sa17}, where the AMC model is cooperatively trained by sharing local model parameters with Blockchain. Additionally, a parameter validity evaluation method is designed to weaken the influence of malicious nodes during the aggregation process. The proposed BFL framework greatly improves the anti-attack ability of FL-based AMC schemes by enriching training samples. A validity evaluation mechanism is also introduced using the Criteria importance through inter-criteria correlation (CRITIC) to evaluate and determine the weights of network parameters. This ensures the security of the parameter aggregation process in the proposed BFL framework. The combination of Blockchain and FL provides a secure and efficient way to train AMC models, making it a promising solution for future research.

\subsubsection{Privacy based Secure Aggregation}

DeTrust-FL \cite{sa13}, is a PP based FL framework that addresses the issue of isolation attacks due to lack of transparency during secure aggregation. A decentralized trust consensus mechanism and functional encryption strategy is utilized by Detrust-FL to ensure privacy of secure aggregated model. The framework achieves a consensus among parties by allowing each party to present its 'decision' on a participation matrix that reflects its role in each secure aggregation round and its expectation of the proportion of benign parties in the FL training. This decentralized approach to secure computing mitigates the risk of privacy leaks caused by disaggregation attacks, which have recently been demonstrated to be a potential threat. Experimental results show that DeTrust-FL outperforms other SMPC enabled solutions in terms of reduced data volume transferred and training time. DeTrust-FL is an efficient and effective solution for PP FL in a decentralized trust setting.

FL is a communication-efficient approach to train ML models using data from multiple devices, but it also poses challenges such as device information leakage and centralized aggregation server risks. To address these challenges, a Structured Transparency empowered cross-silo FL on the Blockchain (ST-BFL) framework is proposed in \cite{sa19}. The framework employs homomorphic encryption, FL-aggregators, FL-verifiers, and smart contracts to enable structured transparency components including input and output privacy, output verification, and flow governance. The framework also includes a reputation system that allows authenticated and authorized clients to query the history and authenticity of the deep learning process, datasets used, and training process in a secure manner. Additionally, ST-BFL incorporates secure aggregation to prevent malicious nodes from introducing backdoor poisonous models and supports Intel SGX TEE to add an extra layer of security.

The authors in \cite{sa21} introduces a novel approach to make FL more robust to device failures or attacks. The proposed algorithm, called RFA, is based on a robust aggregation oracle built upon the geometric median and the smoothed Weiszfeld algorithm. It is designed to ensure convergence even if up to half of the devices send corrupted updates, making it suitable for FL with bounded heterogeneity. The algorithm preserves privacy by leveraging secure multi-party computation primitives. RFA comes in several variants, including a fast one with a single step of robust aggregation and another that adjusts to heterogeneity through on-device personalization. The approach is scalable, efficient, and easy to integrate with existing systems. The paper's contribution lies in presenting an approach that increases trust in the FL process, ensuring privacy and security while maintaining the efficacy of the learning algorithm.

The proposed research study in \cite{sa10} introduces a distributed backdoor attack called Cerberus Poisoning (CerP) that exploits the core assumption of current defense methods in FL. The defensive mechanisms rely on the assumption that a poisoned local model trained with poisoned data is significantly biased compared to those trained with poison-free data. CerP casts the distributed backdoor attack as a joint optimization process of three learning objectives, thereby exploiting the limit of this assumption. The research aims to evaluate the effectiveness of CerP and show that it can successfully evade existing defenses while remaining stealthy and achieving a high attack success rate.

BAFFLE \cite{sa24}, proposed in this paper, is a new defense mechanism that leverages FL to detect and secure against backdoor attacks. BAFFLE uses a feedback loop to integrate the diverse data sets from different clients in the FL process to uncover model poisoning. By incorporating the views of multiple clients, the proposed approach can achieve very high detection rates against state-of-the-art backdoor attacks, even with basic validation methods. The core idea of BAFFLE is to use data from multiple clients not only for training but also for identifying model poisoning. The contribution of this paper is to propose a new defense mechanism to improve the security of FL against backdoor attacks, which can be effective even with simple validation methods.

FL is susceptible to model poisoning attacks (MPAs) where malicious clients attempt to corrupt the global model by transmitting deceptively modified local model updates. Existing defenses, focused on model parameters, struggle to detect these attacks. In \cite{sa25}, propose FLARE, a robust model aggregation mechanism that leverages penultimate layer representations (PLRs) to evaluate the adversarial influence on local model updates. Trust evaluation method used in FLARE assigns trust scores based on pairwise PLR discrepancies, allowing FLARE to aggregate model updates by their trust scores.
\begin{table*}[]
\caption{Secure Aggregation in Trustworthy FL}
\label{tab:my-table7}
\resizebox{\textwidth}{!}{%
\begin{tabular}{@{}ccllccllc@{}}
\toprule
\textbf{\begin{tabular}[c]{@{}c@{}}Proposed \\ Technique\end{tabular}} &
  \textbf{YoP} &
  \multicolumn{1}{c}{\textbf{\begin{tabular}[c]{@{}c@{}}Accompanied \\ Algorithm\end{tabular}}} &
  \multicolumn{1}{c}{\textbf{Description}} &
  DC &
  CN &
  \multicolumn{1}{c}{\textbf{Problem Encountered}} &
  \multicolumn{1}{c}{\textbf{Benefits}} &
  \textbf{Data Set} \\ \midrule
\begin{tabular}[c]{@{}c@{}}BFL-CIDS\\ \cite{sa1}\end{tabular} &
  2020 &
  \begin{tabular}[c]{@{}l@{}}Public BC\\ Ethereum\end{tabular} &
  \begin{tabular}[c]{@{}l@{}}The scheme sends local training parameters to a\\ cloud computing center for global prediction and\\ alert filter identification module (RSP-AFI) is\\ introduced to filter out false alerts\end{tabular} &
   &
  \quad\circledmark[cyan!60] &
  \begin{tabular}[c]{@{}l@{}}Lack of Intrusion\\ detection system\end{tabular} &
  \begin{tabular}[c]{@{}l@{}}Efficient detection of\\ malicious attacks in\\ distributed\\ environments\end{tabular} &
  \begin{tabular}[c]{@{}c@{}}KDDCup99 \\ dataset\end{tabular} \\
\begin{tabular}[c]{@{}c@{}}CMFL\\ \cite{sa2}\end{tabular} &
  2022 &
  \begin{tabular}[c]{@{}l@{}}Committee\\ Consensus\\ protocol(CCP)\end{tabular} &
  \begin{tabular}[c]{@{}l@{}}It utilizes a committee system to monitor the\\ gradient aggregation process\end{tabular} &
  \circledmark &
   &
  \begin{tabular}[c]{@{}l@{}}Byzantine attack,\\ Backdoor Attack\end{tabular} &
  \begin{tabular}[c]{@{}l@{}}Faster convergence\\ and better accuracy\end{tabular} &
  \begin{tabular}[c]{@{}c@{}}FEMNIST, \\ Sentiment140,\\ Shakespeare\end{tabular} \\
\begin{tabular}[c]{@{}c@{}}Sniper\\ \cite{sa3}\end{tabular} &
  \multicolumn{1}{l}{2019} &
  \begin{tabular}[c]{@{}l@{}}Fed-Avg\\ Algorithm\end{tabular} &
  \begin{tabular}[c]{@{}l@{}}Sniper identifies benign local models by solving a\\ maximum clique problem, and suspected (poisoned)\\ local models are ignored\end{tabular} &
   &
  \quad\circledmark[cyan!60] &
  \begin{tabular}[c]{@{}l@{}}Model failure and\\ target error\\ poisoning attacks\end{tabular} &
  \begin{tabular}[c]{@{}l@{}}Better attack drop\\ rate\end{tabular} &
  MNIST \\
\begin{tabular}[c]{@{}c@{}}PBFT based \\ B-FL\\ \cite{sa4}\end{tabular} &
  2022 &
  \begin{tabular}[c]{@{}l@{}}PBFT\\ Consensus\\ protocol, MDP\\ DRL\end{tabular} &
  \begin{tabular}[c]{@{}l@{}}PBFT-based wireless B-FL architecture combines\\ permissioned blockchain and wireless FL system to\\ create a trustworthy AI model training environment\end{tabular} &
   &
  \quad\circledmark[cyan!60] &
  \begin{tabular}[c]{@{}l@{}}Byzantine attacks,\\ model tampering\\ attacks\end{tabular} &
  \begin{tabular}[c]{@{}l@{}}Reduced training\\ latency and\\ improved attack\\ handling\end{tabular} &
  \begin{tabular}[c]{@{}c@{}}MNIST,\\ CIFAR10,\\ Heart\\ activity\end{tabular} \\
\begin{tabular}[c]{@{}c@{}}BREA\\ \cite{sa5}\end{tabular} &
  2020 &
  \begin{tabular}[c]{@{}l@{}}distance-based\\ outlier removal\\ mechanism\end{tabular} &
  \begin{tabular}[c]{@{}l@{}}It employs a robust gradient descent approach that\\ enables secure computations over the secret shares\\ of the local updates to handle such attacks\end{tabular} &
   &
  \quad\circledmark[cyan!60] &
  Byzantine attack &
  \begin{tabular}[c]{@{}l@{}}Improved\\ convergence,\\ accuracy and privacy\end{tabular} &
  \begin{tabular}[c]{@{}c@{}}MNIST,\\ CIFAR10\end{tabular} \\
\begin{tabular}[c]{@{}c@{}}S-space\\ \cite{sa7}\end{tabular} &
  2022 &
  \begin{tabular}[c]{@{}l@{}}Stochastic\\ Descending\\ Gradient (SGD)\end{tabular} &
  \begin{tabular}[c]{@{}l@{}}A novel deep metric learning approach with S-space\\ for robust information extraction and a safer\\ learnable metric in FL applications\end{tabular} &
   &
  \quad\circledmark[cyan!60] &
  Byzantine attack &
  \begin{tabular}[c]{@{}l@{}}Robustness to\\ attacks\end{tabular} &
  \begin{tabular}[c]{@{}c@{}}Malimg,\\ MaleVis\end{tabular} \\
\begin{tabular}[c]{@{}c@{}}TDFL\\ \cite{sa8}\end{tabular} &
  2022 &
  \begin{tabular}[c]{@{}l@{}}Max Clique\\ Filtering\end{tabular} &
  \begin{tabular}[c]{@{}l@{}}TDFL uses a robust truth discovery aggregation\\ scheme to remove malicious model updates\end{tabular} &
   &
  \quad\circledmark[cyan!60] &
  Poisoning attacks &
  \begin{tabular}[c]{@{}l@{}}Reasonable\\ byzantine robustness\end{tabular} &
  \begin{tabular}[c]{@{}c@{}}MNIST,\\ FMNIST,\\ CIFAR10\end{tabular} \\
\begin{tabular}[c]{@{}c@{}}Fed-SCR\\  \\ Fed-GMA\\  \cite{sa9}\end{tabular} &
  2022 &
  Fed-GMA &
  \begin{tabular}[c]{@{}l@{}}Fed-SCR, for detecting faults and anomalies in\\ smart grids and  Fed-GMA, to handle the issue of\\ noisy gradients during aggregation\end{tabular} &
   &
  \quad\circledmark[cyan!60] &
  \begin{tabular}[c]{@{}l@{}}Anomaly detection\\ with small sized\\ labeled and\\ imbalance data\end{tabular} &
  \begin{tabular}[c]{@{}l@{}}Trustworthiness and\\ robust aggregation\end{tabular} &
  Power datasets \\
\begin{tabular}[c]{@{}c@{}}CerP\\  \cite{sa10}\end{tabular} &
  2023 &
  NA &
  \begin{tabular}[c]{@{}l@{}}The defensive mechanisms rely on the assumption\\ that a poisoned local model trained with poisoned\\ data is significantly biased compared to those\\ trained with poison-free data\end{tabular} &
   &
  \quad\circledmark[cyan!60] &
  Backdoor Attack &
  Improved defense &
  NA \\
\begin{tabular}[c]{@{}c@{}}FLTrust\\  \cite{sa11}\end{tabular} &
  2021 &
  \begin{tabular}[c]{@{}l@{}}Cosine\\ Similarity\end{tabular} &
  \begin{tabular}[c]{@{}l@{}}A new approach to bootstrapping trust in federated\\ learning by having the service provider itself collect\\ a small clean training dataset\end{tabular} &
   &
  \quad\circledmark[cyan!60] &
  \begin{tabular}[c]{@{}l@{}}Malicious and\\ byzantine clients\end{tabular} &
  \begin{tabular}[c]{@{}l@{}}Robust against\\ adaptive attacks\end{tabular} &
  \begin{tabular}[c]{@{}c@{}}MNIST,\\ CIFAR 10, \\ FMNIST,\\ HAR\end{tabular} \\
\begin{tabular}[c]{@{}c@{}}Ensemble FL\\ \cite{sa12}\end{tabular} &
  2021 &
  \begin{tabular}[c]{@{}l@{}}Clopper\\ Pearson method\end{tabular} &
  \begin{tabular}[c]{@{}l@{}}The approach builds on existing base FL algorithms\\ by training multiple global models, each learned\\ using a randomly selected subset of clients\end{tabular} &
   &
  \quad\circledmark[cyan!60] &
  \begin{tabular}[c]{@{}l@{}}Malicious and\\ byzantine clients\end{tabular} &
  \begin{tabular}[c]{@{}l@{}}Improved and\\ certified accuracy\end{tabular} &
  \begin{tabular}[c]{@{}c@{}}MNIST,\\ HAR\end{tabular} \\
\begin{tabular}[c]{@{}c@{}}DeTrust-FL\\  \cite{sa13}\end{tabular} &
  2022 &
  \begin{tabular}[c]{@{}l@{}}Differential\\ Privacy\\ mechanism\end{tabular} &
  \begin{tabular}[c]{@{}l@{}}Utilizes a decentralized trust consensus mechanism\\ and functional encryption scheme to ensure private\\ and secure model updates\end{tabular} &
  \circledmark &
   &
  \begin{tabular}[c]{@{}l@{}}Lack of\\ transparency,\\ disaggregation\\ attacks\end{tabular} &
  \begin{tabular}[c]{@{}l@{}}Improved training \\ time and reduced\\ volume of data\\ transferred\end{tabular} &
  CIFAR10 \\
\begin{tabular}[c]{@{}c@{}}BFLC\\  \cite{sa14}\end{tabular} &
  2020 &
  \begin{tabular}[c]{@{}l@{}}Committee\\ Consensus\\ (CCM)\end{tabular} &
  \begin{tabular}[c]{@{}l@{}}An innovative committee consensus mechanism is\\ devised to reduce consensus computing and\\ malicious attacks\end{tabular} &
  \circledmark &
   &
  \begin{tabular}[c]{@{}l@{}}Malicious clients,\\ central servers’\\ constant attack to the\\ global model\end{tabular} &
  \begin{tabular}[c]{@{}l@{}}Anti-malevolence,\\ High efficiency\end{tabular} &
  FEMNIST \\
\begin{tabular}[c]{@{}c@{}}MiTFed\\  \cite{sa15}\end{tabular} &
  2023 &
  \begin{tabular}[c]{@{}l@{}}Ethereum's\\ smart contracts,\\ SMPC\end{tabular} &
  \begin{tabular}[c]{@{}l@{}}BC and SDN are combined for building efficient\\ intrusion detection models\end{tabular} &
  \circledmark &
   &
  DDoS attack &
  \begin{tabular}[c]{@{}l@{}}Flexibility, tamper\\ proof, scalability\end{tabular} &
  NSL-KDD \\
\begin{tabular}[c]{@{}c@{}}FL-WBC\\  \cite{sa16}\end{tabular} &
  2021 &
  \begin{tabular}[c]{@{}l@{}}Coordinate\\ Median\\ aggregation\end{tabular} &
  \begin{tabular}[c]{@{}l@{}}A client-based defense that mitigates model\\ poisoning attacks in FL\end{tabular} &
   &
  \quad\circledmark[cyan!60] &
  Robustness issues &
  \begin{tabular}[c]{@{}l@{}}Effective attack\\ mitigation\end{tabular} &
  \begin{tabular}[c]{@{}c@{}}FMNIST,\\ CIFAR10\end{tabular} \\
\begin{tabular}[c]{@{}c@{}}BFL-AMC\\ \cite{sa17}\end{tabular} &
  2022 &
  \textit{\begin{tabular}[c]{@{}l@{}}Newton’s\\ Cooling Law {[}1{]}\end{tabular}} &
  \begin{tabular}[c]{@{}l@{}}AMC model is cooperatively trained by sharing\\ local model parameters with Blockchain\end{tabular} &
   &
  \quad\circledmark[cyan!60] &
  \begin{tabular}[c]{@{}l@{}}Poor anti-attack\\ capacity\end{tabular} &
  \begin{tabular}[c]{@{}l@{}}Improved anti-attack\\ ability\end{tabular} &
  Sim \\
\begin{tabular}[c]{@{}c@{}}Lightweight\\ hybrid FL \\ framework\\  \cite{sa18}\end{tabular} &
  2020 &
  \begin{tabular}[c]{@{}l@{}}HE, SMPC,\\ Smart Contract\\ (SC)\end{tabular} &
  \begin{tabular}[c]{@{}l@{}}Combines blockchain smart contracts with edge\\ training to manage trust, authentication, and\\ distribution of global or locally trained model\end{tabular} &
   &
  \quad\circledmark[cyan!60] &
  \begin{tabular}[c]{@{}l@{}}Quality of data and\\ integrity issues\end{tabular} &
  \begin{tabular}[c]{@{}l@{}}Wider adoption in\\ IoHT\end{tabular} &
  Sim \\
\begin{tabular}[c]{@{}c@{}}ST-BFL\\  \cite{sa19}\end{tabular} &
  2021 &
  HE, SC &
  \begin{tabular}[c]{@{}l@{}}A reputation system that allows authenticated and\\ authorized clients to query the history and\\ authenticity of the deep learning process\end{tabular} &
   &
  \quad\circledmark[cyan!60] &
  \begin{tabular}[c]{@{}l@{}}Malicious\\ aggregation server\end{tabular} &
  \begin{tabular}[c]{@{}l@{}}Improved\\ Input/output privacy\\ and verification\end{tabular} &
  \begin{tabular}[c]{@{}c@{}}MNIST,\\ FMNIST\end{tabular} \\
\begin{tabular}[c]{@{}c@{}}DDFL\\  \cite{sa20}\end{tabular} &
  2021 &
  \begin{tabular}[c]{@{}l@{}}Social group\\ Utility\\ maximization\end{tabular} &
  \begin{tabular}[c]{@{}l@{}}The framework involves clustering devices based on\\ social similarity, edge betweenness, and physical\\ distance.\end{tabular} &
  \circledmark &
   &
  \begin{tabular}[c]{@{}l@{}}Malicious\\ aggregation server\end{tabular} &
  \begin{tabular}[c]{@{}l@{}}Better robustness\\ and privacy\\ awareness\end{tabular} &
  MNIST \\
\begin{tabular}[c]{@{}c@{}}RFA\\  \cite{sa21}\end{tabular} &
  2022 &
  \begin{tabular}[c]{@{}l@{}}Smoothed\\ Weiszfeld\\ algorithm,\\ SMPC\end{tabular} &
  \begin{tabular}[c]{@{}l@{}}RFA comes in several variants, a fast one with a\\ single step of robust aggregation and one that adjusts\\ to heterogeneity through on-device personalization\end{tabular} &
   &
  \quad\circledmark[cyan!60] &
  Robustness issues &
  \begin{tabular}[c]{@{}l@{}}Communication\\ efficiency,\\ scalability\end{tabular} &
  EMNIST \cite{d18}\\
\begin{tabular}[c]{@{}c@{}}Turbo-\\Aggregate\\  \cite{sa22}\end{tabular} &
  2021 &
  \begin{tabular}[c]{@{}l@{}}Lagrange\\ coding\end{tabular} &
  \begin{tabular}[c]{@{}l@{}}Turbo-Aggregate uses a multi-group circular\\ strategy and additive secret sharing with novel\\ coding techniques to handle user dropouts\end{tabular} &
  \circledmark &
   &
  User dropout &
  \begin{tabular}[c]{@{}l@{}}Improved running\\ time, Reduced user\\ dropout rate\end{tabular} &
  \begin{tabular}[c]{@{}c@{}}Amazon\\ EC2 cloud\end{tabular} \\
\begin{tabular}[c]{@{}c@{}}CRFL\\  \cite{sa23}\end{tabular} &
  2021 &
  Markov Kernel &
  \begin{tabular}[c]{@{}l@{}}The proposed method controls the global model\\ smoothness by exploiting clipping and smoothing\\ on model parameters\end{tabular} &
   &
  \quad\circledmark[cyan!60] &
  \begin{tabular}[c]{@{}l@{}}Poisoning and\\ backdoor attack\end{tabular} &
  \begin{tabular}[c]{@{}l@{}}High Attack success\\ rate on backdoor\\ samples\end{tabular} &
  \begin{tabular}[c]{@{}c@{}}MNIST,\\ EMNIST\end{tabular} \\
\begin{tabular}[c]{@{}c@{}}BAFFLE\\  \cite{sa24}\end{tabular} &
  2022 &
  \begin{tabular}[c]{@{}l@{}}Local Outlier\\ Factor (LOF)\end{tabular} &
  \begin{tabular}[c]{@{}l@{}}BAFFLE uses a feedback loop to integrate the\\ diverse data sets from different clients in the FL\\ process to uncover model poisoning\end{tabular} &
   &
  \quad\circledmark[cyan!60] &
  Backdoor Attack &
  \begin{tabular}[c]{@{}l@{}}Improved detection\\ accuracy\end{tabular} &
  \begin{tabular}[c]{@{}c@{}}FEMNIST,\\ CIFAR10\end{tabular} \\
\begin{tabular}[c]{@{}c@{}}FLARE\\  \cite{sa24}\end{tabular} &
  2022 &
  \begin{tabular}[c]{@{}l@{}}Penultimate\\ Layer\\ Representations\\ (PLRs)\end{tabular} &
  \begin{tabular}[c]{@{}l@{}}Trust evaluation method used in FLARE assigns\\ trust scores based on pairwise PLR discrepancies\end{tabular} &
   &
  \quad\circledmark[cyan!60] &
  \begin{tabular}[c]{@{}l@{}}Model poisoning and\\ backdoor attack\end{tabular} &
  \begin{tabular}[c]{@{}l@{}}Attack defending\\ efficiency\end{tabular} &
  \begin{tabular}[c]{@{}c@{}}FEMNIST,\\ CIFAR10\end{tabular} \\ \\
\multicolumn{9}{l}{\textit{*CN: Centralized, DC: De-centralized}} \\ \bottomrule

\end{tabular}%
}
\end{table*}

\begin{table*}[]
\caption{Secure Data Aggregation in Trustworthy FL}
\label{tab:my-table8}
\resizebox{\textwidth}{!}{%
\begin{tabular}{@{}ccllccllc@{}}
\toprule
\textbf{\begin{tabular}[c]{@{}c@{}}Proposed \\ Technique\end{tabular}} &
  \textbf{YoP} &
  \multicolumn{1}{c}{\textbf{\begin{tabular}[c]{@{}c@{}}Accompanied \\ Algorithm\end{tabular}}} &
  \multicolumn{1}{c}{\textbf{Description}} &
  DC &
  CN &
  \multicolumn{1}{c}{\textbf{Problem Encountered}} &
  \multicolumn{1}{c}{\textbf{Benefits}} &
  \textbf{Data Set} \\ \midrule
\begin{tabular}[c]{@{}c@{}}EPPDA\\  \cite{da1}\end{tabular} &
  2022 &
  \multicolumn{1}{c}{HE} &
  \begin{tabular}[c]{@{}l@{}}Homomorphisms of secret sharing are adopted in\\ EPPDA to aggregate shared data without\\ reconstruction\end{tabular} &
   &
  \quad\circledmark[cyan!60] &
  Reverse attack &
  \begin{tabular}[c]{@{}l@{}}Efficient system\\ efficiency\end{tabular} &
  Sim \\
\begin{tabular}[c]{@{}c@{}}FLPDA\\ \cite{da2}\end{tabular} &
  2022 &
  \begin{tabular}[c]{@{}l@{}}PBFT, Paillier\\ Cryptosystem\\ (PC)\end{tabular} &
  \begin{tabular}[c]{@{}l@{}}FLPDA adopts data aggregation to protect the\\ changes made to individual user models in FL and\\ resist reverse analysis attacks\end{tabular} &
   &
  \quad\circledmark[cyan!60] &
  Information leakage &
  \begin{tabular}[c]{@{}l@{}}Ensures secure and\\ privacy-preserving\\ data aggregation\end{tabular} &
  Sim \\
\begin{tabular}[c]{@{}c@{}}PPDAFL\\ \cite{da3}\end{tabular} &
  2023 &
  \begin{tabular}[c]{@{}l@{}}PBFT\\ Consensus\\ algorithm, PC\end{tabular} &
  \begin{tabular}[c]{@{}l@{}}The PPDAFL scheme combines secret sharing with\\ FL to protect local model changes\\ and resist reverse analysis attacks\end{tabular} &
   &
  \quad\circledmark[cyan!60] &
  Data Island issue &
  \begin{tabular}[c]{@{}l@{}}Guaranteeing\\ Message\\ Authenticity and\\ integrity\end{tabular} &
  Sim \\ \\
\multicolumn{9}{l}{\textit{*CN: Centralized, DC: De-centralized}} \\ \bottomrule
\end{tabular}%
}
\end{table*}
\subsubsection{gradient based Secure Aggregation}

The authors propose a scheme called Sniper in \cite{sa3} to eliminate poisoned local models from malicious participants during training. This paper explores the relations between the number of poisoned training samples, attackers, and attack success rate in a FL system. Sniper identifies benign local models by solving a maximum clique problem, and suspected (poisoned) local models are ignored during global model updating. The authors observed that honest user models and attacker models are in different cliques, which they utilize to propose a filtering defense mechanism in Sniper. During every communication, the parameter server runs Sniper to filter parameters updated by attackers, dropping the attack success rate significantly even with multiple attackers present. The proposed scheme ensures secure aggregation of local models while preserving privacy.

A framework called BREA is proposed in \cite{sa5}, to address the challenge of secure and resilient FL against adversarial users. By protecting local updates with random masks, the true values are concealed from the server. However, Byzantine users can modify the datasets aor local updates to manipulate the global model. BREA is a single server secure aggrgation framework, that utilizes a verifiable outlier detection, integrated stochastic quantization, and secure model aggregation approach to achieve enhanced convergence, privacy, and Byzantine-resilience concurrently. The framework employs a robust gradient descent approach that enables secure computations over the secret shares of the local updates to handle such attacks. BREA also uses a distance-based outlier removal mechanism \cite{sa6}, to remove the effect of potential adversaries and ensure the selection of unbiased gradient estimators.
 
A novel deep metric learning method is presented in \cite{sa7}, using an auxiliary S-space to identify complex similarity regions in FL, aiming to extract reliable information from diverse data in FL situations. The primary novelty lies in an interpretable quantifier for deep metric learning aggregation in FL applications, resulting in a more secure learnable metric for these use cases.

To address vulnerability of FL to poisoning attacks and the limitations of existing defense strategies, a novel method called Truth Discovery based FL (TDFL) is proposed in \cite{sa8}. Unlike previous methods, TDFL can defend against multiple poisoning attacks without additional datasets and tolerate high proportions of Byzantine attackers. In the Honest-majority setting, TDFL uses a robust truth discovery aggregation scheme to remove malicious model updates, while in the Byzantine-majority setting, it employs a maximum clique-based filter to ensure global model quality. The proposed method is tested in different scenarios and under various types of poisoning attacks, including label flipping, arbitrary, Krum, Trim, and backdoor attacks. The experimental results demonstrate the robustness and effectiveness of TDFL, which can identify and filter malicious users with a 100\% detection rate. Additionally, TDFL does not require a validation dataset or a separate server model, making it a practical and effective solution for FL.
 
An FL framework Fed-SCR \cite{sa9}, for detecting faults and anomalies in smart grids with improved privacy and security, leverages a novel lightweight generative network, called SRC-GAN, for semisupervised learning of anomalous patterns from unbalanced power data. A new aggregation scheme, called Fed-GMA, is presented to handle the issue of noisy gradients during aggregation by replacing the averaging operation with a geometric median. Fed-GMA restricts the availability of active participants by granting selective node involvement, which enables different proportions of fog nodes to participate in each training round. This periodic aggregation approach leads to a decrease in the total number of communication rounds and the total communication overhead. Overall, this study provides a promising approach to improving the performance and privacy of FL-based fault and anomaly detection in smart grids.

FLTrust in \cite{sa11}, is a new Byzantine-robust FL method that bridges the gap between the server and the clients. It proposes a new approach to bootstrapping trust in FL by having the service provider itself collect a small clean training dataset, called the root dataset, and maintain a model, called the server model, based on it to bootstrap trust. FLTrust incorporates this root of trust in its new Byzantine-robust aggregation rule. The server assigns a trust score to each local model update, which is larger if the direction of the local model update is more similar to that of the server model update. FLTrust then normalizes each local model update to have the same magnitude as the server model update and computes the average of the normalized local model updates weighted by their trust scores as the global model update. By limiting the impact of malicious local model updates with large magnitudes, FLTrust provides a more secure and private way of aggregating local model updates.

FL-WBC in \cite{sa16} is a client-based defense that mitigates model poisoning attacks in FL. The defense is named White Blood Cell and perturbs the parameter space during local training to identify the space where the long-lasting attack effect on parameters resides. The defense provides a certified robustness guarantee against model poisoning attacks and a convergence guarantee to FedAvg after applying it.

The exponential growth of Internet of Things (IoT) devices has led to an increased use of FL for on-device ML. However, conventional FL is susceptible to privacy leakage, as a malicious aggregation server can infer sensitive information from end-devices' local learning model updates. To address this issue, a socially-aware device-to-device (D2D) communication-enabled distributed aggregation-based dispersed FL (DDFL) framework is proposed \cite{sa20}. This DDFL framework enables robust, privacy-aware, and efficient communication resource usage, with better privacy preservation than traditional FL. The framework involves clustering devices based on social similarity, edge betweenness, and physical distance. The proposed algorithm involves solving three sub-problems: clustering, resource allocation, and local accuracy minimization, using low complexity matching-theory-based solutions and a convex optimizer. The proposed algorithm is validated through numerical experiments, showing its superior performance in terms of learning model accuracy compared to traditional FL. The main contributions of this work are the socially-aware clustering-enabled DDFL framework, a novel clustering algorithm, and a loss function that simultaneously considers packet error rate and local learning accuracy.

The scalability of FL is hindered by the overhead of secure model aggregation across many users, with the existing protocols incurring quadratic overhead. Turbo-Aggregate \cite{sa22}, the first secure aggregation framework for FL, that achieves almost 50\% lower user dropout and reduced overhead for secure aggregation. A multi-group circular strategy along with additive secret sharing with novel coding techniques is used by Turbo-Aggregate, to handle user dropouts while ensuring user privacy. The framework has robustness guarantees, and its running time grows almost linearly with the number of users.

First Certifiably Robust FL (CRFL) framework has been proposed in \cite{sa23}, to train models that are robust against backdoor attacks in FL. Existing methods lack certification of their robustness, whereas CRFL provides a general framework to train models with sample-wise robustness certification on backdoors with limited magnitude. The proposed method controls the global model smoothness by exploiting clipping and smoothing on model parameters, which specifies the relation to FL parameters such as poisoning ratio on instance level, number of attackers, and training iterations. The training dynamics of the aggregated model via Markov Kernel are analyzed, and parameter smoothing is proposed for model inference. Overall, CRFL provides theoretical analysis and certified robustness against backdoor attacks, which aim to manipulate local models to fit the main task and backdoor task, achieving high attack success rates on backdoored data samples.

\subsubsection{Data Aggregation}
A privacy-preserving data aggregation mechanism for FL is proposed to resist reverse analysis attacks \cite{da1}. The proposed mechanism, called EPPDA, is based on secret sharing and can aggregate users' trained models secretly without revealing individual user data to the server. EPPDA also has efficient fault tolerance for user disconnections. The design goals of EPPDA are to protect a single user's model changes and prevent the server from initiating reverse analysis attacks. Homomorphisms of secret sharing are adopted in EPPDA to aggregate shared data without reconstruction. The mechanism enables the server to obtain an aggregated result without knowing anything about an individual user's trained model, thereby enhancing the privacy preservation of FL.\\
A FL-based PP data aggregation scheme (FLPDA) \cite{da2} for Industrial Internet of Things (IIoT) to protect data security and privacy. FLPDA adopts data aggregation to protect the changes made to individual user models in FL and resist reverse analysis attacks from industry administration centers. The PBFT consensus algorithm is used to select one of the IIoT devices in each round of data aggregation as the initialization and aggregation node. The scheme combines Paillier cryptosystem and secret sharing to achieve data fault tolerance and secure sharing. The proposed scheme does not rely on trusted authorities or third parties and has lower overhead in computation, communication, and storage, leading to higher efficiency and faster execution speed in data aggregation for IIoT. The main contribution of this paper is the FLPDA scheme that ensures secure and privacy-preserving data aggregation in IIoT, which can effectively address the challenges of data security and privacy protection in IIoT.\\
In \cite{da3}, researchers present PPDAFL, a secure data integration technique using FL tailored for IIoT applications. PPDAFL combines secret sharing and FL to shield local model updates and counter reverse engineering. Utilizing the PBFT algorithm, the system independently selects initiating and merging nodes. Fault tolerance is maintained even with multiple IIoT device failures or collusion through secret sharing. PPDAFL surpasses existing methods in communication and computation efficiency, speed, and effectiveness. The paper's key contributions include a secure multi-dimensional data integration approach based on FL, protection of local model changes, and combining the Paillier cryptosystem with secret sharing for data security and sharing. The PPDAFL method is well-suited for IIoT data integration scenarios. Table 7 and 8 provides an extensive overview of the research conducted on secure aggregation and data aggregation respectively in FL. 

\subsection{Privacy Preserving FL}
We categorize Privacy-aware Trustworthy FL works into more refined and detailed categories based on the main objectives and primary privacy and trustworthiness of strategies:
\subsubsection{Blockchain and Smart Contract based Privacy Preserving FL}

FL is a ML technique that aims to protect data privacy. However, it is prone to security threats, such as model inversion and membership inference, which need to be addressed when applied to Autonomous Vehicles (AVs). The research work in \cite{pv1} presents a significant contribution to the field of FL by proposing a novel privacy-preserving Byzantine-Fault-Tolerant (BFT) decentralized FL method for Autonomous Vehicles (AVs), called BDFL. The proposed BDFL method addresses the privacy leakage problem in FL while ensuring data security in AVs. BDFL employs a Peer-to-Peer (P2P) FL framework with BFT and uses the HydRand protocol and Publicly Verifiable Secret Sharing (PVSS) scheme to protect AVs' models. Experimental results demonstrate the effectiveness of BDFL in terms of accuracy and training loss on the MNIST dataset. Moreover, the decentralized P2P-FL framework built by the HydRand protocol ensures the method's robustness against node failures. The proposed BDFL method outperforms other BFT-based FL methods, which indicates its superiority in securing data privacy in AVs. Thus, the paper's contribution lies in proposing a feasible and effective FL method for secure model aggregation among AVs in a P2P network while maintaining data privacy.

Authors in \cite{pv3} addresses the privacy-preservation challenges in FL, a promising technique for facilitating data sharing while avoiding data leakage. The proposed approach is based on Homomorphic encryption and blockchain technology to address the Single Point of Failure (SPoF), gradient privacy, and trust issues. The Homomorphic encryption is used for gradient privacy protection. FL trust and BC storage issue are solved suing a reputation mechanism based on smart contract and on/off chain storage process.  The proposed approach is evaluated on the EMNIST dataset, and the results show that it achieves better performance than existing solutions in terms of accuracy and convergence speed while maintaining privacy-preservation.

A privacy-preserving and Byzantine-robust FL model based on blockchain is presented in \cite{pv4}. The proposed model adopts cosine similarity to detect malicious gradients uploaded by malicious clients and utilizes Fully Homomorphic Encryption (FHE) to provide secure aggregation. Additionally, the use of blockchain facilitates transparent processes and regulation enforcement. The proposed scheme achieves both efficiency and credibility, and extensive experiments demonstrate its robustness and efficiency. The main contributions of this work are the provision of a privacy-preserving training mechanism using the FHE scheme CKKS, the removal of malicious gradients via cosine similarity to provide a trusted global model, and the utilization of blockchain to facilitate transparent processes and enforcement of regulations. This work addresses the challenges of privacy preservation, poisoning attacks, and credibility in FL.

A decentralized blockchain-based FL model for ensuring trustworthiness and privacy in VNet systems has been proposed in \cite{pv5}. The framework utilizes a consensus method in the blockchain to enable ML on end devices without centralized training or coordination. A consensus protocol is adopted to guarantee consensus in the fog for critical vehicles. The proposed solution integrates both FL and blockchain to ensure data privacy and network security. The model ensures privacy by adopting blockchain capability along with FL through fog consensus. The adopted practical Byzantine Fault Tolerance protocol overcomes faulty issues. The framework guarantees privacy, trustworthiness, and adherence to delay requirements for vehicles in VNet systems. The proposed solution provides a decentralized approach for mutual ML models on end-devices, promoting privacy-preserving and secure ML.

PriMod-Chain, a framework that combines smart contracts, blockchain, FedML, DP, and IPFS to enhance privacy and trustworthiness in ML (ML) sharing in an IIoT setting is presented in \cite{pv12}. FedML is used as the global ML model sharing approach, while DP ensures privacy on the models. The framework includes smart contracts and EthBC for traceability, transparency, and immutability. IPFS provides low latency, fast decentralized archiving, and secure P2P content delivery. The framework was tested for its feasibility in terms of privacy, security, reliability, safety, and resilience. PriModChain proved to be a feasible solution for trustworthy privacy-preserving ML in IIoT systems and generated excellent results towards the five pillars of trustworthiness.

\subsubsection{Cryptography and Encryption based Privacy Preserving FL}

To addresses the challenge of preserving privacy while defending against poisoning attacks in FL, especially in edge computing environments, authors in \cite{pv2} proposes a differential privacy-based FL model designed for edge deployment that is resistant to poisoning attacks. The proposed model utilizes a weight-based detection algorithm that improves detection rates using small validation datasets, reducing communication costs. Moreover, the model protects the privacy of user data and model parameters on honest devices by leveraging differential privacy technology, with noise added dynamically to minimize disturbance and improve accuracy. The main contributions of the paper are a secure and privacy-preserving FL model for edge networks, a weight-based detection scheme to resist poisoning attacks, and an improved differential privacy technique for FL in edge networks. These contributions enable accurate neural network model training while maintaining privacy and security for both data and models in edge computing settings.

DetectPMFL \cite{pv7}, is a privacy-preserving momentum FL scheme for industrial agents, which considers the issue of unreliable agents. A detection method is designed to calculate the credibility of all agents while preventing the server from accessing the model parameters of the agents. Additionally, the privacy issues of convolutional neural networks (CNNs) are investigated, and the Cheon-Kim-Kim-Song (CKKS) encryption scheme is adopted to protect agents' data.

Biscotti \cite{pv6}, P2P distributed method for multi-party ML employs BC and cryptographic elements to facilitate a confidential ML procedure among collaborating clients. This strategy tackles the trust requirement in centralized systems and mitigates the risk of poisoning assaults by malevolent participants. The proposed solution is scalable, resilient to faults, and safeguards against recognized attacks. The effectiveness of this technique is demonstrated through evaluation, making it a cutting-edge solution for secure multi-party ML. However, Biscotti is a fully decentralized approach that provides privacy-preserving ML without the need for a trusted centralized infrastructure.

The main contribution of authors in \cite{pv11} is a decentralized FL scheme called PTDFL that enhances both privacy protection and trustworthiness while maintaining accuracy even in the presence of untrusted nodes. The scheme is designed for dynamic scenarios where nodes can join or leave at any time. To achieve these goals, the scheme uses lightweight primitives such as Lagrange interpolation and pseudorandom generators to encrypt gradients and ensure the trustworthiness of aggregated results. It also leverages zero-knowledge succinct non-interactive arguments of knowledge (zk-SNARK) to enhance the trustworthiness of gradients. Additionally, the paper proposes a novel local aggregation strategy that doesn't require a trusted third party to ensure the aggregated results' trustworthiness. Finally, PTDFL is also designed to support data owners joining and leaving during the FL task. The proposed scheme solves the problem of enhancing privacy protection and trustworthiness while maintaining accuracy in a dynamic and decentralized setting.

\subsubsection{Statistical based Privacy Preserving FL}

The authors presents FEDOBD in \cite{pv13}, a Federated Opportunistic Block Dropout approach that aims to reduce communication overhead while preserving model performance in large-scale deep models. FEDOBD divides models into semantic blocks and evaluates block importance, rather than individual parameter importance, to opportunistically discard unimportant blocks. The block importance measure is not based on the client loss function, enabling the approach to handle complex tasks. FEDOBD's evaluation shows that it outperforms the state-of-the-art method AFD, demonstrating its effectiveness in reducing communication overhead without compromising model performance. FEDOBD addresses the problem of communication overhead in FL by intelligently selecting and dropping unimportant blocks of a deep model.

Authors in \cite{pv8} proposes a framework called TP2SF that aims to maintain trustworthy privacy-preserving security in IoT networks. The framework consists of three main components: a trust management module, a two-level privacy-preserving module using enhanced Proof of Work (ePoW) and Principal Component Analysis (PCA), and an intrusion detection module using XGBoost. The proposed privacy-preserving module addresses inference and poisoning attacks by using ePoW and PCA, respectively. The framework also employs feature selection using Pearson correlation coefficient (PCC) to identify the relevant features for the smart city environment. The proposed framework demonstrates promising results in maintaining trustworthiness and privacy while detecting suspicious activities in IoT networks. Overall, TP2SF offers a comprehensive approach to privacy-preserving security in IoT, addressing various challenges in trust, privacy, and security.

\subsubsection{Trust and Reputation Management based Privacy Preserving FL}

A Deep Reinforcement Learning (DRL)-based reputation management mechanism for improving the security and reliability of FL has been proposed in \cite{pv9}. The proposed mechanism uses the concept of reputation as a metric to evaluate the reliability and trustworthiness of FL workers, and employs the DRL algorithm, Deep Deterministic Policy Gradient (DDPG), to improve the accuracy and stability of FL models. The paper compares the performance of the proposed method with conventional and DQN-based reputation methods, and demonstrates that the DRL-based mechanism outperforms the other two methods. The DDPG algorithm is capable of handling continuous and complex action spaces, and is used to detect unreliable workers in the FL environment. The reputation score of a worker is represented by the reliability of its local model updates, which is evaluated by the server using attack detection schemes. The proposed method addresses the problem of identifying reliable and trustworthy workers in FL, which is critical for FL security.

This authors proposes a trusted decentralized FL algorithm based on the concept of trust as a metric for measuring the trustworthiness of network entities in collaborative multi-agent systems \cite{pv14}. The proposed method updates trust relations among agents based on evidence of their contributive or non-contributive collaboration towards achieving specific goals. Trust estimates are used in decisions such as access control, resource allocation, and agent participation. The paper presents a mathematical framework for trust computation and aggregation and discusses its incorporation within a decentralized FL setup. The proposed algorithm enhances the security of FL training by enabling trust-based decision-making. Trust can be computed and aggregated based on the specific application, making the approach adaptable to various contexts.
Table 9 presents an extensive overview of the research conducted on privacy preserving and security aware trustworthiness in FL. This table highlights various methods, techniques, and approaches employed to ensure trustworthiness during the privacy preserving process.

SCFL in \cite{pv10}, is a Social-aware Clustered FL technique that attains equilibrium between data confidentiality and effectiveness by capitalizing on users' social relationships. It allows trusted parties to establish social groupings and merge their unprocessed model updates before uploading them to the cloud for comprehensive aggregation. Utilizing game theory, the social cluster organization is refined, and a just allocation system discourages free-riders. Additionally, an adaptable privacy safeguard is devised for clusters with minimal trust, enabling the dynamic cleansing of participants' model updates. An iterative, two-sided matching process results in an optimized disjoint partition with Nash-stable equilibrium. SCFL notably enhances model performance without sacrificing privacy, presenting an affordable and viable strategy for addressing the privacy and efficiency obstacles in FL.

\begin{table*}[]
\caption{Privacy Preserving Trustworthy FL}
\label{tab:my-table9}
\resizebox{\textwidth}{!}{%
\begin{tabular}{@{}cllllllll@{}}
\toprule
\textbf{\begin{tabular}[c]{@{}c@{}}Proposed \\ Technique\end{tabular}} &
  \multicolumn{1}{c}{\textbf{YoP}} &
  \multicolumn{1}{c}{\textbf{Description}} &
  \multicolumn{1}{c}{\textbf{\begin{tabular}[c]{@{}c@{}}Accompanied \\ Algorithm\end{tabular}}} &
  \multicolumn{1}{c}{\textbf{CN}} &
  \multicolumn{1}{c}{\textbf{DC}} &
  \multicolumn{1}{c}{\textbf{\begin{tabular}[c]{@{}c@{}}Problem \\ Encountered\end{tabular}}} &
  \multicolumn{1}{c}{\textbf{Outcomes}} &
  \multicolumn{1}{c}{\textbf{Dataset}} \\ \midrule
\begin{tabular}[c]{@{}c@{}}BDFL\\ \cite{pv1}\end{tabular} &
  2021 &
  \begin{tabular}[c]{@{}l@{}}Decentralized FL for AVs with\\ Byzantine fault tolerance\end{tabular} &
  \begin{tabular}[c]{@{}l@{}}P2P FL,\\ HydRand, PVSS\end{tabular} &
   &
  \quad\circledmark[cyan!60] &
  Privacy leakage &
  Improved accuracy &
  MNIST \\
\begin{tabular}[c]{@{}c@{}}Differential \\ privacy-based \\ FL\\ \cite{pv2}\end{tabular} &
  2022 &
  \begin{tabular}[c]{@{}l@{}}Privacy-focused FL for edge networks,\\ poison attack resistant\end{tabular} &
  \begin{tabular}[c]{@{}l@{}}Weight-based\\ detection\end{tabular} &
  \circledmark &
   &
  \begin{tabular}[c]{@{}l@{}}Poisoning\\ attacks\end{tabular} &
  Reduced comm. costs &
  \begin{tabular}[c]{@{}l@{}}MNIST, \\ FMNIST, \\ CIFAR10\end{tabular} \\
\begin{tabular}[c]{@{}c@{}}PP-FL \\ based on HE\\ \cite{pv3}\end{tabular} &
  2021 &
  \begin{tabular}[c]{@{}l@{}}FL using homomorphic encryption,\\ blockchain for privacy\end{tabular} &
  \begin{tabular}[c]{@{}l@{}}Smart contract,\\ on/off-chain\\ storage, HE\end{tabular} &
  \circledmark &
   &
  \begin{tabular}[c]{@{}l@{}}SPoF, gradient\\ privacy, trust\end{tabular} &
  Enhanced performance &
  EMNIST \\
\begin{tabular}[c]{@{}c@{}}PBFL\\ \cite{pv4}\end{tabular} &
  2022 &
  \begin{tabular}[c]{@{}l@{}}Byzantine-robust FL via blockchain and\\ cosine similarity\end{tabular} &
  \begin{tabular}[c]{@{}l@{}}FHE, CKKS,\\ blockchain\end{tabular} &
  \circledmark &
   &
  \begin{tabular}[c]{@{}l@{}}Privacy\\ preservation,\\ poisoning\\ attacks\end{tabular} &
  Robustness, efficient &
  \begin{tabular}[c]{@{}l@{}}MNIST, \\ FMNIST\end{tabular} \\
\begin{tabular}[c]{@{}c@{}}FL in VNet\\ \cite{pv5}\end{tabular} &
  2020 &
  \begin{tabular}[c]{@{}l@{}}Decentralized blockchain-based FL for\\ VNet systems\end{tabular} &
  \begin{tabular}[c]{@{}l@{}}Fog consensus,\\ PBFT\end{tabular} &
   &
  \quad\circledmark[cyan!60] &
  \begin{tabular}[c]{@{}l@{}}Trustworthines,\\ privacy\end{tabular} &
  Decentralized learning &
  Sim \\
\begin{tabular}[c]{@{}c@{}}Biscotti\\ \cite{pv6}\end{tabular} &
  2020 &
  \begin{tabular}[c]{@{}l@{}}P2P distributed ML using blockchain and\\ cryptography\end{tabular} &
  \begin{tabular}[c]{@{}l@{}}Private, secure\\ ML\end{tabular} &
   &
  \quad\circledmark[cyan!60] &
  \begin{tabular}[c]{@{}l@{}}Trust, poisoning\\ attacks\end{tabular} &
  Scalable, fault-resilient &
  \begin{tabular}[c]{@{}l@{}}MNIST, \\ Credit card \\ datset \cite{d15}\end{tabular} \\
\begin{tabular}[c]{@{}c@{}}DetectPMFL\\ \cite{pv7}\end{tabular} &
  2022 &
  \begin{tabular}[c]{@{}l@{}}Privacy-preserving momentum FL for\\ industrial agents\end{tabular} &
  CKKS encryption &
  \circledmark &
   &
  \begin{tabular}[c]{@{}l@{}}Unreliable\\ agents, privacy\end{tabular} &
  Credibility calculation &
  MNIST \\
\begin{tabular}[c]{@{}c@{}}TP2SF\\ \cite{pv8}\end{tabular} &
  2021 &
  \begin{tabular}[c]{@{}l@{}}Trustworthy privacy-preserving security\\ in IoT networks\end{tabular} &
  \begin{tabular}[c]{@{}l@{}}ePoW, PCA,\\ XGBoost\end{tabular} &
  \circledmark &
   &
  \begin{tabular}[c]{@{}l@{}}Inference,\\ poisoning\\ attacks\end{tabular} &
  Improved detection &
  \begin{tabular}[c]{@{}l@{}}BoT-IoT\cite{d16}, \\ ToN-IoT\cite{d17}\end{tabular} \\
\begin{tabular}[c]{@{}c@{}}DRL-based\\ Reputation\\ \cite{pv9}\end{tabular} &
  2022 &
  \begin{tabular}[c]{@{}l@{}}DRL-based reputation management for\\ secure, reliable FL in IoT\end{tabular} &
  \begin{tabular}[c]{@{}l@{}}Trustworthiness,\\ security\end{tabular} &
  \circledmark &
   &
  \begin{tabular}[c]{@{}l@{}}Unreliable FL\\ workers\end{tabular} &
  Improved accuracy &
  MNIST \\
\begin{tabular}[c]{@{}c@{}}SCFL\\ \cite{pv10}\end{tabular} &
  2022 &
  \begin{tabular}[c]{@{}l@{}}Social-aware clustered FL with\\ customized privacy preservation\end{tabular} &
  \begin{tabular}[c]{@{}l@{}}Game theory,\\ Nash equilibrium\end{tabular} &
  \circledmark &
   &
  \begin{tabular}[c]{@{}l@{}}Data\\ confidentiality,\\ effectiveness\end{tabular} &
  Enhanced performance &
  \begin{tabular}[c]{@{}l@{}}MNIST, \\ CIFAR10\end{tabular} \\
\begin{tabular}[c]{@{}c@{}}PTDFL\\ \cite{pv11}\end{tabular} &
  2023 &
  \begin{tabular}[c]{@{}l@{}}Decentralized FL scheme for privacy\\ protection and trustworthiness\end{tabular} &
  \begin{tabular}[c]{@{}l@{}}Lagrange\\ interpolation, PRG\end{tabular} &
   &
  \quad\circledmark[cyan!60] &
  \begin{tabular}[c]{@{}l@{}}Dynamic\\ scenarios,\\ untrusted nodes\end{tabular} &
  Maintained accuracy &
  \begin{tabular}[c]{@{}l@{}}MNIST, \\ CIFAR10\end{tabular} \\
\begin{tabular}[c]{@{}c@{}}PriModChain\\ \cite{pv12}\end{tabular} &
  2020 &
  \begin{tabular}[c]{@{}l@{}}Framework combining FedML, DP,\\ blockchain, smart contracts, IPFS\end{tabular} &
  \begin{tabular}[c]{@{}l@{}}IIoT, privacy,\\ security\end{tabular} &
  \circledmark &
   &
  \begin{tabular}[c]{@{}l@{}}Privacy, trust in\\ IIoT\end{tabular} &
  Trustworthiness &
  MNIST \\
\begin{tabular}[c]{@{}c@{}}FEDOBD\\ \cite{pv13}\end{tabular} &
  2022 &
  \begin{tabular}[c]{@{}l@{}}Reduces comm. overhead by\\ opportunistically dropping unimportant\\ blocks\end{tabular} &
  \begin{tabular}[c]{@{}l@{}}Block dropout,\\ deep models\end{tabular} &
  \circledmark &
   &
  \begin{tabular}[c]{@{}l@{}}Communication\\ overhead\end{tabular} &
  Preserved performance &
  \begin{tabular}[c]{@{}l@{}}CIFAR10, \\ CIFAR 100,\\ IMDB\end{tabular} \\
\begin{tabular}[c]{@{}c@{}}Trusted\\ Decentralized \\ FL\\ \cite{pv14}\end{tabular} &
  2022 &
  \begin{tabular}[c]{@{}l@{}}Trust-based decision-making for secure\\ decentralized FL\end{tabular} &
  \begin{tabular}[c]{@{}l@{}}Trust\\ computation,\\ aggregation\end{tabular} &
   &
  \quad\circledmark[cyan!60] &
  \begin{tabular}[c]{@{}l@{}}Trustworthiness\\ in multi-agent\\ systems\end{tabular} &
  Enhanced security &
  Sim\\ \\
\multicolumn{9}{l}{\textit{*CN: Centralized, DC: De-centralized}} \\ \bottomrule
\end{tabular}%
}
\end{table*}

\subsection{Discussion}

The main limitations and pitfalls in the current literature on trustworthy verifiability in FL are the reliance on trusted third parties, efficiency trade-offs, and scalability limitations. To address these issues and improve the learning process, future research should focus on the development of trustless solutions, addressing scalability challenges, and optimizing privacy-preserving techniques. Additionally, researchers should explore adaptive and context-aware mechanisms that consider the unique characteristics of various application scenarios and adapt the learning process accordingly. A common limitation in secure and verifiable FL frameworks \cite{v1, v2, v3, v4, v5, v6, v7}  is the reliance on trusted third parties for authentication and aggregation. This dependence could introduce security vulnerabilities and single points of failure. Future research should focus on developing trustless solutions that eliminate the need for third parties and improve the overall security of FL systems. FL systems incorporating trust \cite{v8, v9, v10, v11, v12, v13}  often rely on blockchain technology to enhance transparency and trustworthiness. However, these papers fail to consider the efficiency trade-offs and scalability limitations of blockchain-based solutions. Future research should address these limitations and explore more efficient and scalable alternatives that can support large-scale FL systems. Privacy-preserving FL with additional features \cite{v13,v14,v15, v16,v17,v18,v19}  tackles a diverse range of issues, such as privacy preservation, IP protection, and explainability. However, these studies often overlook the performance overhead introduced by their proposed solutions. For instance, privacy-preserving techniques like homomorphic encryption and verifiable computing employed in \cite{v13, v14, v17} can be computationally expensive. Future research should focus on optimizing these techniques to balance privacy and efficiency without compromising the learning process.

Security and attack resistance are major concerns in FL. Some papers focus on detecting and preventing poisoning attacks \cite{sa3, sa7, sa10, sa16, sa24}  or malicious clients \cite{sa5, sa6, sa11, sa12, sa13, sa17}. However, most of these methods assume some degree of trust in the clients or the aggregator, which may not hold in real-world applications. To enhance trustworthiness, a critical review should consider the assumptions made by these approaches and examine their robustness to various threat models. The research work \cite{da1, da2} proposes different schemes to protect data privacy during FL, but they may overlook the trade-off between privacy and learning efficiency. Moreover, they may not adequately address the issue of model robustness against adversarial attacks. The decentralized and blockchain-based methods have been investigated in \cite{sa4, sa14, sa15, sa18, sa19, sa20}. These approaches aim to enhance the trustworthiness of FL by leveraging decentralized consensus mechanisms. However, they may suffer from scalability and communication overhead issues. Additionally, the reliance on blockchain technologies may introduce new vulnerabilities and limitations.

\section{Open Issues and Future Research Directions}
In this section, we aim to emphasize the primary challenges related to ensuring Trustworthy FL. Additionally, we will identify potential areas of future research that warrant further exploration. By doing so, we hope to provide researchers with a clearer understanding of the crucial aspects that require attention within the realm of FL. Addressing these key challenges in Trustworthy FL is essential for ensuring the robustness, reliability, and widespread adoption of this technology. As research progresses, it is crucial to focus on these aspects to develop improved methods and techniques in FL.
\subsection{Interpretability Aware Trustworthy FL }

Trustworthy FL faces several key challenges that must be addressed to ensure its effectiveness and reliability. Some of these challenges are related to interpretability, model selection, feature and sample selection, and data sharing.

\subsubsection{Interpretability} Ensuring that the models used in FL are interpretable is crucial for building trust among stakeholders. This involves making the decision-making process of the models transparent and understandable to both technical and non-technical users. Developing techniques to improve the interpretability of complex models, such as deep learning architectures, is an ongoing challenge in this area.
\subsubsection{Model selection} Choosing the most appropriate model for a particular task in a federated setting is a significant challenge. Researchers must consider factors such as the heterogeneity of data across participating devices, communication constraints, and computational resources available on each device. Developing robust and efficient methods for model selection in FL is essential for ensuring the quality and reliability of the learning process.
FL faces challenges that demand accurate and efficient model update validation methods for non-IID datasets, to detect poisoning attacks. Furthermore, optimizing the number of workers in FL is necessary to balance performance and resource costs.
\subsubsection{Feature and sample selection}In FL, the data distribution across devices can be non-uniform, leading to potential issues with feature and sample selection. Addressing these challenges requires the development of robust methods to handle such heterogeneity and to identify the most relevant features and samples for model training. This, in turn, can improve the overall performance and efficiency of the learning process.
\subsubsection{Data sharing} Trustworthy FL relies on the secure and privacy-preserving sharing of data between devices. This involves designing protocols that allow devices to share information without revealing sensitive information about individual data points or compromising user privacy. Developing methods for secure and efficient data sharing, while maintaining the utility of the shared information, is a critical challenge in the field of FL.

\subsection{Fairness Aware Trustworthy FL}
Challenges in Trustworthy FL related to fairness include addressing biases and discrimination in training data or procedures, which can reduce the fairness of the resulting model. Ensuring a model is fair involves training it with balanced and unbiased data, allowing for generalization across the entire class distribution. Challenges in Trustworthy FL related to fairness include quantifying the impact of fairness on the final accuracy and convergence speed. The lack of existing literature providing theoretical analysis of fairness in FL highlights the need for further investigation into this issue.
\subsubsection{Client Selection}
There are several key issues regarding client selection in FL that include a decrease in performance with increasing clients, a communication bottleneck due to multiple transactions, and the possibility of off-chain collusion by malicious clients, which cannot be defended against by the proposed commitment hash scheme.

Selecting FL worker nodes based on reputation is challenging due to the cold start problem, where historical interaction data is required for reputation evaluation. The potential tampering of reputation values also introduces uncertainty. One approach to mitigate this problem is to use contract theory.

In addition, the simulation of the straggler effect in the FL process involved the inclusion of various straggler robots that fail to transmit their local model update within a specific time, leading to a decrease in the overall accuracy of the global model.

FL faces critical challenges including: 1) the assumption that all workers participate in training unconditionally, which is not realistic as some may refuse due to data privacy concerns or lack of incentives; 2) the risk of malicious attacks by workers or free riding behavior where workers provide fake parameters to improve their reputation; 3) the presence of a parameter server resulting in remote data communication and the difficulty of finding a trusted third-party server, limiting the application of FL. In the context of Trustworthy FL, there are several challenges that need to be addressed to ensure a more secure and reliable learning process. These challenges include:

\paragraph{Attack susceptibility} FL processes are vulnerable to various types of attacks, such as poisoning attacks, where malicious workers send incorrect local model updates to mislead the learning process.
\paragraph{Unreliable local updates} Benign workers may unintentionally contribute unreliable local updates due to factors such as unstable channel conditions, device mobility, and energy constraints.
\paragraph{Model convergence issues} The presence of unreliable workers with erroneous local training data can hinder model convergence or prolong convergence time compared to situations involving only reliable workers.
\paragraph{Dynamic device behaviors} Real-world FL deployments often involve devices with changing behaviors, transitioning between reliable and unreliable or malicious states, making it difficult to maintain a stable learning environment.
Offline and dropout workers: Workers may go offline or drop out of the FL task due to unstable network connections, high device mobility, or energy constraints, which can negatively impact the overall performance of the learning task.
\paragraph{Vulnerability to adversarial attacks} Addressing the issue of FL susceptibility to attacks, such as poisoning attacks, where malicious workers provide incorrect local model updates, remains a significant challenge.
\paragraph{Malicious worker withdrawal} Detected malicious workers may attempt to withdraw from the FL task, further complicating the learning process.
Real-time monitoring challenges: Existing trust models often lack dynamic monitoring mechanisms, making it difficult for FL parameter servers to monitor worker behaviors and minimize the adverse effects from malicious workers in real-time.

\subsubsection{Contribution evaluation and Incentive Mechanism}
A key challenge in FL is designing a fair reward system that compensates local devices proportionally to their data contributions, while addressing issues such as devices' unwillingness to federate and potential dishonest behavior from untruthful devices that could lead to inaccurate global model updates.

\paragraph{Incentivizing participation} Identifying appropriate incentives to encourage clients' participation and sharing of model updates, considering the system costs and security risks involved, is a critical aspect of future FL research.

\paragraph{Evaluating clients' contributions} Developing accurate, efficient, and fair methods to evaluate each client's contribution within FL systems, considering the unique challenges of this learning environment, is crucial for effective incentive mechanisms.

\paragraph{Real-time monitoring} Future research should focus on devising real-time monitoring mechanisms that enable FL parameter servers to detect and mitigate malicious or unreliable worker behaviors, ensuring a secure and reliable learning process.

\paragraph{Reward allocation and fairness} Designing reward allocation schemes that account for the quality of each participant's contribution, maintain collaborative fairness, and offer suitable incentives to retain high-quality participants is vital for FL success.

\paragraph{Free Rider problem}
The free rider issue is a common problem in collaborative systems, including FL, where some participants, known as free riders, take advantage of the shared resources, knowledge, or benefits without contributing to the system. Free riders can negatively impact FL in various ways such as Inequitable distribution of resources, reduced overall performance, and lower participant motivation.

To tackle the free rider problem in FL, future research should concentrate on a few crucial areas. Creating innovative incentive mechanisms can encourage active participation by rewarding contributors, thereby discouraging free riders. Establishing reputation systems can identify free riders and promote positive involvement by assessing and ranking participants based on their input. Strengthening secure and privacy-preserving communication channels can foster trust and reduce free riding tendencies. Moreover, consistently monitoring participant contributions and imposing penalties on identified free riders can prevent potential exploitation of the system.

\paragraph{Model performance and bias:} Future research should emphasize designing reward systems that incentivize both high predictive performance and low bias in FL models, especially in sensitive applications such as healthcare.

\subsection{Security and Privacy Aware Trustworthy FL}

In this section, we outline the key challenges in security and privacy for Trustworthy FL along with the potential future research directions:
\paragraph{Distributed poisoning attacks}
Developing robust defenses against coordinated poisoning attacks, particularly when malicious clients collude, is a significant challenge. Future research should investigate strategies to identify and mitigate such attacks in FL environments.

\paragraph{Byzantine-resilience and privacy preservation}
Reconciling the need for Byzantine-resilience with preserving privacy in FL environments presents a complex problem. Research should focus on developing frameworks that strike a balance between robustness and privacy protection. urrent Byzantine-robust FL methods are still vulnerable to local model poisoning attacks. Research should explore establishing a root of trust to prevent malicious clients from corrupting global models.

\paragraph{Secure aggregation and malicious clients}
Existing secure aggregation schemes are often based on semi-honest assumptions, making them vulnerable to malicious clients. Future work should explore mechanisms to enhance the security of FL systems against malicious clients while minimizing communication overhead.

\paragraph{Training integrity and decentralization} Ensuring training integrity in FL is crucial, as incomplete or lazy participation can degrade model accuracy. Research should focus on designing privacy-preserving and verifiable decentralized FL frameworks to guarantee data privacy and training integrity while addressing trust concerns and communication bottlenecks.

\paragraph{Defending against data and model poisoning attacks} FL faces challenges from data poisoning and model poisoning attacks. Investigating appropriate defenses to protect model performance and detect anomalies is crucial for future research.

\paragraph{Resource allocation in FL} Efficient resource allocation schemes are needed to enable the participation of more devices in FL and maintain learning accuracy without significantly increasing convergence time.

\paragraph{Robustness against server failures} The centralized aggregation server in FL might fail due to physical damage or security attacks. Future research should investigate methods to ensure the robustness of FL systems in such scenarios.

\paragraph{Scalability of secure model aggregation}
The overhead of secure model aggregation creates a bottleneck in scaling secure FL to large numbers of users. Research should focus on developing efficient and scalable secure aggregation protocols.

\paragraph{Handling user dropouts and unavailability} 
FL systems need to be robust against user dropouts and unavailability, which can lead to privacy breaches and degraded performance. Future work should focus on designing protocols that can handle user dropouts and maintain privacy guarantees.

\paragraph{Certifiable robustness against backdoor attacks} 
The goal of certifiable robustness in FL is to protect the global model against adversarial data modifications made to local training datasets. Research should explore methods to certify robustness in FL systems while defending against backdoor attacks.

\paragraph{Security and privacy in edge computing deployments}
Addressing security and privacy threats in FL approaches for edge computing, such as resource-constrained IoT devices and data privacy disclosure, is crucial for future research.

In FL, ensuring trustworthy security and privacy presents multiple challenges. Developing a Byzantine-resilient and privacy-preserving FL framework is crucial to protect against collusion between malicious clients engaging in distributed poisoning attacks. Additionally, there is a need to design privacy-preserving and verifiable decentralized FL frameworks that can guarantee data privacy and training integrity while alleviating trust concerns and communication bottlenecks caused by centralization. Addressing the impact of non-IID data on convergence and model performance is essential for effective learning in real-world scenarios. Efficient resource allocation schemes are required to increase the number of participating devices and maintain robustness against user dropouts and unavailability while preserving privacy.

Moreover, defending against various types of attacks, such as data poisoning, model poisoning, and backdoor attacks, is necessary to maintain model performance and integrity. This requires achieving certifiable robustness against adversarial data modification in local training data. In edge computing deployments for FL, balancing robustness and latency is critical to prevent interruptions due to physical damage or security attacks. Overcoming the overhead of secure model aggregation and its scalability is essential for practical applications. Finally, addressing challenges in social cluster-based FL, such as forming stable and optimized social cluster structures, quantifying contributions, ensuring fair revenue allocation, and designing flexible and differentiated perturbation mechanisms, is vital to strike a balance between privacy and utility.

\section{Conclusion}
Federated Learning (FL) is a significant development in AI, enabling collaborative model training across distributed devices while maintaining data privacy. Addressing trustworthiness issues in FL is crucial. In this survey, we present a comprehensive overview of Trustworthy FL, exploring existing solutions and well-defined pillars. While there is substantial literature on trustworthy centralized ML/DL, more efforts are needed to identify trustworthiness pillars and metrics specific to FL models and develop relevant solutions. We propose a taxonomy encompassing three main pillars: Interpretability, Fairness, and Security \& Privacy. We also discuss trustworthiness challenges and suggest future research directions, providing valuable insights for researchers and practitioners in the field.


%

\section*{Acknowledgment}

This work is supported by the Zayed health science center under fund 12R005.

\ifCLASSOPTIONcaptionsoff
  \newpage
\fi

\end{document}